\documentclass{article}

\usepackage{microtype}
\usepackage{graphicx}
\usepackage{subcaption}

\usepackage{booktabs} % for professional tables

\usepackage{hyperref}

\usepackage[accepted]{icml2024}

\usepackage{amsmath}
\usepackage{amssymb}
\usepackage{mathtools}
\usepackage{amsthm}

\usepackage[capitalize,noabbrev]{cleveref}

\theoremstyle{plain}

\theoremstyle{definition}

\theoremstyle{remark}

\usepackage[textsize=tiny]{todonotes}

\icmltitlerunning{Bias of SGD or the Architecture: Disentangling the Effects of Overparameterization of Neural Networks}

\usepackage{setting}

\begin{document}

\twocolumn[
\icmltitle{Bias of Stochastic Gradient Descent or the Architecture: Disentangling the Effects of Overparameterization of Neural Networks}



\icmlsetsymbol{equal}{*}

\begin{icmlauthorlist}
\icmlauthor{Amit Peleg}{uni,ai}
\icmlauthor{Matthias Hein}{uni,ai}
\end{icmlauthorlist}

\icmlaffiliation{uni}{University of Tübingen, Germany}
\icmlaffiliation{ai}{Tübingen AI Center, Germany}

\icmlcorrespondingauthor{Amit Peleg}{amit.peleg@uni-tuebingen.de}

\icmlkeywords{Machine Learning, ICML}

\vskip 0.3in
]



\printAffiliationsAndNotice{}  % leave blank if no need to mention equal contribution

%\input{0_abstract}
\begin{abstract}
Neural networks typically generalize well when fitting the data perfectly, even though they are heavily overparameterized. 
Many factors have been pointed out as the reason for this phenomenon, including an implicit bias of stochastic gradient descent (SGD) and a possible simplicity bias arising from the neural network architecture. 
The goal of this paper is to disentangle the factors that influence generalization stemming from optimization and architectural choices by studying \emph{random} and \emph{SGD-optimized} networks
that achieve zero training error.
We experimentally show, in the low sample regime, that overparameterization in terms of increasing width is beneficial for generalization, and this benefit is due to the bias of SGD and not due to an architectural bias. In contrast, for increasing depth, overparameterization is detrimental for generalization, but random and SGD-optimized networks behave similarly, so this can be attributed to an architectural bias. 
\end{abstract}

\section{Introduction}
\label{sec:intro}

The generalization of neural networks challenges common wisdom in classical statistical learning theory \cite{VCdimension}: the number of parameters in today's neural networks is much larger than necessary to fit the data \cite{zhang2017understanding}, and further increasing network size, i.e., more overparameterization, yields better generalization \cite{hestness2017deep, allen2019convergence}. 
The underlying mechanisms of generalization despite overparameterization remain, to a large extent, an open question.

While the contribution of certain aspects has been examined in the past (e.g., batch size~\cite{keskar2016large} and learning rate~\cite{li2019towards}), isolating the effect of gradient-based optimization is more challenging due to its integral role in today's well-performing networks.
The implicit bias of Stochastic Gradient Descent (\sgd{}) is thus often thought to be the main reason behind generalization \cite{arora2019implicit,shah2020pitfalls}.
A recent thought-provoking study by \citet{chiang2022loss} suggests the idea of the volume hypothesis for generalization: 
well-generalizing basins of the loss occupy a significantly larger volume in the
weight space of neural networks than basins that do not generalize well.
They argue that the generalization performance of neural networks is primarily a bias of the architecture and that the implicit bias of \sgd{} is only a secondary effect. To this end, they randomly sample networks that achieve zero training error (which they term Guess and Check (\gnc)) and argue that the generalization performance of these networks is qualitatively similar to networks found by \sgd{}. 

In this work, we revisit the approach of \citet{chiang2022loss} and study it in detail to disentangle the effects of implicit bias of \sgd{} from a potential bias of the choice of architecture. As we have to compare to randomly sampled neural networks, our analysis is restricted to binary classification tasks in the low sample regime. Specifically, we analyze the behavior of the LeNet~\cite{lecun1998gradient}, Multi-Layer Perceptron (MLP), and ResNet~\cite{he2016deep} architectures.

In summary, we make the following contributions:
\begin{itemize}[leftmargin=*]
    \item We show that some findings of \citet{chiang2022loss} are based on a sub-optimal initialization of \sgd{}, which makes \sgd{} artificially worse. Moreover, we argue that their normalized loss cannot be used to compare networks of different architectures. We suggest an alternative normalization and analysis scheme that allows us to study the effects of increasing overparameterization (in terms of width and depth).
    \item We show that increasing overparameterization in terms of width improves the generalization of \sgd{}, while for randomly sampled networks, it remains mostly unaffected. This indicates a clear implicit bias of \sgd{}.
    \item On the contrary, increasing overparameterization in terms of depth is detrimental to generalization both for \sgd{} and \gnc{}. As they behave similarly, 
    the negative impact of increased depth can be mainly attributed to a bias in the architecture. 
\end{itemize}
Note that we do not claim that increasing depth is always detrimental to generalization, but rather that improvements regarding depth are due to architectural bias and not due to \sgd{}. In contrast, increasing width, even for randomly sampled networks, does not harm generalization, and the implicit bias of \sgd{} leads to networks with higher margins, significantly improving generalization.
\section{Related Work}
Understanding the generalization properties of deep neural networks is a long-standing research topic that has been tackled from different angles \cite{Cybenko1989ApproximationBS,HORNIK1991251UniversalApprox,Wolpert1995Mathematics,bartlett2002rademacher,Hoffer2017TrainLonger,jakubovitz2019generalization}. 

\textbf{Overparameterization and generalization.} Traditional learning theory with complexity measures, e.g., the VC dimension \cite{VCdimension}, suggests that for generalization, models should not be able to fit any possible training data  \cite{UnderstandingMLbook}. In particular, heavily over-parametrized models should overfit the data \cite{hastie2001elements}. However, even though networks are large enough to fit the data perfectly \cite{Haeffele17GlobalOptimality,nguyen18aOptLandscape}, we essentially observe the opposite \cite{Belkin2018ReconcilingMM,neal2019a,bartlett20Benign}. 
Specifically, \citet{zhang2017understanding} showed that a CNN architecture can obtain perfect train accuracy both on random and non-random labels on the CIFAR10 dataset, proving that the models are expressive enough to memorize 
while still being able to generalize.
Additionally, \citet{arpit2017closer} illustrated that the networks do not just memorize the data but learn simple patterns first, and \citet{neal2019a} reported that both bias and variance can decrease as the
number of parameters grows. 

\textbf{Implicit bias from the optimizer.} The success of modern deep learning is often attributed to the implicit bias of the optimizers \cite{neyshabur2015RealIndBias}: \citet{soudry2018ImplBiasGD} showed that SGD on linearly separable data converges to the maximum-margin linear classifier, and \citet{arora2019implicit} 
hypothesized that the implicit regularization of gradient-based optimizers goes beyond what can be expressed by standard regularizers.
\citet{galanti2022sgd} investigated the effects of small batch size on the network's rank. \citet{liu2020understanding} related the gradient signal and noise to generalization, \citet{advani2020high} highlighted that learning with gradient descent effectively takes place in a subspace of the weights, and \citet{Lee2020WideNetworks} observed simplified learning dynamics for wider networks. 
\citet{andriushchenko2023LargeLRsgd} showed that large step-size SGD leads to low-rank features.

\textbf{Loss landscape perspectives.} Several works connect generalization properties to the geometry of the loss landscape, often involving sharpness quantities \cite{dziugaite2017ComputingNonVacGB,keskar2017largeBatch,Jiang2020Fantastic,foret2021SAM}. The relevance of this relation, however, is unclear \cite{andrishchenko23sharpness}. 
This connection between flatness and well-performing models is also present in the Bayesian literature \cite{izmailov18AveragingWeights,Wilson2020BDL}.

\textbf{The volume hypothesis.} 
\citet{Perez18simplefunctions} observed a bias towards simple functions in deep networks, irrespective of optimization. \citet{huang2020understanding}  hypothesized that the volume of bad minima in weight space might be smaller than that of well-generalizing minima.  \citet{mingard21isSGDBayesian} further argued that, in simplified terms, SGD behaves like a Bayesian sampler and that the inductive bias in deep learning does not primarily stem from the optimizer. However, their approximation requires infinite width and thus does not apply to practical networks. \citet{geiping2022stochastic} questioned the relevance of stochasticity in SGD by illustrating that gradient descent with explicit regularization can perform comparably. \citet{chiang2022loss} argued that the implicit bias of \sgd{} is only a secondary effect, and the main contribution to generalization comes from the architecture. 
\section{Setup}
\label{sec:method}
Our main goal in this work is to disentangle the architectural bias -- the bias inherent in the function class obtained by randomly sampling the network parameters -- from the bias of \sgd{} regarding generalization in the overparameterized setting. Our setup is similar to \citet{chiang2022loss}. However, we deviate in our evaluation, arguing below that their conclusions are potentially inaccurate due to issues in their application of \sgd{} and the quantities they considered.
To achieve our objective, we utilize a random sampling procedure, termed \gnc{} in \citet{chiang2022loss}, as well as \sgd{} to attain neural networks that achieve zero training error. However, due to the exponentially increasing computational cost of \gnc{}, we limit our study to binary classification tasks from MNIST \cite{lecun1998gradient} and CIFAR10 \cite{krizhevsky2009learning} in the low sample regime.

\subsection{Neural Network Architecture}
\label{sec:norm}
Following the setup of
\citet{chiang2022loss}, we focus on
versions of the LeNet architecture \cite{lecun1998gradient} in all of our experiments. 
Additionally, we confirm our main results with
an MLP and a small ResNet~\cite{he2016deep}.
Details on the architectures can be found in Appendix~\ref{sec:network_arch}.
For the LeNet, the ReLU activation function, $a(z)=\max\{z,0\}$, is applied after each layer (except for the pooling layers).
The resulting function $f:\R^d \rightarrow \R^K$ induced by this network can be written as,
\[ f(W_1,\ldots,W_L,x)=W_L \circ a \circ W_{L-1} \circ \ldots \circ a \circ W_1 x,\] with input dimension $d$ and $K$ classes.
Note that the output vector of $f$ contains the logits of the classes.

When using the ReLU activation function, the function $f$ is one-homogeneous in the weights $W_l$ of each layer and in the input $x$. That is, for each $\gamma>0$, 
\begin{equation*}
\label{eq:homogenous}
f(W_1,\mydots,\gamma W_l,\mydots,W_L,x)=\gamma f(W_1,\mydots,W_l,\mydots,W_L,x).
\end{equation*}
For compactness, we use the notation $f(W,x)$, where $W=(W_1,\ldots,W_L)$.

We use the $\{-1,1\}$ label encoding for binary classification and denote the two components of $f$ as $f_1$
and $f_{-1}$. Let $\ST=(x_i,y_i)_{i=1}^n$ be the training set, where $x_i \in \R^d$, and $y_i \in \{-1,1\}$. The one-homogeneity of the ReLU network implies that if a network realizes zero-training error, that is 
\begin{equation*}
y_i \left(f_1(W,x_i)-f_{-1}(W,x_i)\right)\geq 0, \quad i=1,\ldots,n,
\end{equation*}
and one uses cross-entropy loss (which reduces to logistic loss in the binary case), 
\[ l(y,f(W,x)) = \log\left(1+e^{-y_i(f_{1}(W,x_i)-f_{-1}(W,x_i))}\right),\]
then simply upscaling the weights of any layer to infinity yields zero loss. 
Since upscaling the weights does not change the classifier, the training loss itself is not a suitable criterion for distinguishing two ReLU networks that achieve zero training error. 
Therefore, as the goal of this paper is to study the overparameterized regime, where the network is large enough to achieve zero training error, one needs a suitable normalization, which is discussed in the next paragraph.

\subsection{Geometric Margin as a Criterion to Compare Networks Achieving Zero Training Error}
\label{sec:normalized losses}
As discussed above, the (cross-entropy) loss itself is not a suitable criterion to distinguish networks that achieve zero training error.
A proper criterion should eliminate the degree of freedom for upscaling weights  \cite{farhang2022investigating}. Let 
\[ g(W,x):=f_{1}(W,x)-f_{-1}(W,x),\]
and denote with $L(g(W))$ its Lipschitz constant for the $\ell_2$-distance with respect to the input $x$. Then, the criterion
\begin{equation}\label{eq:lipschitznorm} 
    \rho(x) \triangleq \frac{g(W,x)}{L(g(W))},
\end{equation}
eliminates the scaling degree of freedom in the weights as the Lipschitz constant $L(g(W))$ has the same homogeneity as $g(W,x)$ with respect to the weights. Furthermore, as shown below, $|\rho(x)|$ provides a lower bound on the distance to the decision boundary. Let $z$ be on the decision boundary, that is $g(W,z)=0$. Then,
\begin{align*}
 |g(W,x)|=|g(W,x)-g(W,z)| \leq L(g(W)) \norm{x-z}_2,
\end{align*}
thus $|\rho(x)|$ is a lower bound on the $\ell_2$-distance of $x$ to the decision boundary of $g$. 
However, it is difficult to determine the Lipschitz constant of a neural network. 

As a ReLU-network is piecewise linear as a function of $x$, it holds
\[ L(g(W))=\max_{x \in \R^d} \norm{\nabla_x g(W,x)}_2.\]
We determine a lower bound on $L(g(W))$ by taking the maximum over the union of train and test data, i.e.,
\begin{equation}
\label{eq:lipschitznorm estimate} 
\tilde{\rho}(x)=\frac{g(W,x)}
{\max_{z \in \ST \cup \mathcal{S}_\text{test}}  \norm{\nabla_x g(W,z)}_2}.
\end{equation}
We then plug $\tilde{\rho}(x)$ into the loss instead of $g(W,x)$ and term it \textbf{Lipschitz normalized loss}. We note that by replacing the true Lipschitz constant with a lower bound, we cannot argue anymore that $|\tilde{\rho}(x)|$ is a lower bound of the distance to the decision boundary. However, this data-based estimate is a tighter estimate of $L(g(W))$ than a direct upper bound that uses the $1$-Lipschitzness of the ReLU activation, 
\[ L(g(W))\leq \prod_{l=1}^L \norm{W_l}_{2,2},\]
where $\norm{W_l}_{2,2}$ denotes the spectral norm of the weight matrix. In fact, the normalization of  \citet{chiang2022loss} by the product of Frobenius norms,
\begin{equation}\label{eq:weightnorm}
\gamma(x)=\frac{g(W,x)}{ \prod_{l=1}^L \norm{W_l}_{F}},
\end{equation}
is even a coarser upper bound. To show that, let $\sigma_i(W)$ be the singular values of $W$, then $\norm{W}_F=\sqrt{\sum_{i} \sigma^2_i(W)}\geq \max_i \sigma_i(W)=\norm{W}_{2,2}$. Note that this gap grows with the rank of $W$. In their analysis, $\gamma(x)$ is plugged into the loss, which we term \textbf{weight normalized loss}. While their normalized loss is fine when comparing neural networks of the \emph{same} architecture, it is problematic when comparing \emph{different} architectures with different ranks of $W$, e.g., networks of different widths.   
In Section~\ref{sec:width}, we show
that this ambiguity in the definition of the normalized loss can change the interpretation of the results of \citet{chiang2022loss}. While we believe the \textbf{Lipschitz normalized loss} is a more accurate estimate, we base our main observations on quantities independent of this normalization
to avoid the resulting ambiguities. If not specified differently in the paper, ``normalized loss'' always denotes our Lipschitz normalized loss. 

\setlength\cellspacetoplimit{0pt}
\setlength\cellspacebottomlimit{0pt}
\renewcommand\tabularxcolumn[1]{>{\centering\arraybackslash}S{p{#1}}}

\begin{figure*}[htp]
  \setlength\tabcolsep{0pt}
  \adjustboxset{width=\linewidth,valign=c}
  \centering
  \begin{tabularx}{1.0\linewidth}{
  @{}l 
  S{p{0.05\textwidth}} 
  *{4}{S{p{0.2245\textwidth}}} 
  S{p{0.052\textwidth}}}
    %%%%%%%%%%%%%%%%%%%%%
    &
    & \multicolumn{1}{c}{\textbf{\quad \enspace Uniform [-1, 1]}}
    & \multicolumn{1}{c}{\textbf{\quad \enspace Uniform [-0.2, 0.2]}}
    & \multicolumn{1}{c}{\textbf{\quad \enspace Kaiming Uniform}}
    & \multicolumn{1}{c}{\textbf{\quad \enspace Kaiming Gaussian}}
    & \multicolumn{1}{c}{} \\
    %%%%%%%%%%%%%%%%%%%%%%
    % SGD 2 
    %%%%%%%%%%%%%%%%%%%%%%%%
    &\rotatebox[origin=c]{90}{\textbf{\makecell{\enspace\sgd{}\\ \enspace Test accuracy}}}
    &\includegraphics{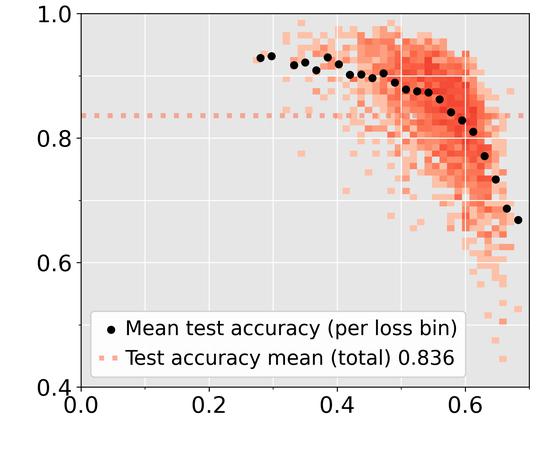}
    & \includegraphics{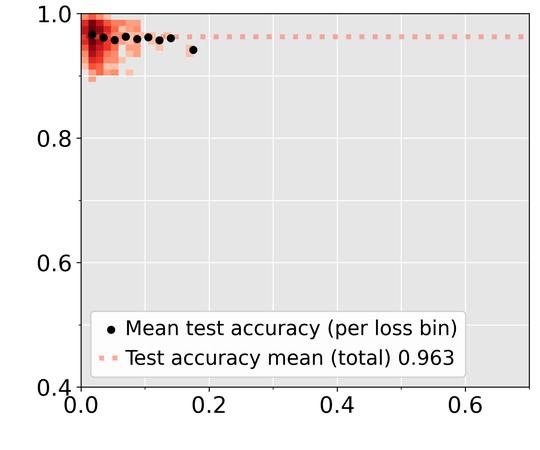}
    & \includegraphics{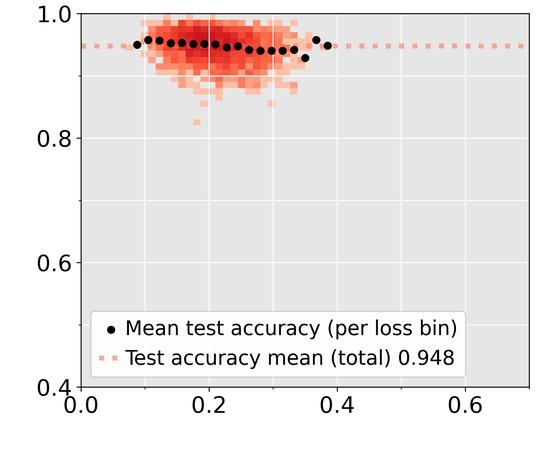}
    & \includegraphics{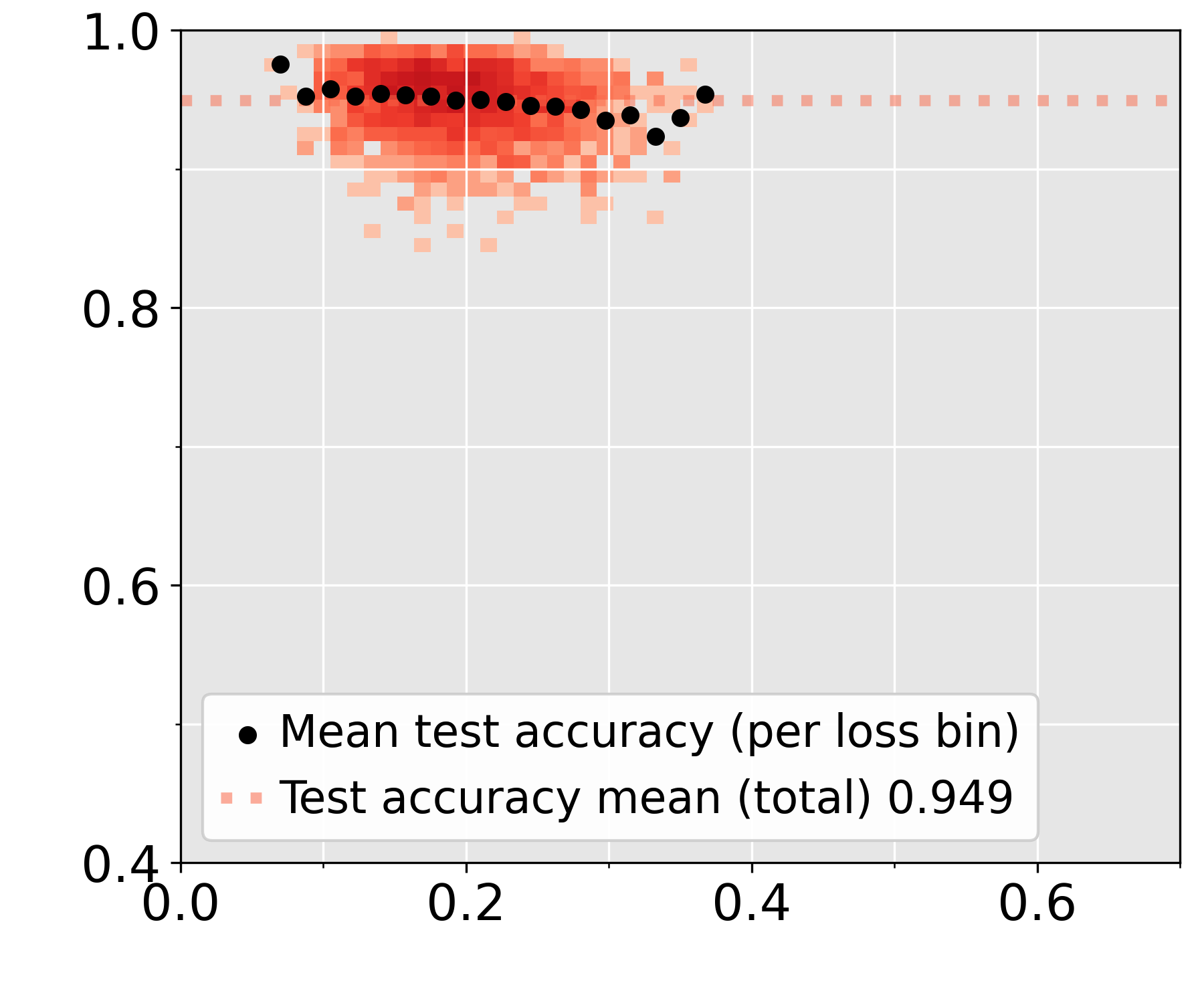}
    & 
    \includegraphics{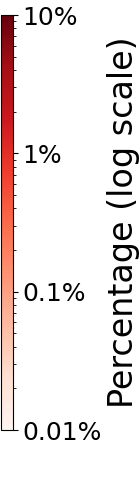} \\
    %%%%%%%%%%%%%%%%%%%%%%
    % GNC 2
    %%%%%%%%%%%%%%%%%%%%%%%%
    &\rotatebox[origin=c]{90}{\textbf{\makecell{\enspace \gnc{}\\ \enspace Test accuracy}}}
    & \includegraphics{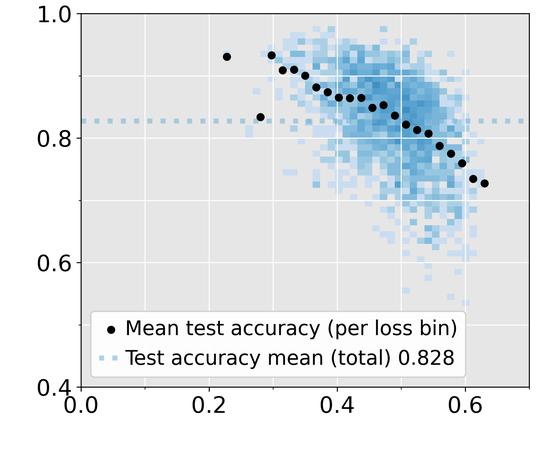}
    & \includegraphics{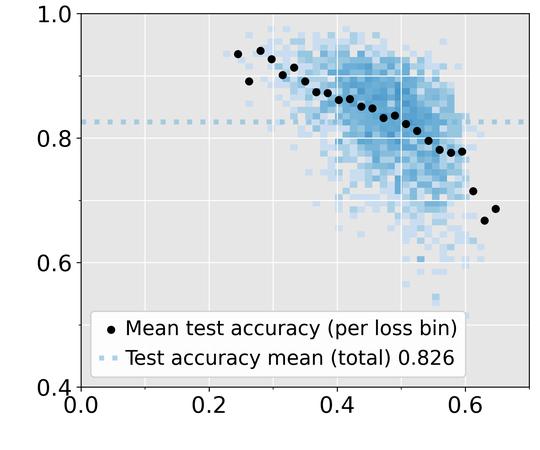}
    & \includegraphics{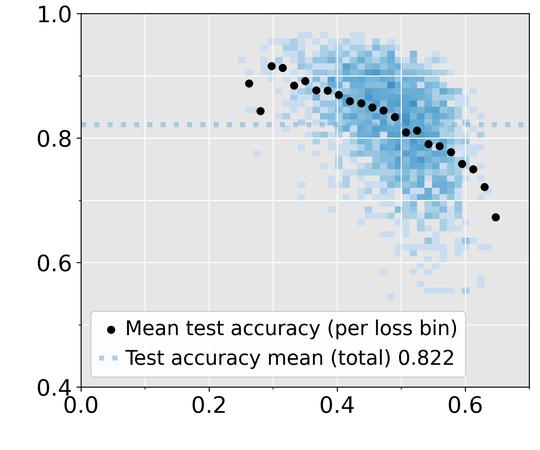}
    & \includegraphics{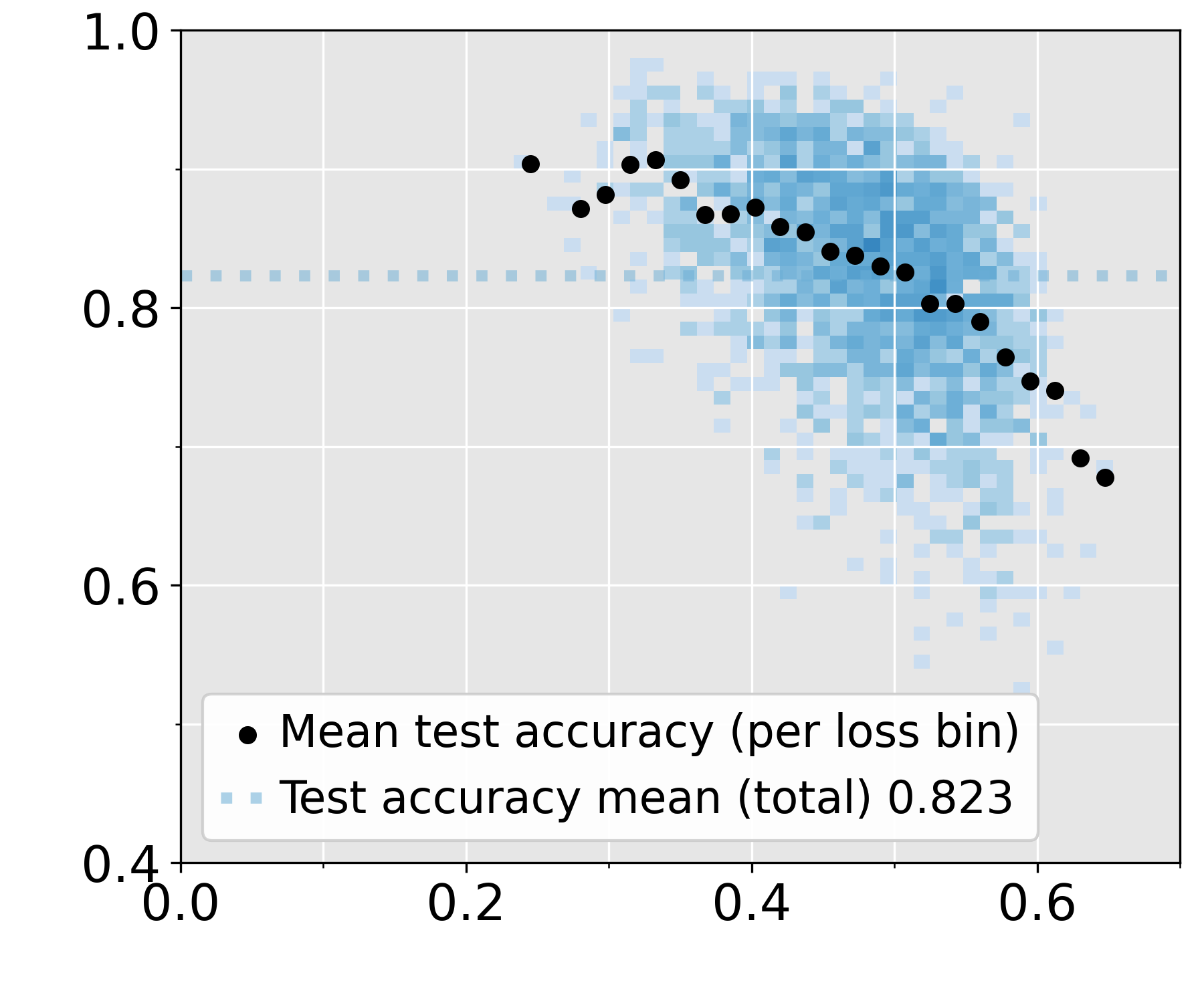}
    & \includegraphics{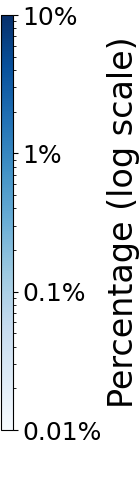} \\
    %%%%%%%%%%%%%%%%%%%%%%%%%%%%%%%%%%%%%%%%%%%%%%%%%%%%%%%%%
    % x axis labels
    %%%%%%%%%%%%%%%%%%%%%%%%%%%%%%%%%%%%%%%%%%%%%%%%%%%%%%%%%
    &
    & \multicolumn{1}{c}{\hspace{0.2em} \textbf{Train loss (normalized)}}
    & \multicolumn{1}{c}{\hspace{0.5em} \textbf{Train loss (normalized)}}
    & \multicolumn{1}{c}{\hspace{0.15em} \textbf{\enspace Train loss (normalized)}}
    & \multicolumn{1}{c}{\hspace{0.25em} \textbf{\enspace Train loss (normalized)}}
    &
  \end{tabularx}
  \caption{\textbf{Generalization of \colorsgd{\sgd{} (optimized)} versus \colorgnc{\gnc{} (randomly sampled)} in dependency of the prior on the weights $\Pr(W)$:} 
    We ``train'' $2000$ LeNet models to $100\%$ train accuracy for $16$ training samples from classes \emph{0} and \emph{7} of MNIST.
    Test accuracies for \gnc{} are similar across initializations, and the normalized loss (see Section~\ref{sec:method}) is similar across the uniform distributions.
  \textbf{Column 1:} For $\mathcal{U}[-1, 1]$ initialization, as used by \citet{chiang2022loss}, the normalized losses and the test accuracies of \sgd{} and \gnc{} are similar, except for the convergence of \sgd{} towards more low-margin solutions. 
  The claim in \citet{chiang2022loss} that the average test accuracy of \gnc{} resembles \sgd{}, conditional on the normalized loss bin (black dots), is an artifact of the suboptimal convergence of \sgd{} caused by this initialization.
  \textbf{Columns 2-4:} For other initializations, \sgd{} (first row) improves considerably both in terms of loss and accuracy.
  In contrast, \gnc{} remains unaffected, as it is independent of the scale of the weights in each layer.   
  Results for different numbers of samples and other classes from MNIST and CIFAR10 are in Appendix~\ref{sec:init-appendix}.
}
  \label{fig:different_initializations}
  \vspace{-0.1cm}
\end{figure*}

\subsection{Training of SGD and Guess and Check (\gnc)}
We deliberately avoid 
commonly used 
techniques for improving generalization, such as augmentation, weight decay, and sophisticated learning rate schedulers \cite{ruder2016overview}. 
Instead, we train \sgd{} using cross-entropy loss for $60$ epochs and a batch size of $2$, with a fixed learning rate.
We run \sgd{} from random initialization (see Section~\ref{sec:init}) and only keep networks that achieve zero training error.
Due to the strong overparameterization and our small number of training samples, zero training error is attained for almost every initialization.

The main analysis tool of this work is randomly sampling networks from a prior on the weights and only accepting networks that achieve zero training error. This is called Guess and Check algorithm (\gnc{}) in \citet{chiang2022loss}, and we keep this name for simplicity. 
We discuss the effects of different priors in Section~\ref{sec:init}.  

The advantage of random sampling networks from a prior  $P(w)$ is that we can directly estimate the probability of achieving zero training error for a given training set $\ST=(x_i,y_i)_{i=1}^n$. Given that the prior distribution has compact support, e.g., the uniform distribution, this is proportional to the  ``volume'' in the weight space of perfectly fitting solutions. Let $M_{\ST}$ be the number of sampled networks needed to get $N$ networks that have zero training error, then we can estimate this probability as
\[ \Pr_{W \sim P(W)}(\TrainError(W)=0\,|\,\ST) \approx \frac{N}{M_{\ST}}.\]
This probability indicates how difficult it is to fit the data perfectly. A random function $f:\R^d \rightarrow \{-1,1\}$ yields probability $\Pr_{f}(\TrainError(W)=0\,|\,\ST)=\frac{1}{2^n}$ for $n$ training samples. 
This baseline is indicated with a black dotted line when we plot the 
probability to
find a perfectly fitting network vs the training set size in Figures~\ref{fig:different_widths} and \ref{fig:different_depths}. Note that random networks can be biased towards constant predictions and thus be worse than a random function for a small number of training samples.
\section{Experiments}
\label{sec:experiments}
In the case of overparameterization, \sgd{} typically converges to a solution with zero training error. It is also a common belief that \sgd{} is biased towards networks that generalize better. 
Thus, we clearly expect, even in a low sample regime,  that \sgd{} finds, on average, better generalizing networks than \gnc{}. 
It is, however, unclear to which extent the inductive bias of the function class, which is implemented by the neural network architecture, contributes to better generalization. 
By increasing the overparameterization of the networks and comparing the average test error over networks that have achieved zero training error, we can disentangle the effects of the architecture and the optimization. 
Specifically, we overparametrized the networks by increasing width or depth.
Then, when \sgd{} and \gnc{} behave similarly, the change can be purely attributed to the architecture.
In contrast, when they behave differently, the optimization process has a clear additional bias. 

As all networks found by \gnc{} have zero training error, their average test error serves as an estimator for the expected generalization ability of random networks that achieve zero training error, which is
\[ \Exp[\TestError(W) \,|\,\TrainError(W)=0].\]
We use average test accuracy as the main quantity to compare \sgd{} and \gnc{}, as it remains unaffected by potential issues related to weight normalization.
In the remainder of this section, we explore in Section~\ref{sec:init} the effect of initialization of the network (different priors for the weights) and the behavior when increasing the width and depth of the network (Section~\ref{sec:width} and Section~\ref{sec:depth}, respectively). 

\subsection{Dependency of \sgd{} and \gnc{} on the Prior $P(W)$}
\label{sec:init}
Networks randomly sampled from $P(W)$ correspond to the initialization used in \sgd{} and are the ones checked for zero training error in \gnc{}. We note that \gnc{} is independent of the scale of the weights of each layer as the ReLU-network is one-homogeneous in every layer, and thus, it only changes the scale of the logits but not the classification. This is different for \sgd{}, where it is well-known that the scale of each layer's initialization significantly affects the training dynamics. This is why, in practice, the Kaiming initialization, also known as He initialization \cite{he2015delving}, is often used, in which the weights of each layer are sampled from a uniform or Gaussian distribution and then normalized. 

\citet{chiang2022loss} find that conditioned on a specific bin of the weight normalized loss, \gnc{} and \sgd{} have similar test errors.
From this, they conclude that the main effect of generalization in the overparameterized setting is not due to a bias of \sgd{} but is inherent in the network architecture. 
However, they do not apply the standard Kaiming initialization but sample each weight uniformly at random in $[-1,1]$. 
This leads to large logits and low loss once zero training error is achieved, resulting in small gradients. Consequently, \sgd{} essentially stops the optimization early in the training process, as illustrated in Figure~\ref{fig:initialization_gradients} in the appendix,
leaving the margin to the decision boundary small. 
Thus, they effectively compare
a suboptimal early-stopped
\sgd{}-optimized network with a randomly sampled network, which leads to their impression that \gnc{} and \sgd{} behave similarly.

\setlength\cellspacetoplimit{0pt}
\setlength\cellspacebottomlimit{0pt}
\renewcommand\tabularxcolumn[1]{>{\centering\arraybackslash}S{p{#1}}}
\begin{figure*}[ht!]
\centering
\setlength\tabcolsep{0pt}
\adjustboxset{width=\linewidth,valign=c}
\centering
\begin{tabularx}{1.0\linewidth}{
@{}l 
S{p{0.022\textwidth}} 
S{p{0.047\textwidth}} 
*{4}{S{p{0.219\textwidth}}} 
S{p{0.051\textwidth}}}
    %%%%%%%%%%%%%%%%%%%%%%%%%
    &
    &
    & \multicolumn{1}{c}{\quad \textbf{Width} $\nicefrac{\mathbf{1}}{\mathbf{6^*}}$}
    & \multicolumn{1}{c}{\quad $\nicefrac{\mathbf{1}}{\mathbf{6}}$}
    & \multicolumn{1}{c}{\quad $\nicefrac{\mathbf{2}}{\mathbf{6}}$}
    &\multicolumn{1}{c}{\quad $\nicefrac{\mathbf{4}}{\mathbf{6}}$}
    &\\
    %%%%%%%%%%%%%%%%%%%%%%%%%%%%%%%%%%%
    % GNC 1
    %%%%%%%%%%%%%%%%%%%%%%%%%%%%%%%%%%%
    &\multirow{2}{*}[-4em]{\rotatebox{90}{\textbf{\gnc}}}
    &\rotatebox[origin=c]{90}
    {\footnotesize \quad \textbf{\makecell{Test accuracy vs\\ weight norm. loss}}}
    &\includegraphics{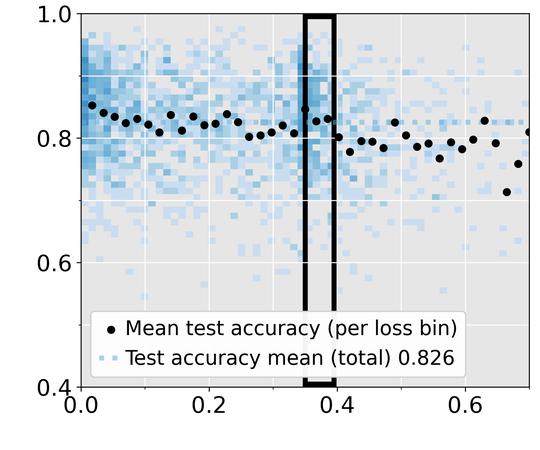}
    &\includegraphics{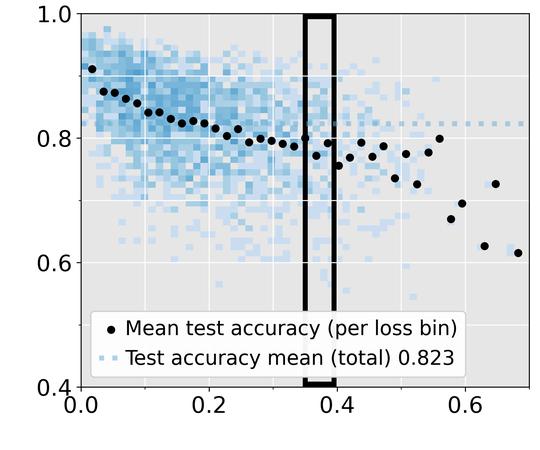}
    &\includegraphics{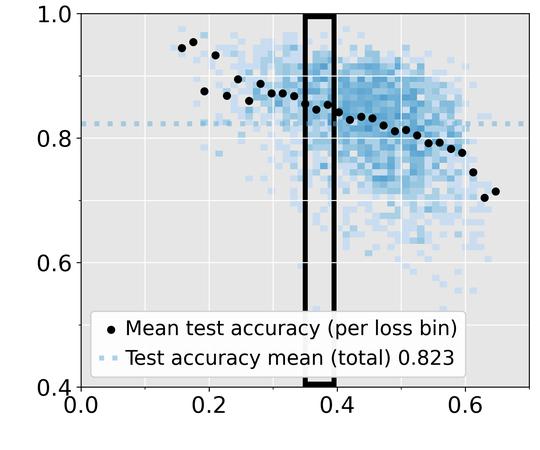}
    &\includegraphics{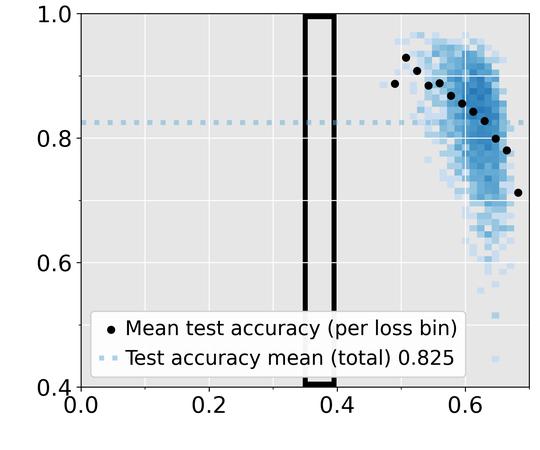}
    &\includegraphics{figures/different_loss_factor/colorbar.png} \\
     %%%%%%%%%%%%%%%%%%%%%%%%%%%%%%%%%%%
    % GNC 2
    %%%%%%%%%%%%%%%%%%%%%%%%%%%%%%%%%%%
    &
    &\rotatebox[origin=c]{90}
    {\footnotesize \enspace \textbf{\makecell{Test accuracy vs \\ Lipschitz norm. loss}}}
    &\includegraphics{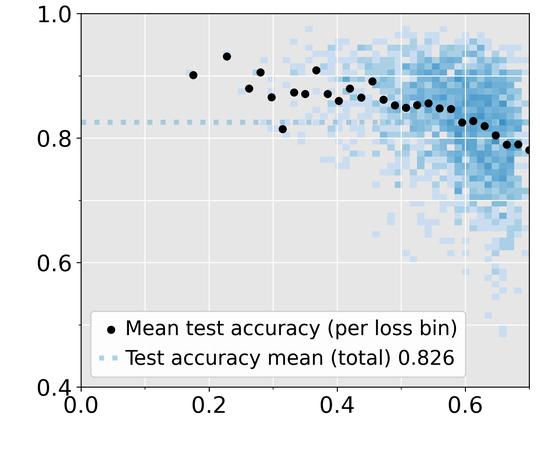}
    &\includegraphics {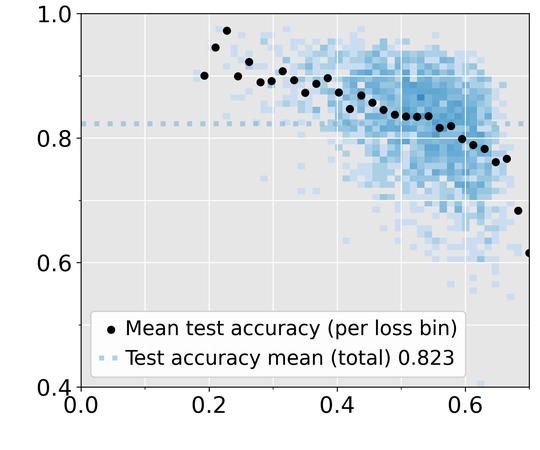}
    &\includegraphics{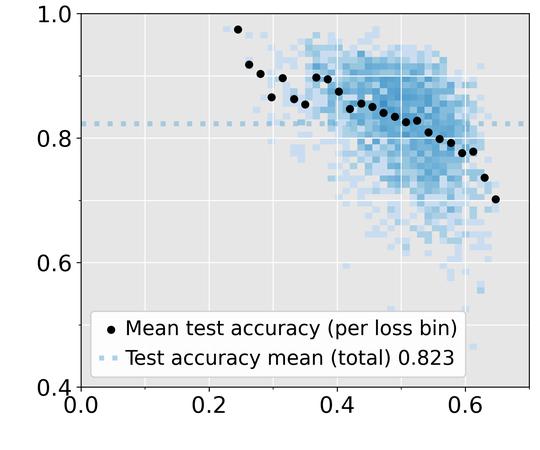}
    &\includegraphics{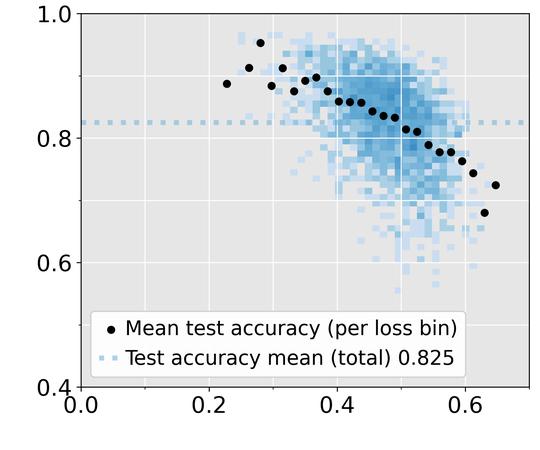}
    &\includegraphics{figures/different_loss_factor/colorbar.png} \\
    %%%%%%%%%%%%%%%%%%%%%%%%%%%%%%%%%%%
    % SGD 1
    %%%%%%%%%%%%%%%%%%%%%%%%%%%%%%%%%%%
    &\multirow{2}{*}[-4em]{\rotatebox{90}{\textbf{\sgd}}}
    &\rotatebox[origin=c]{90}    {\footnotesize \quad \textbf{\makecell{Test accuracy vs\\ weight norm. loss}}}
    & \includegraphics{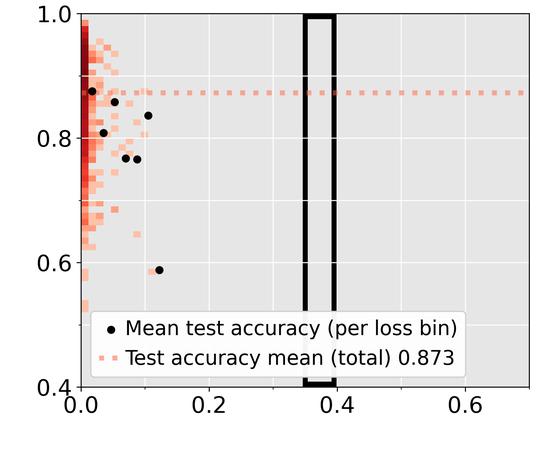}
    &\includegraphics{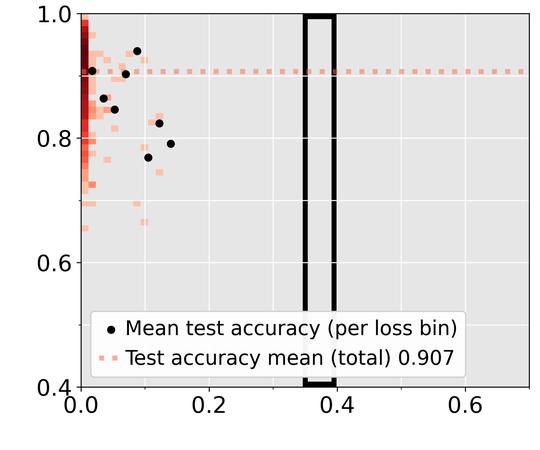}
    &\includegraphics{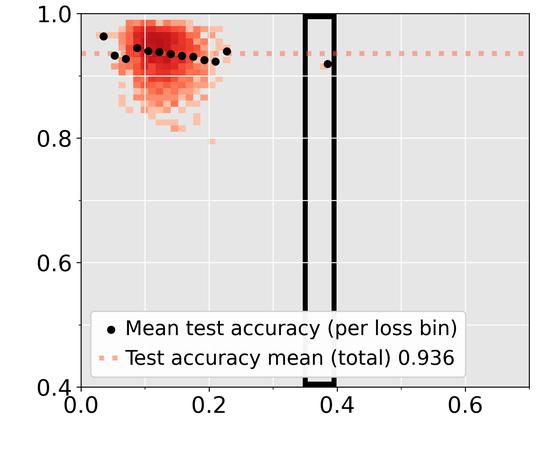}
    &\includegraphics{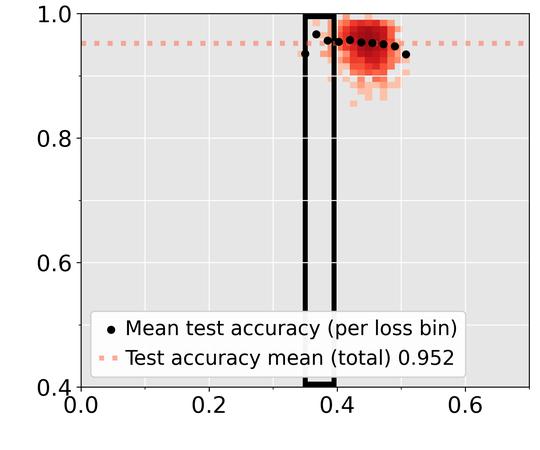}
    &\includegraphics{figures/different_loss_factor/colorbar_SGD.png} \\
    %%%%%%%%%%%%%%%%%%%%%%%%%%%%%%%%%%%
    % SGD 2
    %%%%%%%%%%%%%%%%%%%%%%%%%%%%%%%%%%%
    &
    &\rotatebox[origin=c]{90}    {\footnotesize  \enspace \textbf{\makecell{Test accuracy vs \\ Lipschitz norm. loss}}}
    &\includegraphics{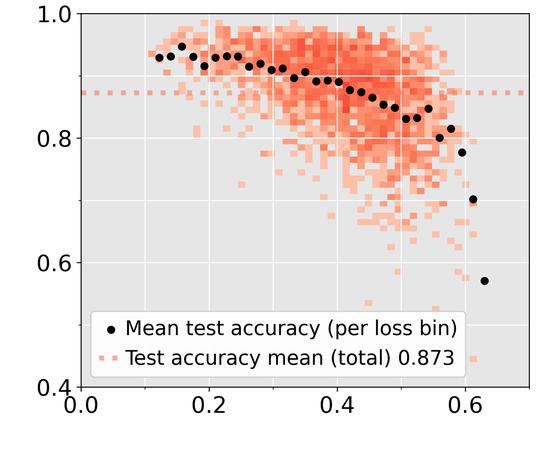}
    &\includegraphics{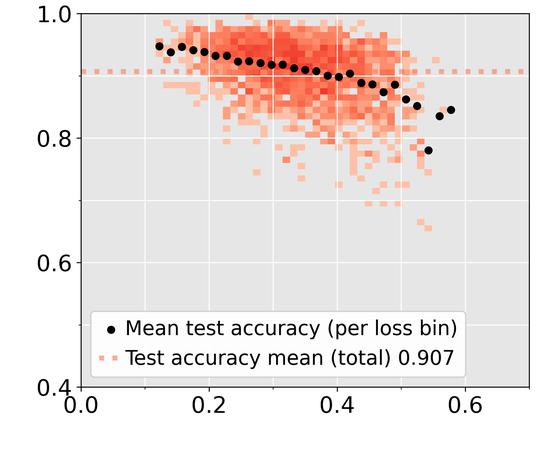}
    &\includegraphics{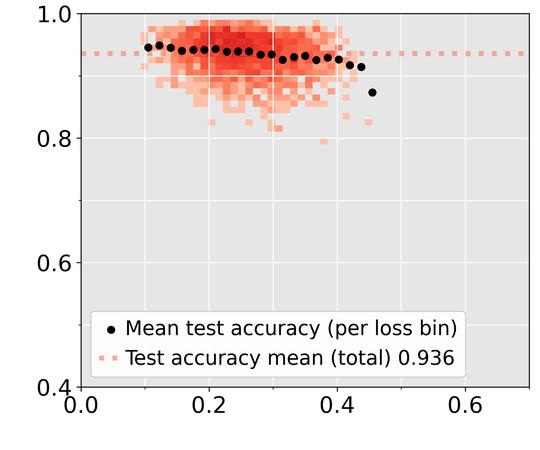}
    &\includegraphics{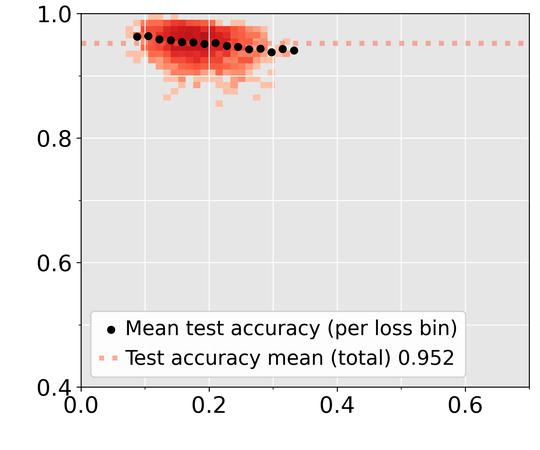}
    &\includegraphics{figures/different_loss_factor/colorbar_SGD.png} \\
    %%%%%%%%%%%%%%%%%%%%%%%%%%%%%%%%%
    &
    &
    & \multicolumn{1}{c}{\footnotesize \hspace{0.7em} \textbf{Train loss (normalized)}}
    & \multicolumn{1}{c}{\footnotesize \hspace{0.7em} \textbf{Train loss (normalized)}}
    & \multicolumn{1}{c}{\footnotesize \hspace{0.25em} \textbf{\enspace Train loss (normalized)}}
    & \multicolumn{1}{c}{\footnotesize \hspace{0.5em} \textbf{\enspace Train loss (normalized)}}
    &  
    \end{tabularx}
    \caption{
    \textbf{Analysis of overparameterization when increasing the width.}
    Test accuracy vs weight normalized loss \eqref{eq:weightnorm} of \citet{chiang2022loss} and our Lipschitz normalized loss \eqref{eq:lipschitznorm estimate} of {\color{amaranth}{\textbf{\sgd}}} and {\color{azure}{\textbf{\gnc}}} for classes \emph{0} \& \emph{7} of MNIST and 16 training samples across 2000 LeNet models. 
    \textbf{Row 4:} Widening the networks enhances 
    geometric margin (lower normalized loss) and average test accuracy for \sgd{}, while for \gnc{} \textbf{(Row 2)}, the margin improves only slightly, and average test accuracy remains the same. This suggests that the improvement is mainly due to the bias of \sgd{} and not due to an architectural bias (see Figure~\ref{fig:different_widths}). \textbf{Rows 1 and 3:} \citet{chiang2022loss} compare networks conditional on the (weight) normalized loss bin (illustrated by black boxes), which led them to conclude that \gnc{} improves with increasing width.
    With our Lipschitz normalized loss, one would arrive at the opposite conclusion, which shows the problem of normalization. Results for different numbers of samples and other classes from MNIST and CIFAR10  are in Appendix~\ref{sec:width-appendix}.
    }
    \label{fig:different_widths_loss_with_nornalization}
\end{figure*}

In Figure~\ref{fig:different_initializations}, we plot the test accuracy vs (Lipschitz) normalized loss for \sgd{} and \gnc{} using three different uniform initializations (their Uniform $[-1,1]$, Uniform $[-0.2,0.2]$, Kaiming Uniform) and the Kaiming Gaussian initialization. 
For each algorithm, we ``trained" $2000$ models to achieve $100\%$ train accuracy on a fixed subset of 16 samples from classes \emph{0} and \emph{7} from the MNIST dataset (similar results for different numbers of training samples and other classes from MNIST and CIFAR10 can be found in Appendix~\ref{sec:init-appendix}).

As argued in Section~\ref{sec:norm}, 
one observes 
that  
\gnc{} is not affected by the scale of the uniform initialization, whereas the outcome of \sgd{} depends heavily on it.
For the Uniform $[-1,1]$ used in \citet{chiang2022loss}, the average test accuracy is indeed almost the same for \sgd{} and \gnc{} ($83.6\%$ vs. 82.8\%).
However, with proper initializations, the results
of \sgd{} improve drastically to 96.3\%.
Additionally, the plot demonstrates that it is essentially impossible to compare \sgd{} and \gnc{} conditional on loss bins, as they achieve significantly different normalized losses. 
We stress that these findings are independent of which normalized loss is used. In Figure~\ref{fig:different_initializations_weight_normalization},
we show the same comparison as Figure~\ref{fig:different_initializations}, but with the weight normalized loss of \citet{chiang2022loss}.

In addition to Figure~\ref{fig:initialization_gradients}, which illustrates 
the 
problems with suboptimal initialization, Figure~\ref{fig:sgd_epochs_convergence} in the appendix shows the trajectory of \sgd{} with good initialization (Kaiming uniform) as one varies the number of training epochs.
The distribution of normalized losses of \sgd{} after one epoch (i.e., not fully optimized networks) is comparable to that of \gnc{} and \sgd{} with suboptimal initialization. However, even in this case, conditional on the bin of the normalized loss, the average test accuracy of \sgd{} is better than \gnc{}.

Therefore, for the remainder of the paper, we use 
Kaiming initialization for \sgd{} to get properly optimized networks.
However, we sample each layer from a uniform distribution so that our results of \gnc{} are directly comparable to the ones of \citet{chiang2022loss}, and since Kaiming Gaussian did not result in major differences both for \sgd{} and \gnc{}.

\setlength\cellspacetoplimit{0pt}
\setlength\cellspacebottomlimit{0pt}
\renewcommand\tabularxcolumn[1]{>{\centering\arraybackslash}S{p{#1}}}
\begin{figure}[htbp]
\centering
\setlength\tabcolsep{0pt}
\adjustboxset{width=\linewidth,valign=c}
\begin{tabularx}{1.0\linewidth}
{@{}l
S{p{0.03\textwidth}} 
S{p{0.02\textwidth}} 
*{2}{S{p{0.44\linewidth}}}}
    &
    &
    & \multicolumn{1}{c}{\footnotesize \textbf{\quad \enspace Mean test accuracy}}
    & \multicolumn{1}{c}
    {\footnotesize
    \textbf{\enspace \makecell{Probability to sample\\ 100\% train accuracy}}}\\
    %%%%%%%%%%%%%%%% Lenet %%%%%%%%%%%
    &\multirow{2}{*}[-1.5em]{\rotatebox{90}{\textbf{LeNet}}}
    &\rotatebox[origin=c]{90}{\textbf{\quad 0 vs 7}}
    &\includegraphics[width=\linewidth]{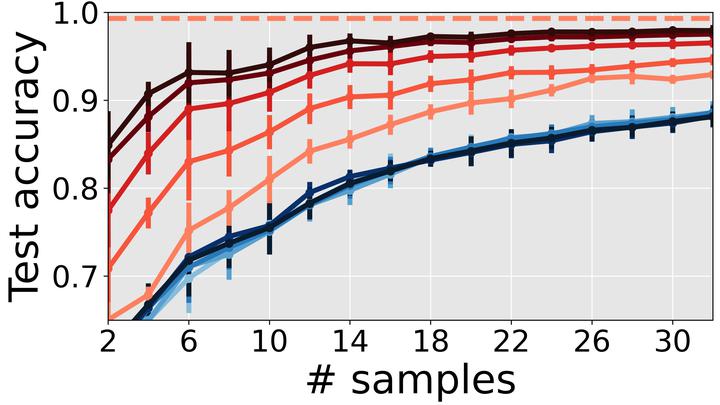}
    &\includegraphics[width=\linewidth]{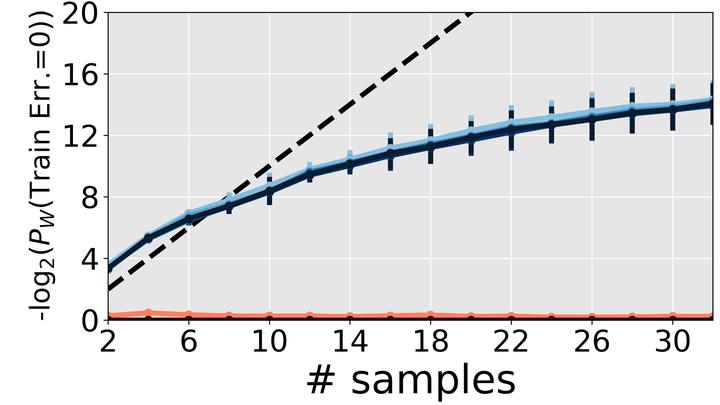}\\
    &&\rotatebox[origin=c]{90}{\textbf{\quad Bird vs Ship}}
    &\includegraphics[width=\linewidth]{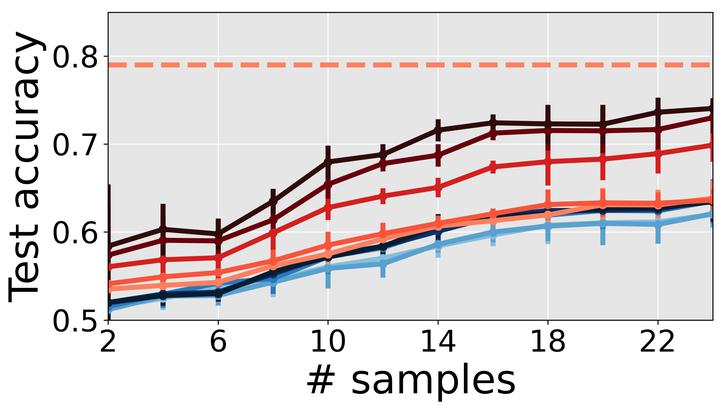}
    &\includegraphics[width=\linewidth]{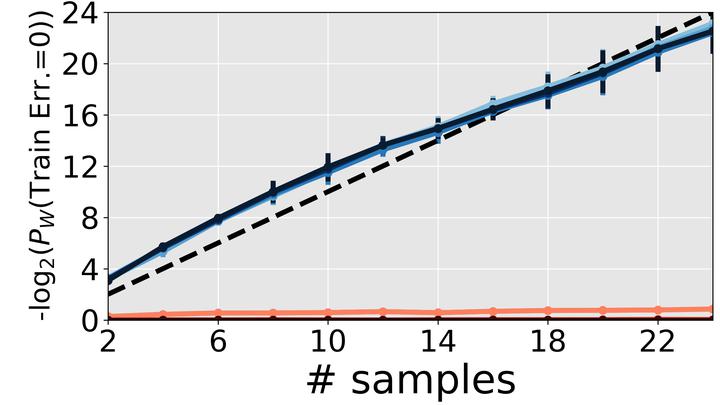}\\
    &\multicolumn{4}{c}{\includegraphics[width=\linewidth]{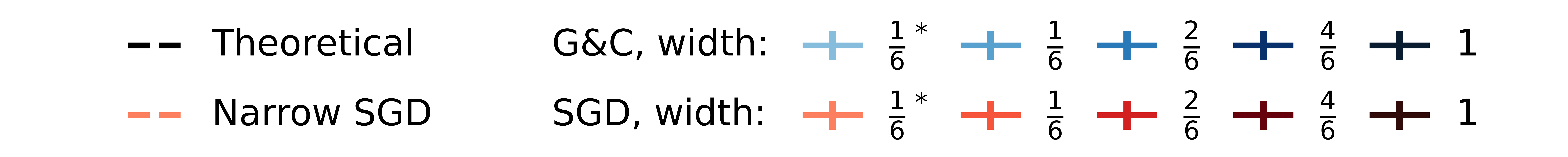}}\\
    \noalign{{\color{gray}\hrule height 1.5pt}} \\
    % %%%%%%%%%%% mlp %%%%%%%%%%%
    &\multirow{2}{*}[-1.5em]{\rotatebox{90}{\textbf{MLP}}}
    &\rotatebox[origin=c]{90}{\textbf{\quad 0 vs 7}}
    &\includegraphics[width=\linewidth]{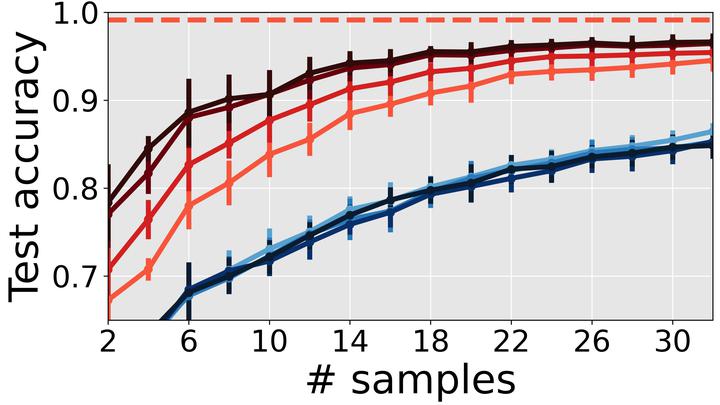}
    &\includegraphics[width=\linewidth]{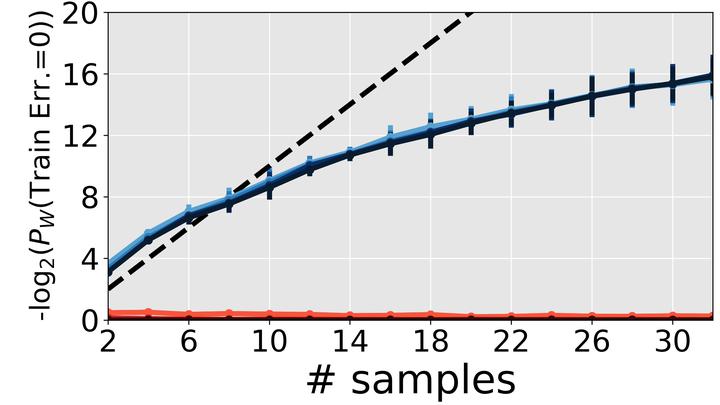}\\
    &&\rotatebox[origin=c]{90}{\textbf{\quad Bird vs Ship}}
    &\includegraphics[width=\linewidth]{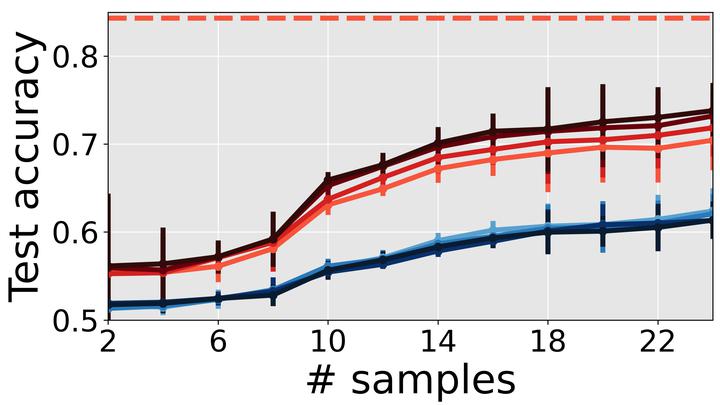}
    &\includegraphics[width=\linewidth]{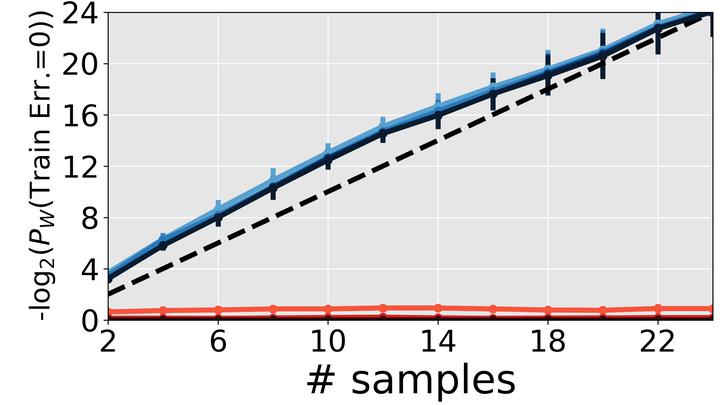}\\
    &\multicolumn{4}{c}{\includegraphics[width=\linewidth]{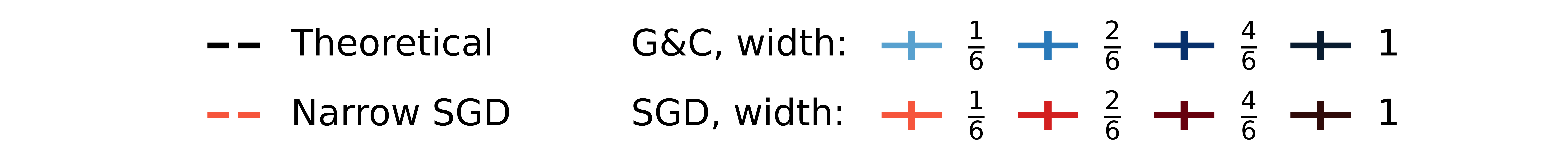}}\\
    \noalign{{\color{gray}\hrule height 1.5pt}} \\
    %%%%%%%%%%% resnet %%%%%%%%%%%
    &\multirow{2}{*}[-1.5em]{\rotatebox{90}{\textbf{ResNet}}}
    &\rotatebox[origin=c]{90}{\textbf{\quad 0 vs 7}}
    &\includegraphics[width=\linewidth]{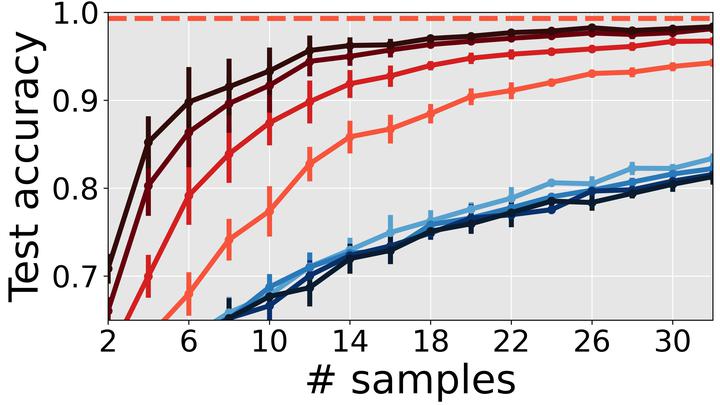}
    &\includegraphics[width=\linewidth]{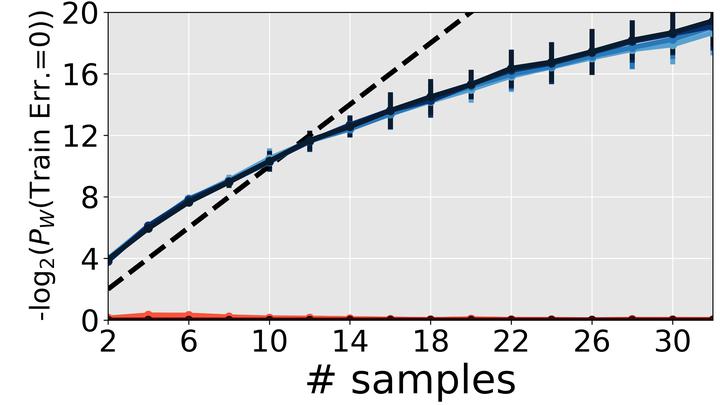}\\
    &&\rotatebox[origin=c]{90}{\textbf{\quad Bird vs Ship}}
    &\includegraphics[width=\linewidth]{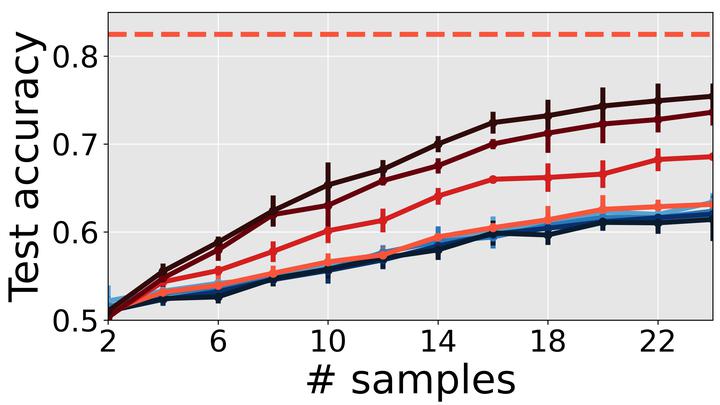}
    &\includegraphics[width=\linewidth]{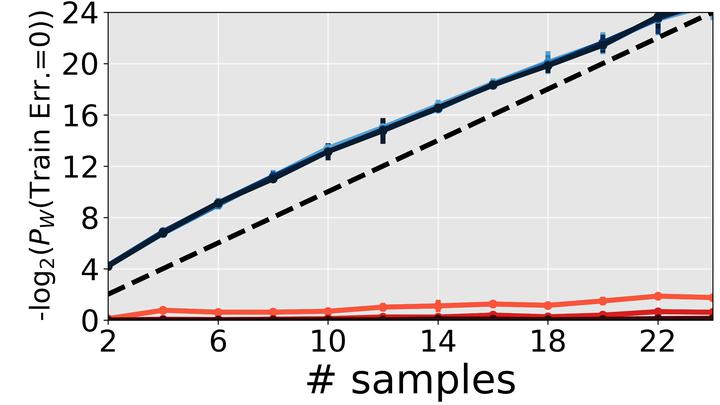}\\
    &\multicolumn{4}{c}{\includegraphics{figures/mlp/legend_widths_mlp.png}}\\
    \end{tabularx}
    \caption{
    \textbf{Increasing width is a positive optimization bias.} 
    From top to bottom, the architectures are LeNet, MLP, and ResNet. Within each architecture, the first row corresponds to MNIST and the second to CIFAR10.
    \textbf{Left:}  
    Mean test accuracy vs number of training samples 
    across network widths.
    \sgd{} improves for wider networks, while \gnc{} behaves similarly for all widths.  
    Thus, for increasing width, \sgd{} has a bias towards better generalizing networks independent of an architectural bias.
    However, there is also no overfitting for \gnc{} and thus no sign that overparameterization hurts. 
    \textbf{Right:} 
    We report the negative log probability of \gnc{} to find a network fitting the training data (${\Pr_W(\TrainError=0)}$).
    This number remains the same for different widths, indicating that the pool of ``fitting networks'' does not change with increasing width.  
    More class pairs of MNIST and CIFAR10 are provided in Appendix~\ref{sec:width-appendix}.
    }
    \label{fig:different_widths}
    \vspace{-0.2cm}
\end{figure}

The consistent convergence of properly initialized \sgd{} towards functions that generalize better than \gnc{} within the pool of networks fitting the training data shows that such well-generalizing functions occupy a small volume in the set of networks with zero training error. 
We explicitly test this by initializing \sgd{} with networks found by \gnc{}, that is, randomly sampled networks using Kaiming uniform, which fit the data perfectly.
The test accuracy of the converged \sgd{} versus the test accuracy of the initialization of \sgd{} found by \gnc{} is shown in Figure~\ref{fig:sgd_from_gnc} in the appendix. In almost all cases, \sgd{} improves compared to the \gnc{}-initialization,
which again demonstrates the positive bias of \sgd{} regarding generalization in the heavily overparameterized regime.
In addition, the normalized loss and test accuracy values for these optimized networks, depicted in Figure~\ref{fig:sgd_from_gnc_hist}, exhibit no significant difference from those associated with \sgd{} with the standard Kaiming initialization in Figure~\ref{fig:different_initializations}.
This suggests that initializing a network with a solution that already perfectly fits the data does not give any advantage to the convergence of \sgd{}.

To summarize, 
in this section, 
we examined the influence of the prior on \sgd{} and \gnc{}. 
We found that the performance of \sgd{} heavily depends on proper initialization, while \gnc{} is unaffected by the scale of the weights.
This contrasts \citet{chiang2022loss}, who reported similar behavior of \sgd{} and \gnc{}, using suboptimal initialization. Since proper initialization significantly improves \sgd{} results while keeping \gnc{} performance unaffected, we conclude that there exists a positive bias of \sgd{} for generalization in overparameterized regimes.

\setlength\cellspacetoplimit{0pt}
\setlength\cellspacebottomlimit{0pt}
\renewcommand\tabularxcolumn[1]{>{\centering\arraybackslash}S{p{#1}}}
\begin{figure}[htbp]
\centering
\setlength\tabcolsep{0pt}
\adjustboxset{width=\linewidth,valign=c}
\begin{tabularx}{1.0\linewidth}
{@{}l
S{p{0.03\textwidth}} 
S{p{0.02\textwidth}} 
*{2}{S{p{0.44\linewidth}}}}
    &
    &
    & \multicolumn{1}{c}{\footnotesize \textbf{\quad \enspace Mean test accuracy}}
    & \multicolumn{1}{c}
    {\footnotesize
    \textbf{\enspace \makecell{Probability to sample\\ 100\% train accuracy}}}\\
    %%%%%%%%%%%%%%%%%%%%%   Lenet   %%%%%%%%%%%%%%
    &\multirow{2}{*}[-1.5em]{\rotatebox{90}{\textbf{LeNet}}}
    &\rotatebox[origin=c]{90}{\textbf{\quad 0 vs 7}}
    &\includegraphics[width=\linewidth]{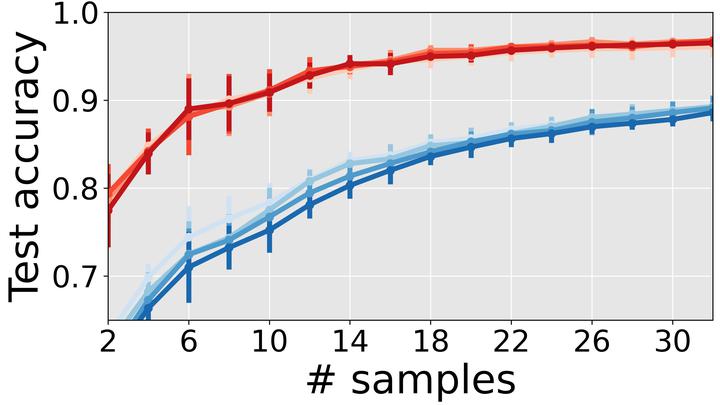}
    &\includegraphics[width=\linewidth]{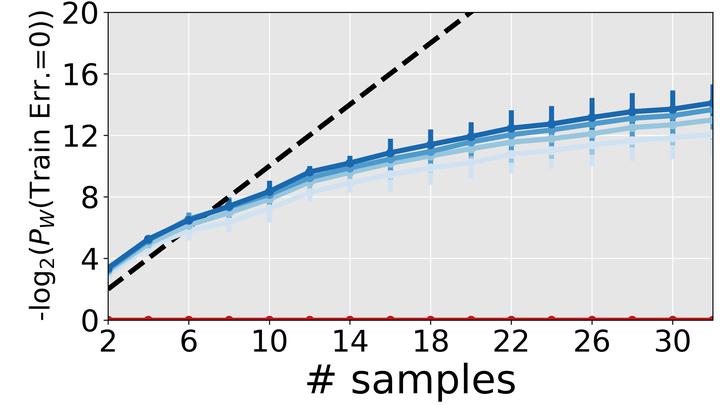}\\
    &&\rotatebox[origin=c]{90}{\textbf{\quad Bird vs Ship}}
    &\includegraphics[width=\linewidth]{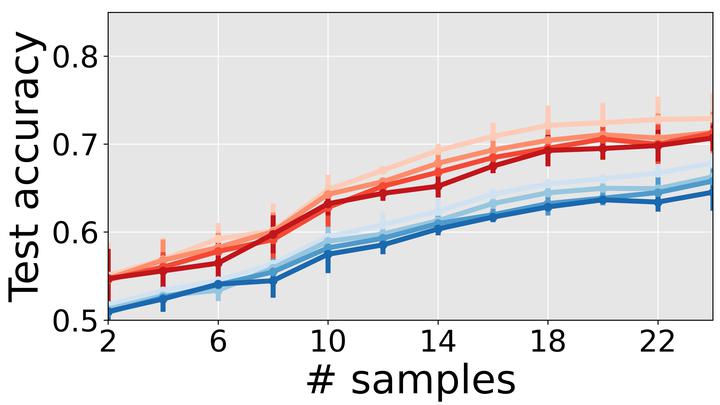}
    &\includegraphics[width=\linewidth]{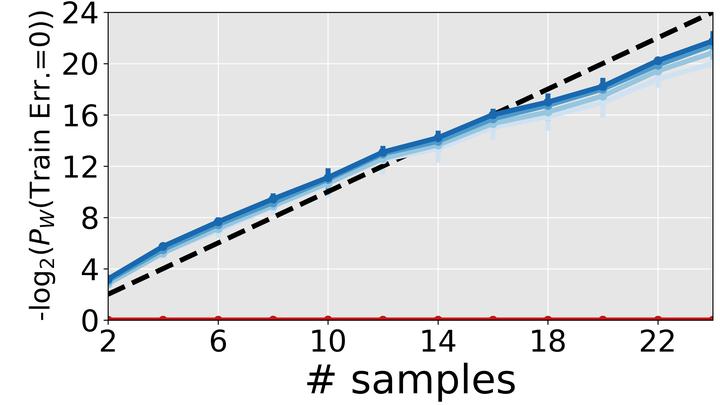}\\
    &\multicolumn{4}{c}{\includegraphics[width=\linewidth]{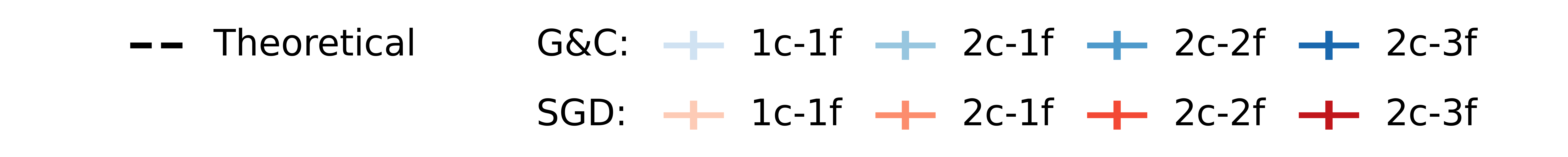}}\\
    \noalign{{\color{gray}\hrule height 1.5pt}} \\
    %%%%%%%%%%%%%%%%%%   mlp   %%%%%%%%%%%%%
    &\multirow{2}{*}[-1.5em]{\rotatebox{90}{\textbf{MLP}}}
    &\rotatebox[origin=c]{90}{\textbf{\quad 0 vs 7}}
    &\includegraphics[width=\linewidth]{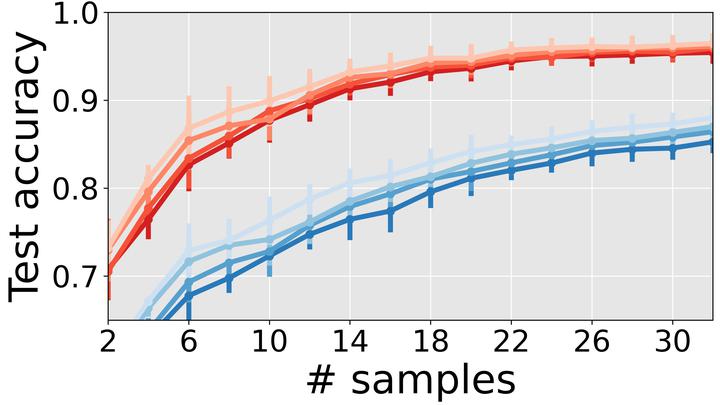}
    &\includegraphics[width=\linewidth]{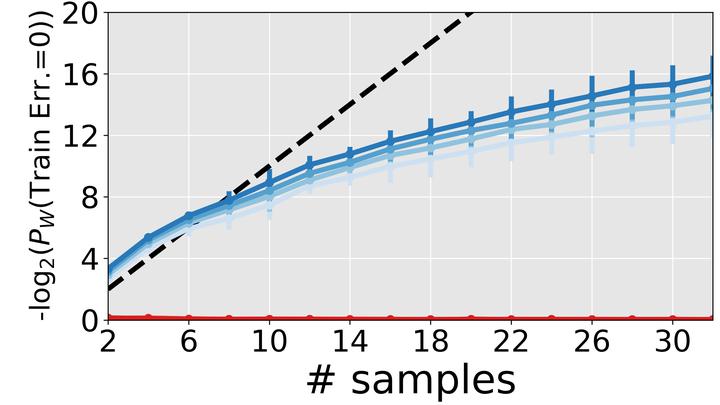}\\
    &&\rotatebox[origin=c]{90}{\textbf{\quad Bird vs Ship}}
    &\includegraphics[width=\linewidth]{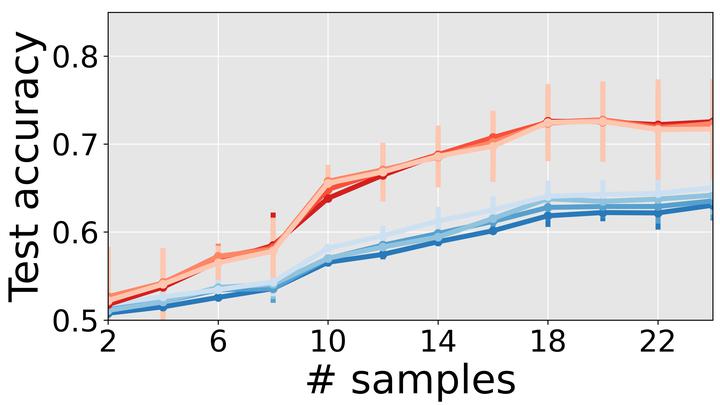}
    &\includegraphics[width=\linewidth]{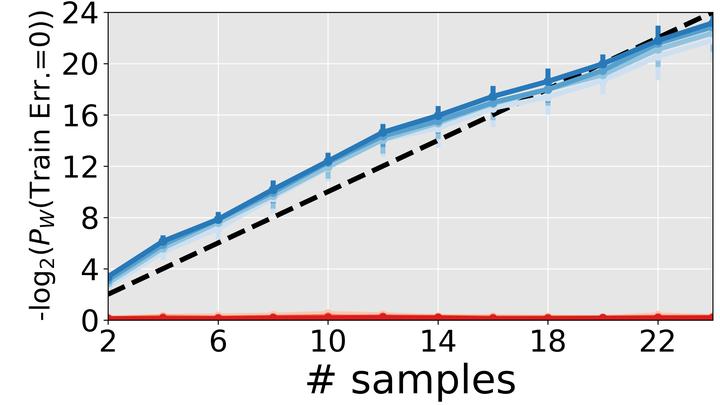}\\
    &\multicolumn{4}{c}{\includegraphics[width=\linewidth]{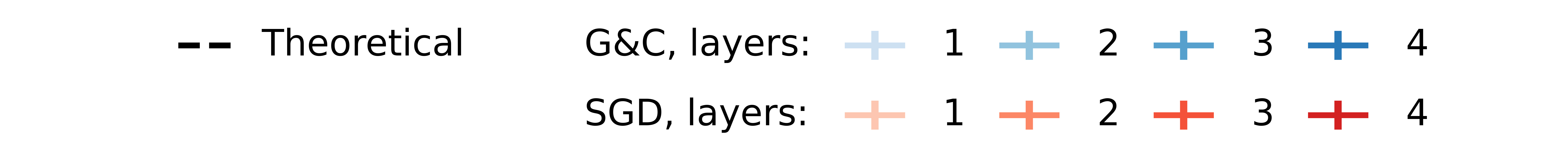}}\\
    \noalign{{\color{gray}\hrule height 1.5pt}} \\
    %%%%%%%%%%%%%%%%%%%%%%  resnet   %%%%%%%%%%%%%
    &\multirow{2}{*}[-1.5em]{\rotatebox{90}{\textbf{ResNet}}}
    &\rotatebox[origin=c]{90}{\textbf{\quad 0 vs 7}}
    &\includegraphics[width=\linewidth]{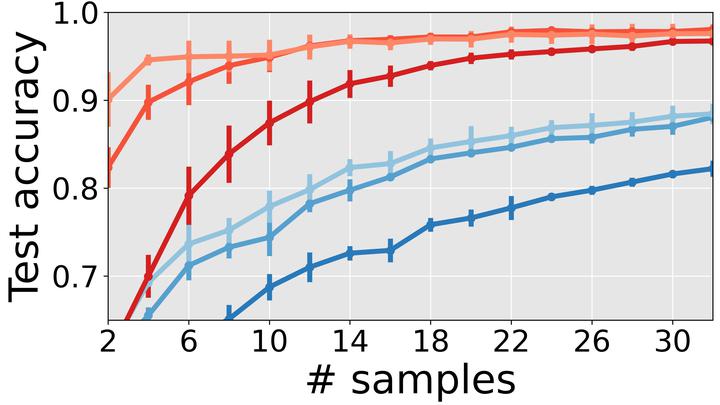}
    &\includegraphics[width=\linewidth]{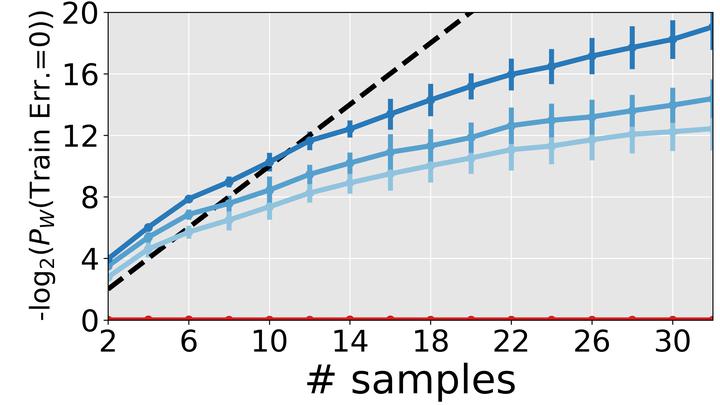}\\
    &&\rotatebox[origin=c]{90}{\textbf{\quad Bird vs Ship}}
    &\includegraphics[width=\linewidth]{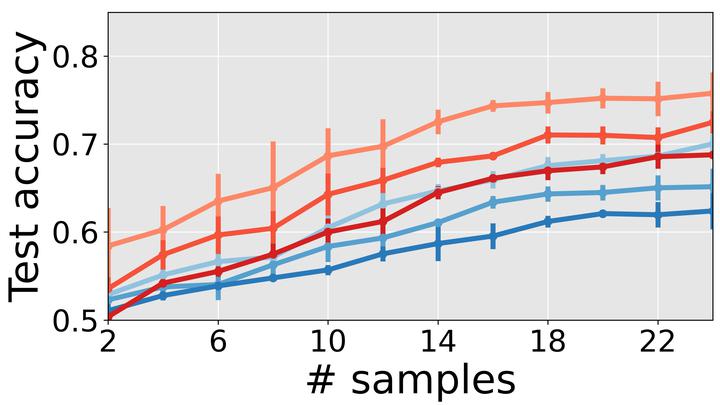}
    &\includegraphics[width=\linewidth]{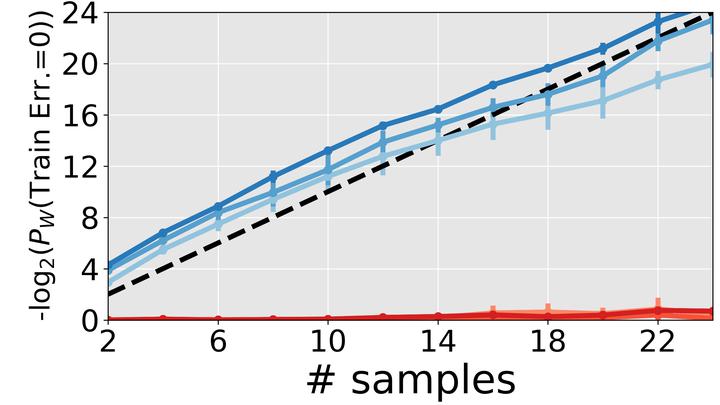}\\
    &\multicolumn{4}{c}{\includegraphics[width=\linewidth]{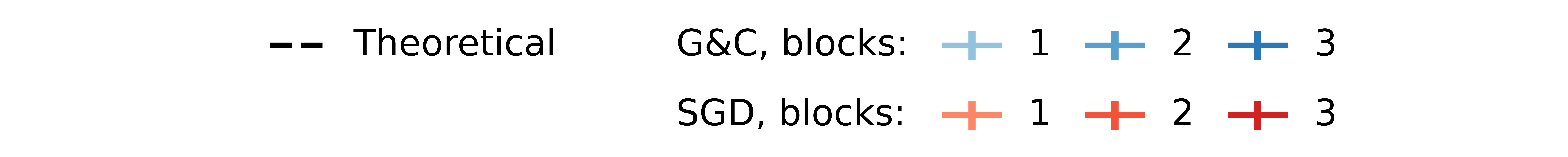}}\\
    \end{tabularx}
    \caption{\textbf{Increasing depth is a negative architectural bias.} 
    From top to bottom, the architectures are LeNet, MLP, and ResNet. Within each architecture, the first row corresponds to MNIST and the second to CIFAR10.
    Configuration ``2c-1f" means two convolutional layers followed by a fully connected layer.
    \textbf{Left:} Mean test accuracy vs number of training samples 
    across network depths.
    \gnc{} always performs worse as depth increases, whereas \sgd{} stagnates  
    or gets worse.
    Thus, overparameterization in terms of depth results in overfitting instead of better generalization, unlike for the width.
     Since both \gnc{} and \sgd{} follow a similar trend, the decrease in performance with increased depth can be attributed to architectural bias. 
    \textbf{Right:} Deeper networks 
    have a lower probability for \gnc{} to fit the training data, indicating that the network produces more complex functions.
    More class pairs of MNIST and CIFAR10 are provided in Appendix~\ref{sec:depth-appendix}.
    } 
    \label{fig:different_depths}
    \vspace{-0.2cm}
\end{figure}

\subsection{Overparameterization in Terms of Increasing Width}
\label{sec:width}
Following~\citet{chiang2022loss}, we examine the influence of increasing the width of the network.  
To this end, we shrink all layers proportionally to the original LeNet architecture, 
e.g., width $\nicefrac{2}{6}$ keeps 2 out of the 6 convolutions in the first layer and $\operatorname{floor}(\nicefrac{2}{6} \cdot 16) = 5$ out of 16 convolutions in the second layer. Width $\nicefrac{1}{6}^*$ means a reduction of $\nicefrac{1}{6}$ for the convolution layers and $\nicefrac{1}{24}$ for the fully connected layers
(details in  Tables~\ref{tab:width_parameters} and~\ref{tab:width_parameters_cifar10} in the appendix).
We note that all the networks of smaller widths, even the smallest one, are overparameterized in the sense that the network can fit any set of training data (up to $32$ samples for MNIST and $24$ for CIFAR10). 
In fact, the result of \citet{nguyen18aOptLandscape} shows that if the number of neurons in the convolutional layer of the LeNet architecture exceeds the number of training points, then the set of parameters that do not produce linearly independent features at the convolutional layer has Lebesgue measure zero. This allows fitting 576 data points for MNIST and 784 data points for CIFAR10. 
Thus, we can see this experiment as starting with mild overparameterization for $\nicefrac{1}{6}^*$ and increasing to heavy overparameterization for the original LeNet.

\textbf{Comparison to the results of \citet{chiang2022loss}:} 
\citet{chiang2022loss} claim that widening the network 
widens the volume of good minima.
They deduce this by observing that the average test error of \gnc{}, conditional on a bin of their weight normalized loss, improves with increasing width, e.g., the boxes shown in the first row of Figure~\ref{fig:different_widths_loss_with_nornalization} for widths larger than $\nicefrac{1}{6}$.
However, as argued in Section~\ref{sec:normalized losses}, their weight normalized loss does not allow a direct comparison of different architectures conditional on loss bins, as it is only weakly related to the geometric properties of the function implemented by the network. 
Additionally, the comparison is complicated since some losses do not include samples, as evidenced by the empty boxes in Figure~\ref{fig:different_widths_loss_with_nornalization}.
To make it comparable across different architectures, we propose using our Lipschitz normalized loss. 

We illustrate the discrepancy of both normalized losses in Figure~\ref{fig:different_widths_loss_with_nornalization}, where we report the test accuracy vs the normalized loss for both the weight normalization of \citet{chiang2022loss} and the Lipschitz normalization for four different widths $(\nicefrac{1}{6}^*,\nicefrac{1}{6},\nicefrac{2}{6},\nicefrac{4}{6})$. First, we observe that the plots of the weight normalized loss and Lipschitz normalized loss are completely different and also behave differently for varying widths. Indeed, we see an \emph{improvement} of \gnc{} in terms of average test accuracy for a fixed bin of the weight normalized loss, confirming the observation of \citet{chiang2022loss}. 
However, when looking at a fixed loss bin of the Lipschitz normalized loss across widths, the average test accuracy is \emph{decreasing}, so we get the opposite statement of \citet{chiang2022loss}. 
It is noteworthy that the Lipschitz normalized loss decreases with width, indicating that the geometric margin to the decision boundary increases, strongly for \sgd{} and only slightly for \gnc{}. This shows that with increasing width, \sgd{} has an implicit bias towards networks with larger geometric margins. The strong improvement in geometric margin is also linked to significant improvement in average test accuracy for \sgd{} despite the \emph{increasing} overparameterization.

\setlength\cellspacetoplimit{0pt}
\setlength\cellspacebottomlimit{0pt}
\renewcommand\tabularxcolumn[1]{>{\centering\arraybackslash}S{p{#1}}}
\begin{figure*}[ht!]
\centering
\setlength\tabcolsep{0pt}
\adjustboxset{width=\linewidth,valign=c}
\centering
\begin{tabularx}{1.0\linewidth}{
@{}l  
  S{p{0.05\textwidth}} 
  *{4}{S{p{0.2245\textwidth}}} 
  S{p{0.052\textwidth}}}
    
    %%%%%%%%%%%%%%%%%%%%%%%%%
    &
    &\multicolumn{1}{c}{\quad \textbf{ $\mathbf{1}$ conv, $\mathbf{1}$ fc}}
    &\multicolumn{1}{c}{\quad \textbf{$\mathbf{2}$ conv, $\mathbf{1}$ fc}}
    &\multicolumn{1}{c}{\quad \textbf{$\mathbf{2}$ conv, $\mathbf{2}$ fc}}
    &\multicolumn{1}{c}{\quad \textbf{standard}}
    & \\
    %%%%%%%%%%%%%%%%%%%%%%%%%%%%%%%%%%%
    % GNC 1
    %%%%%%%%%%%%%%%%%%%%%%%%%%%%%%%%%%%
    &\rotatebox[origin=c]{90}{\textbf{\makecell{\gnc{}\\Test accuracy}}}
    &\includegraphics{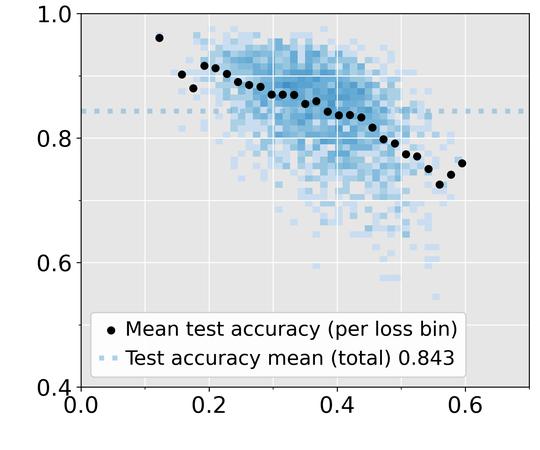}
    &\includegraphics{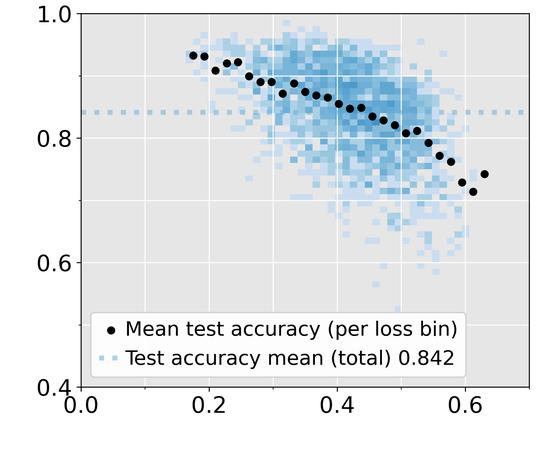}
    &\includegraphics{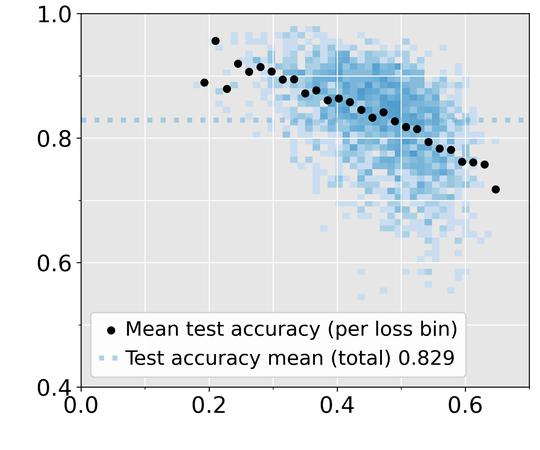}
    &\includegraphics{figures/different_widths/2d_hist_train_loss_normalize_grad_input_test_acc_mnist_guess_initialization_uniform_width033_seed202_permseed202_16_samples.jpg}
    &\includegraphics{figures/different_loss_factor/colorbar.png} \\
    %%%%%%%%%%%%%%%%%%%%%%%%%%%%%%%%%%%
    % SGD 1
    %%%%%%%%%%%%%%%%%%%%%%%%%%%%%%%%%%%
    &\rotatebox[origin=c]{90}{\textbf{\makecell{\sgd{}\\Test accuracy}}}
    &\includegraphics{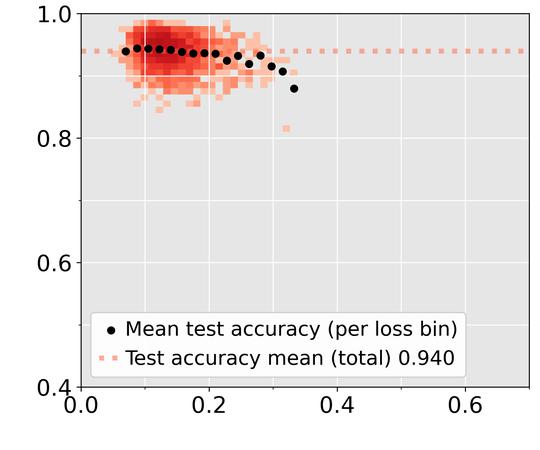}
    &\includegraphics{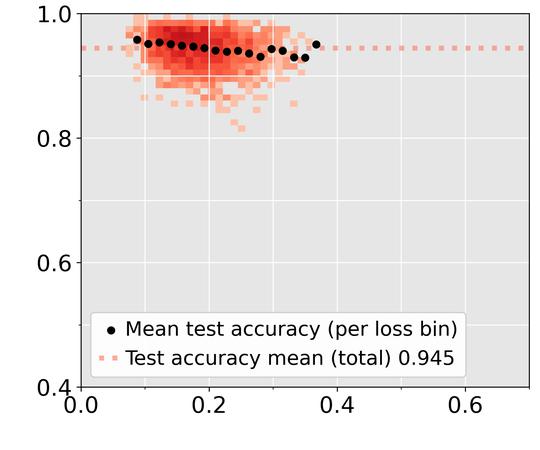}
    &\includegraphics{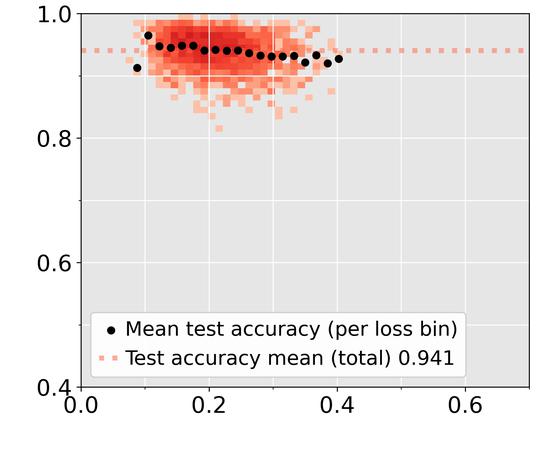}
    &\includegraphics{figures/different_widths/2d_hist_train_loss_normalize_grad_input_test_acc_mnist_sgd_initialization_kaiming_lr01_width033_epoch60_seed202_permseed202_16_samples.jpg}
    &\includegraphics{figures/different_loss_factor/colorbar_SGD.png} \\
    %%%%%%%%%%%%%%%%%%%%%%%%%%%%%%%%
    &
    & \multicolumn{1}{c}{\hspace{0.2em} \textbf{Train loss (normalized)}}
    & \multicolumn{1}{c}{\hspace{0.5em} \textbf{Train loss (normalized)}}
    & \multicolumn{1}{c}{\hspace{0.15em} \textbf{\enspace Train loss (normalized)}}
    & \multicolumn{1}{c}{\hspace{0.25em} \textbf{\enspace Train loss (normalized)}}
    &
    \end{tabularx}
    \caption{\textbf{Qualitative analysis of overparametrization in the depth.} 
    In contrast to increasing width, increasing depth decreases the geometric margin (higher normalized loss). This decrease holds both for {\color{azure}{\textbf{\gnc}}} (top)  and {\color{amaranth}{\textbf{\sgd}}} (bottom).
    We show 2000 LeNet models for each depth for classes 0 and 7 using a training set of size 16.  Results are more pronounced for harder class pairs (see     Figure~\ref{fig:different_depths_loss_mnist_appendix}).
    }
    \label{fig:different_depths_loss_mnist}
\end{figure*} 

\textbf{Behavior of \gnc{} for increasing width:} 
Increasing the width of over-parameterized neural networks is known to improve the generalization of \sgd{} \cite{hestness2017deep,allen2019convergence}.
This is evident in the left column of Figure~\ref{fig:different_widths}, where we observe a clear improvement of \sgd{} across all scenarios. 

Specifically, we report the average test accuracy of networks obtained by \sgd{} and \gnc{} as a function of increasing sample size for classes 0 vs 7 from MNIST and bird vs ship from CIFAR10, with the LeNet, MLP, and small ResNet architectures (See Appendix~\ref{sec:network_arch} for details about the different architectures). 
For each training set size and each algorithm, we generate $100$ fitting networks. 
The plotted curves are the average (and standard deviation) over four repetitions of this experiment, each with a different subset of training samples (See Appendix~\ref{sec:appendix-dataset} for more details).

In contrast to \sgd{}, the test accuracy of \gnc{} as a function of the width remains similar, except for a slight improvement in the LeNet architecture for CIFAR10 for widths up to $\nicefrac{2}{6}$.
This indicates that although \gnc{} shows no clear sign of overfitting with increasing width, it also exhibits no further improvement with overparameterization. 
Moreover, even the probability $\Pr_W(\TrainError=0)$ of finding a perfectly fitting network (right column of Figure~\ref{fig:different_widths}) remains constant. 
So, we can observe that even though the volume of random networks fitting the data does not seem to improve with the width, \sgd{} manages to converge to better networks. 
\textbf{Thus, for increasing width, we have clear evidence of an implicit bias of \sgd{}, which is helpful for improved generalization}. This bias results in convergence to networks of higher geometric margin (smaller Lipschitz normalized loss) with increasing width, as shown in  Figure~\ref{fig:different_widths_loss_with_nornalization}.

The orange dashed line in the plot of the test accuracy in Figure~\ref{fig:different_widths} shows the average test accuracy of the narrowest network for a significantly larger training set size of 400 samples (still perfectly fitting the training data). 
It shows that even for the narrowest network, there exists a set of weights that can lead to good generalization, but \sgd{} does not converge to them.
It is thus an interesting open question whether variants of \sgd{} or novel optimizers could possess implicit biases that lead to better generalization even in narrower networks - or formulated differently: Is the current trend towards larger architecture only an artifact of the optimizer, or is it really necessary to achieve better generalization?

\subsection{Overparameterization in Terms of Increasing Depth}
\label{sec:depth}
Next, we explore the impact of overparameterization when adding layers to the network architecture. We work backward by starting with the $\nicefrac{2}{6}$ architecture and then always discarding the largest layer.
This choice is motivated by our observation in Section~\ref{sec:width}, where we noted a decline in \gnc{} performance on CIFAR10 below this width for the LeNet architecture. 
The number of parameters in each configuration of the LeNet architecture can be found in Tables~\ref{tab:depth_parameters_mnist} and~\ref{tab:depth_parameters_cifar} for MNIST and CIFAR10, respectively.
The results are summarized in Figure~\ref{fig:different_depths} using a similar setup as the one described in the previous subsection for increasing width.

Interestingly, while overparameterizing the network in terms of width improved the result of \sgd{} and at least did not hurt \gnc{} performance, the effect of depth is the opposite.
In the left column of Figure \ref{fig:different_depths}, we observe that \gnc{} consistently performs worse with increasing depth, while the behavior of \sgd{} either decreases or remains the same, depending on the architecture and the dataset.
For example, in the LeNet architecture, for easy problems like 0 vs 7, \sgd{} can compensate for the additional complexity due to depth with its 
bias towards more simple solutions. 
For more complex problems (CIFAR10, MNIST 3 vs 5, see also Figure~\ref{fig:different_depths_mnist} in the appendix), it suffers in the same way as \gnc{}. 

Thus, it seems that we see classical overfitting in overparameterized networks: more complex networks overfit on the small training set and, therefore, do not generalize as well.  
This is also reflected in $\Pr_W(\TrainError=0)$, as the probability of finding a shallow network fitting the data is much higher than for a deep one.
Since \gnc{} and \sgd{} behave similarly in this context, one can state that there is a negative architectural bias with increasing depth, at least for the training set size considered here.

Note that we do not claim that increasing depth is detrimental in general.
Instead, \textbf{we attribute 
the effect of depth
to an architectural bias since it affects \gnc{} and does not seem to influence the optimization process separately.}
If sufficient data is available, it is well documented in the literature that deeper networks
can yield strong generalization performance \cite{he2016deep,xie2019intriguing}.
We speculate that this also holds for \gnc{}, but it is out of reach to verify numerically due to the exponentially increasing cost of sampling.

Additionally, we present in Figure~\ref{fig:different_depths_loss_mnist} the loss-accuracy plots for different depths, similar to the analysis in Figure~\ref{fig:different_widths_loss_with_nornalization}.
It can be seen that shallower networks with fewer parameters have lower Lipschitz normalized loss, which means larger geometric margins.
This finding is in contrast with the results regarding overparameterization in terms of width, where wider networks with more parameters achieve larger geometric margins.

\section{Conclusion}
We performed an extensive comparison of \sgd{} and \gnc{} to disentangle the effects of the implicit bias of \sgd{} versus the architectural bias of the network. 
Focusing on binary classification tasks in the low sample regime from the MNIST and CIFAR10 datasets, we demonstrated that the improvements of \sgd{} with increasing overparameterization in terms of width can be attributed to its implicit bias, as \gnc{} does not improve with width. 
Conversely, we showed that overparameterization in terms of depth is detrimental to generalization in the low sample regime and exhibits signs of classical overfitting. As \sgd{} and \gnc{} behave similarly in this case, this can be attributed to a negative architectural bias of the neural network. 
Disentangling the effects of optimization and architecture could provide deeper insights into why overparameterized networks perform so well in practice and ideally help to achieve similar generalization abilities with smaller networks.

\clearpage
\section*{Acknowledgements}
The authors thank Václav Voráček for valuable discussions, Naama Pearl and Maximilian Müller for their assistance with writing the manuscript, and the International Max Planck Research School for Intelligent Systems (IMPRS-IS) for supporting AP. We acknowledge support from the Deutsche Forschungsgemeinschaft (DFG, German Research Foundation) under Germany's Excellence Strategy (EXC number 2064/1, project number 390727645), as well as in the priority program SPP 2298, project number 464101476.
\section*{Impact Statement}
This paper presents work whose goal is to advance the field of Machine Learning. There are many potential societal consequences of our work, none of which we feel must be specifically highlighted here.

\bibliography{references}
\bibliographystyle{icml2024}

%%%%%%%%%%%%%%%%%%%%%%%%%%%%%%%%%%%%%%%%%%%%%%%%%%%%%%%%%%%%%%%%%%%%%%%%%%%%%%%
%%%%%%%%%%%%%%%%%%%%%%%%%%%%%%%%%%%%%%%%%%%%%%%%%%%%%%%%%%%%%%%%%%%%%%%%%%%%%%%
% APPENDIX
%%%%%%%%%%%%%%%%%%%%%%%%%%%%%%%%%%%%%%%%%%%%%%%%%%%%%%%%%%%%%%%%%%%%%%%%%%%%%%%
%%%%%%%%%%%%%%%%%%%%%%%%%%%%%%%%%%%%%%%%%%%%%%%%%%%%%%%%%%%%%%%%%%%%%%%%%%%%%%%
\newpage
\appendix
\onecolumn
\section{Appendix}
\label{sec:appendix}
In the following sections of the appendix, we expand upon the experiments introduced in the main experiment section (Section~\ref{sec:experiments}). Initially, we provide details on the architectures and datasets in Appendix~\ref{sec:implementation}. Appendix~\ref{sec:init-appendix} further explores experiments related to the initialization analysis outlined in Section~\ref{sec:init}. Appendix~\ref{sec:width-appendix} presents supplementary experiments for the width analysis discussed in Section~\ref{sec:width}. In Appendix~\ref{sec:depth-appendix}, we present additional experiments related to the depth analysis covered in Section~\ref{sec:depth}.

\section{Implementation Details}
\label{sec:implementation}

\subsection{Network Architectures}
\label{sec:network_arch}
\subsubsection{LeNet}
\label{sec:lenet_arch}
\begin{figure}[h]
  \centering
  \includegraphics[width=\columnwidth]{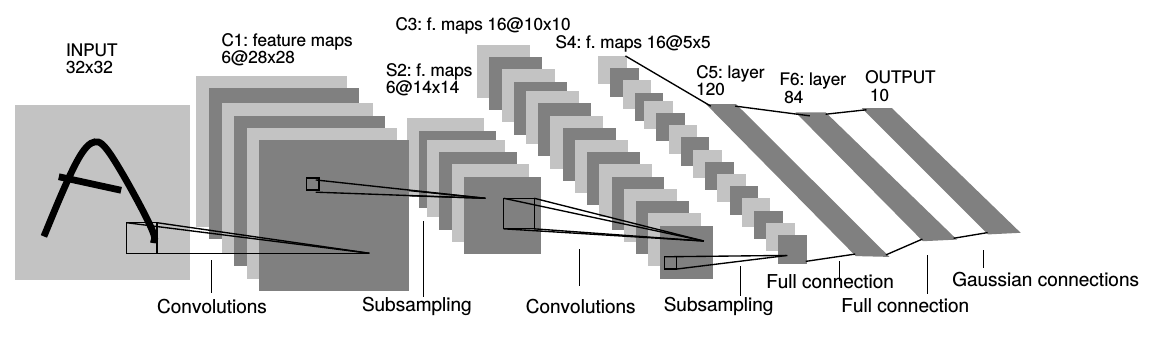} 
  \caption{LeNet architecture. The figure is taken from the original paper \cite{lecun1998gradient}}
  \label{fig:lenet_architecture}
\end{figure}

The LeNet architecture, as appears in~\citet{lecun1998gradient}, is depicted in Figure~\ref{fig:lenet_architecture}.
The first two layers consist of convolutions with a kernel size of $5 \times 5 \times k$, where $k$ represents the input channels' dimension of that layer. 
For MNIST, the first layer has 1 channel, and for CIFAR10, it has 3 channels.
The remaining layers are fully connected.

The original architecture's number of layers and parameter counts are summarized in the row indicated by a width factor of 1 in Table~\ref{tab:width_parameters} (MNIST) and Table~\ref{tab:width_parameters_cifar10} (CIFAR10). Other rows illustrate the corresponding numbers for varying width factors (see Section~\ref{sec:width}). 
Additionally, we include the corresponding tables (Table~\ref{tab:depth_parameters_mnist} and Table~\ref{tab:depth_parameters_cifar}) for varying numbers of layers (Section~\ref{sec:depth}) for a width factor of $\nicefrac{2}{6}$.

\subsubsection{MLP}
The MLP network has five layers with 120, 60, 30, 12, and 2 neurons, respectively.
Since we want the network to be over-parametrized but remain within a feasible number of parameters, we resized the input images by 2 with a max pooling layer, yielding a network with 101,768 parameters for the CIFAR10 dataset and 33,128 for MNIST. 
Increasing the width is done in the same manner as for the LeNet. 
When changing the depth, we use a network with a width of 2/6 and gradually remove the largest layer, as was done for the LeNet architecture.

\begin{table}[ht]
\caption{
\textbf{Parameter count in the LeNet architecture for the MNIST task with varying \emph{width} factors}.
A width factor of 1 indicates the standard LeNet architecture.
The corresponding parameter numbers in the convolutional layers are indicated in parentheses.
In the last fully connected layer, the parameter count solely depends on the input size and preceding layers.
For improved readability, bias terms are excluded from the per-layer parameter count.}
\vskip 0.1in
        \centering
        \begin{tabularx}{\linewidth}{ @{}c  *{2}{>{\centering\arraybackslash}X}  *{3}{>{\centering\arraybackslash}X}  @{}c } 
            \hline
            \multicolumn{1}{c}{\textbf{Width}}  & \multicolumn{2}{c}{\textbf{Convolutions layers}} & \multicolumn{3}{c}{\textbf{Fully connected layers}} & \textbf{Total} \\
            \hline
            \hline
            $\nicefrac{1}{6}^*$ & 1 (25)  & 2 (50)   & 5 (160) & 3 (15) & 2 (6) & 269 \\
            \hline
            \nicefrac{1}{6} & 1 (25)  & 2 (50)   & 20 (640) & 14 (280) & 2 (28) & 1062 \\
            \hline
            \nicefrac{2}{6} & 2 (50)  & 5 (250)  & 40 (3200) & 28 (1120) & 2 (56) & 4753 \\
            \hline
            \nicefrac{3}{6} & 3 (75)  & 8 (600)  & 60 (7680) & 42 (2520) & 2 (84) & 11074 \\
            \hline
            \nicefrac{4}{6} & 4 (100) & 10 (1000) & 80 (12800) & 56 (4480) & 2 (112) & 18644 \\
            \hline
            \nicefrac{5}{6} & 5 (125) & 13 (1625) & 100 (20800) & 70 (7000) & 2 (140) & 29880 \\
            \hline
            1 & 6 (150) & 16 (2400) & 120 (30720) & 84 (10080) & 2 (168) & 43746 \\
            \hline
        \end{tabularx}

        \label{tab:width_parameters}
    \end{table}

\newcolumntype{L}[1]{>{\raggedright\let\newline\\\arraybackslash\hspace{0pt}}m{#1}}
\newcolumntype{C}[1]{>{\centering\let\newline\\\arraybackslash\hspace{0pt}}m{#1}}
\newcolumntype{R}[1]{>{\raggedleft\let\newline\\\arraybackslash\hspace{0pt}}m{#1}}

\begin{table}[t]
\caption{
    \textbf{Parameter count in the LeNet architecture for the CIFAR10 task with varying \emph{width} factors}.}
\vskip 0.1in
\centering
\begin{tabularx}{\linewidth}{@{}c 
*{2}{>{\centering\arraybackslash}X} 
*{3}{>{\centering\arraybackslash}X} 
@{}c} 
    \hline
    \multicolumn{1}{c}{\textbf{Width}}  & \multicolumn{2}{c}{\textbf{Convolutions layers}} & \multicolumn{3}{c}{\textbf{Fully connected layers}} & \textbf{Total} \\
    \hline
    \hline
    $\nicefrac{1}{6}^*$ & 1 (75)  & 2 (50)   & 5 (250) & 3 (15) & 2 (6) & 409 \\
    \hline
    \nicefrac{1}{6} & 1 (75)  & 2 (50)   & 20 (1000) & 14 (280) & 2 (28) & 1472 \\
    \hline
    \nicefrac{2}{6} & 2 (150)  & 5 (250)  & 40 (5000) & 28 (1120) & 2 (56) & 6653 \\
    \hline
    \nicefrac{3}{6} & 3 (225)  & 8 (600)  & 80 (12000) & 42 (2520) & 2 (84) & 15544 \\
    \hline
    \nicefrac{4}{6} & 4 (300) & 10 (1000) & 80 (20000) & 56 (4480) & 2 (112) & 26044 \\
    \hline
    \nicefrac{5}{6} & 5 (375) & 13 (1625) & 100 (32500) & 70 (7000) & 2 (140) & 41830 \\
    \hline
    1 & 6 (450) & 16 (2400) & 120 (48000) & 84 (10080) & 2 (168) & 61326 \\
    \hline
\end{tabularx}
\label{tab:width_parameters_cifar10}
\end{table}
\begin{table}[ht]
\caption{
\textbf{Parameter count in the LeNet architecture for the MNIST task with varying \emph{depth} factors}.
The \emph{width} of the network for all of these experiments is \nicefrac{2}{6}.
The first row is the standard layers configuration of two convolution layers, followed by 3 fully connected layers.
\emph{c} and \emph{f} represent a convolutional layer and a fully connected layer, respectively.}
\vskip 0.1in
\centering
\begin{tabularx}{\linewidth}{ 
@{}l  
*{2}{>{\centering\arraybackslash}X}  
*{3}{>{\centering\arraybackslash}X}  
@{}c } 
    \hline
    \multicolumn{1}{c}{\textbf{Layers}}  & \multicolumn{2}{c}{\textbf{Convolutions layers}} & \multicolumn{3}{c}{\textbf{Fully connected layers}} & \textbf{Total} \\
    \hline
    \hline
    \enspace \; 2c-3f & 2 (50)  & 5 (250)   & 40 (3200) & 28 (1120) & 2 (56) & 4753 \\
    \hline
    \enspace \; 2c-2f & 2 (150)  & 5 (250)   & ----- & 28 (2240) &  2 (56)  & 2263 \\
    \hline
    \enspace \; 2c-1f & 2 (150)  & 5 (250)  & ----- & ----- & 2 (160)  & 469 \\
    \hline
    \enspace \; 1c-1f & 2 (50)  & -----   & ----- & ----- & 2 (144) & 198 \\
    \hline
\end{tabularx}
\label{tab:depth_parameters_mnist}
\end{table}

\begin{table}[ht]
\caption{
\textbf{Parameter count in the LeNet architecture for the CIFAR10 task with varying \emph{depth} factors}.}
\vskip 0.1in
\centering
    \begin{tabularx}{\linewidth}{ @{}l  *{2}{>{\centering\arraybackslash}X}  *{3}{>{\centering\arraybackslash}X}  @{}c } 
        \hline
        \multicolumn{1}{c}{\textbf{Layers}}  & \multicolumn{2}{c}{\textbf{Convolutions layers}} & \multicolumn{3}{c}{\textbf{Fully connected layers}} & \textbf{Total} \\
        \hline
        \hline
        \enspace \; 2c-3f & 2 (150)  & 5 (250)   & 40 (5000) & 28 (1120) & 2 (56) & 6653 \\
        \hline
        \enspace \; 2c-2f & 2 (150)  & 5 (250) & -----  & 28 (3500) & 2 (56) & 3993 \\
        \hline
        \enspace \; 2c-1f & 2 (150)  & 5 (250) & ----- & ----- & 2 (250) & 659 \\
        \hline
        \enspace \; 1c-1f & 2 (150)  & ----- & ----- & -----  & 2 (196) & 350 \\
        \hline
    \end{tabularx}       \label{tab:depth_parameters_cifar}
\end{table}

\subsubsection{ResNet}
For the ResNet-based architecture, we followed \citet{davidcpage2023cifar10-fast}, which trains a custom ResNet architecture in 75 seconds with 94\% accuracy on CIFAR10. The specific implementation contains three blocks, including a total of eight convolutional layers and one linear layer. We reduce the number of channels in each layer such that for a width factor of 1 and CIFAR10, the network contains 58,350 learnable parameters (similar to the LeNet).
% MNIST: 58,242
As with the other architectures, we used a constant learning rate and avoided augmentations of the dataset.

\subsection{Dataset}
\label{sec:appendix-dataset}
For each binary classification task, we used a subset of the training images and the entire test set for the class pair. 
For instance, for MNIST 0 vs 7 with 16 training samples, a subset of 16 samples was selected from the original training set (eight images of 0 and eight images of 7), and the test set contains the entire original test set of 2008 images. %1028, 980
For the loss vs test accuracy plots (e.g., Figures~\ref{fig:different_initializations},~\ref{fig:different_widths_loss_with_nornalization},~\ref{fig:different_depths_loss_mnist}), the same subset was used across all experiments to ensure that randomnesses stem only from the initialization (\gnc{}) and the additional randomness of the algorithm (\sgd{}).
For the samples vs test accuracy plots (e.g., Figures~\ref{fig:different_widths},~\ref{fig:different_depths}), we average results over four different training subsets (same four subsets for each experiment). Additionally, when increasing the number of samples, we only add images to the existing sets, maintaining the previous training images (i.e., the training sets are increasing sets).

\section{Dependency of \sgd{} and \gnc{} on the Prior $P(W)$}
\label{sec:init-appendix}
In Section~\ref{sec:init}, we analyze the effect of different initializations on the generalization capabilities of \sgd{} and \gnc{}.
This section extends the experiments in Section~\ref{sec:init} in the main paper.

\begin{figure}[htb]
    \centering
    \includegraphics[width=0.3\linewidth]{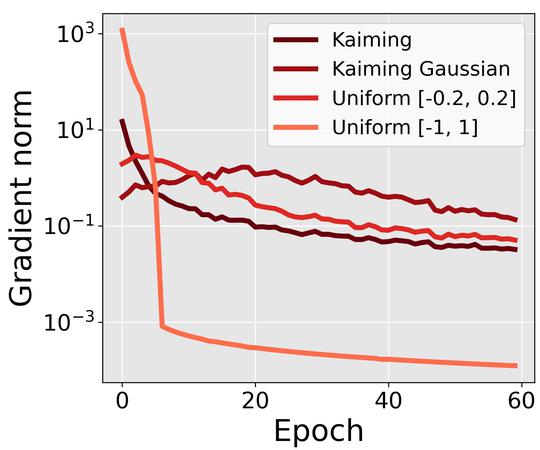}
    \caption{$\mathbf{\mathcal{U}[-1, 1]}$ \textbf{initialization prevents effective gradient-based learning.} When using $\mathcal{U}[-1, 1]$, the initialized weights are large, leading to high logit values and, in turn, to low loss and small gradients as soon as one reaches zero training loss, effectively stopping \sgd{}. For proper initialization, like the commonly used Kaiming or the downscaled $\mathcal{U}[-0.2, 0.2]$, logit values are lower, and \sgd{} optimizes properly, see Section~\ref{sec:init} for more details.} 
    \label{fig:initialization_gradients}
\end{figure}

\begin{figure*}[htb]
  \setlength\tabcolsep{0pt}
  \adjustboxset{width=\linewidth,valign=c}
  \centering
  \begin{tabularx}{1.0\linewidth}{
  @{}l 
  S{p{0.05\textwidth}} 
  *{4}{S{p{0.2245\textwidth}}} 
  S{p{0.052\textwidth}}}
    %%%%%%%%%%%%%%%%%%%%%%
    &
    & \multicolumn{1}{c}{\textbf{\quad Uniform [-1, 1]}}
    & \multicolumn{1}{c}{\textbf{\quad Uniform [-0.2, 0.2]}}
    & \multicolumn{1}{c}{\textbf{\quad Kaiming Uniform}}
    & \multicolumn{1}{c}{\textbf{\quad Kaiming Gaussian}}
    & \multicolumn{1}{c}{} \\
    
    %%%%%%%%%%%%%%%%%%%%%%
    % SGD 
    %%%%%%%%%%%%%%%%%%%%%%%%
    &\rotatebox[origin=c]{90}{\textbf{\makecell{\quad \sgd{}\\ \quad Test accuracy}}}
    &\includegraphics{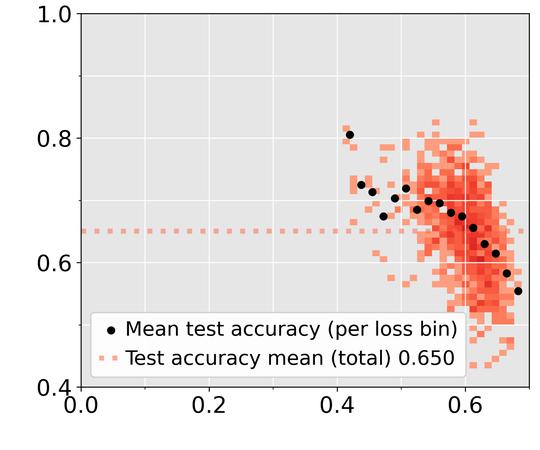}
    & \includegraphics{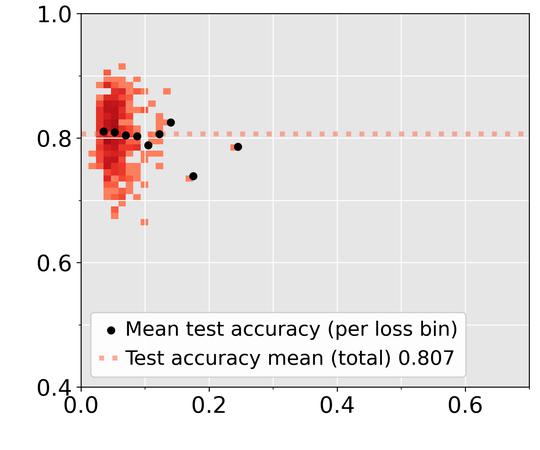}
    & \includegraphics{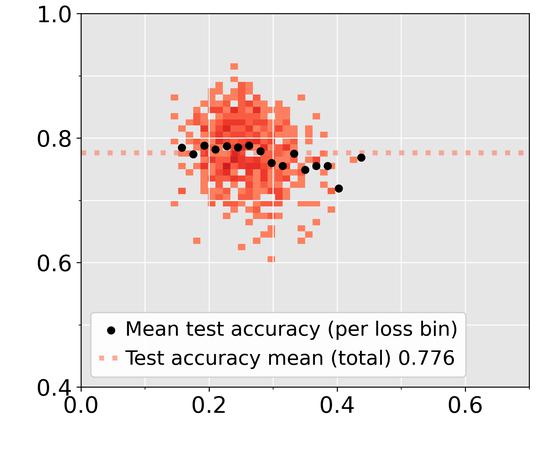}
    & \includegraphics{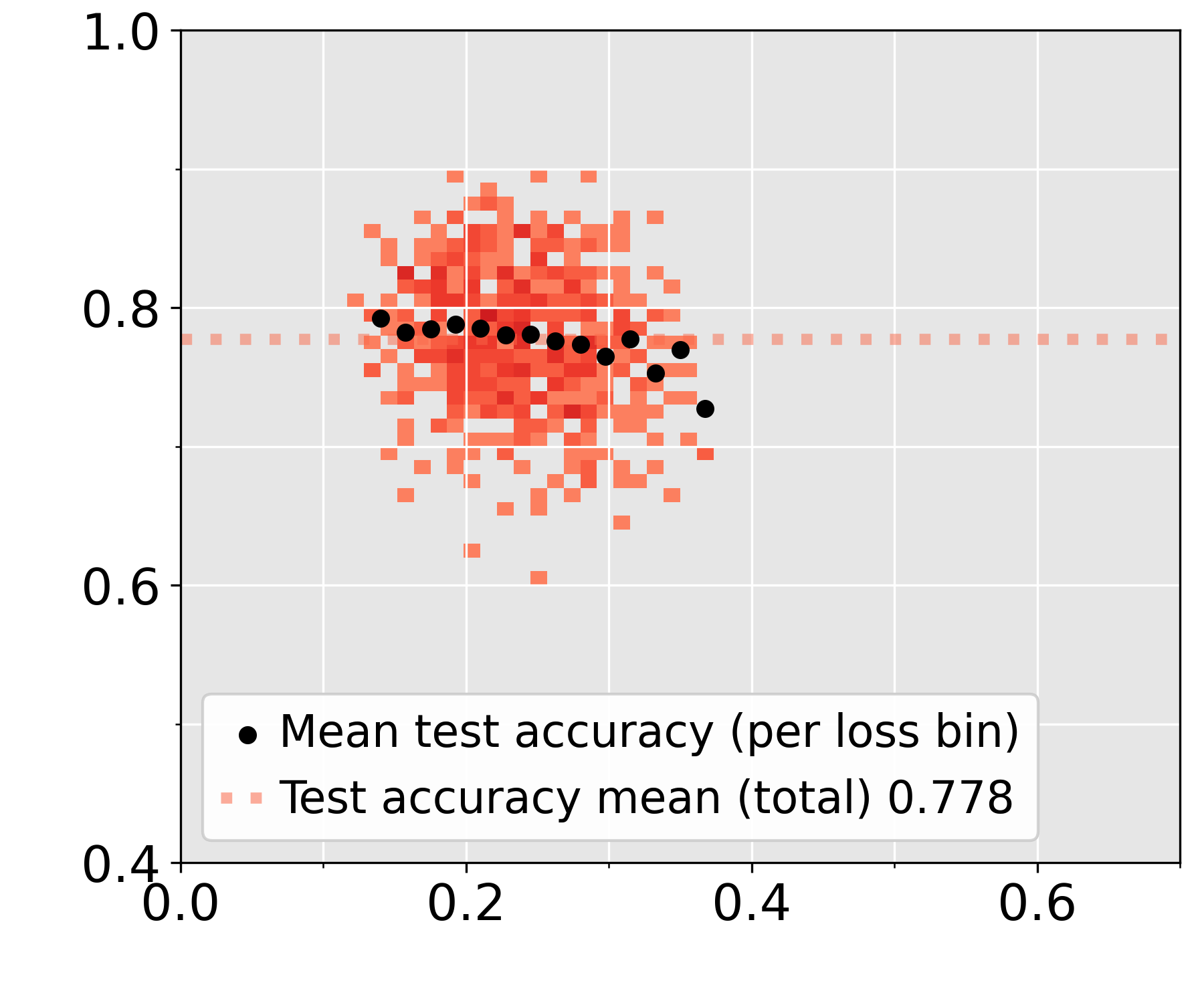}
    & \includegraphics{figures/different_loss_factor/colorbar_SGD.png} \\
    
    %%%%%%%%%%%%%%%%%%%%%%
    % GNC 
    %%%%%%%%%%%%%%%%%%%%%%%%
    &\rotatebox[origin=c]{90}{\textbf{\makecell{\quad \gnc{}\\ \quad Test accuracy}}}
    & \includegraphics{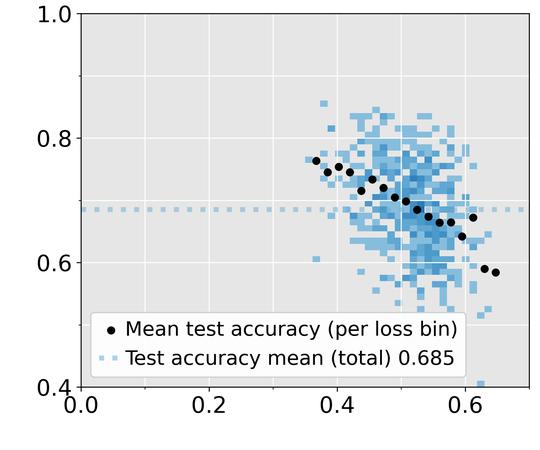}
    & \includegraphics{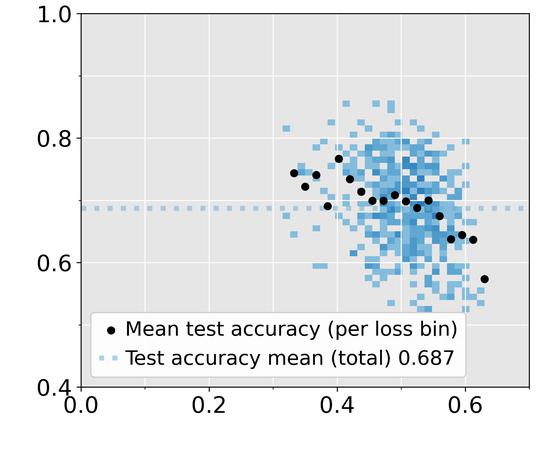}
    & \includegraphics{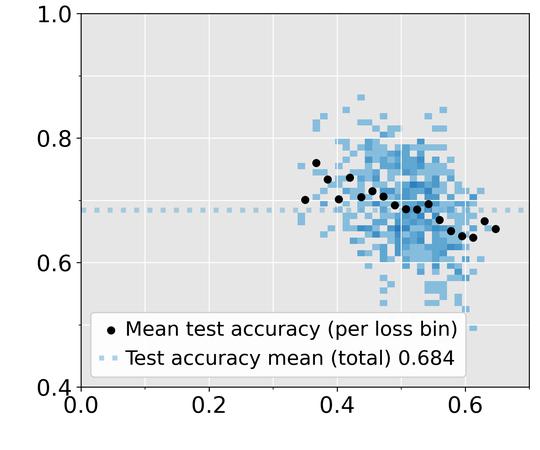}
    & \includegraphics{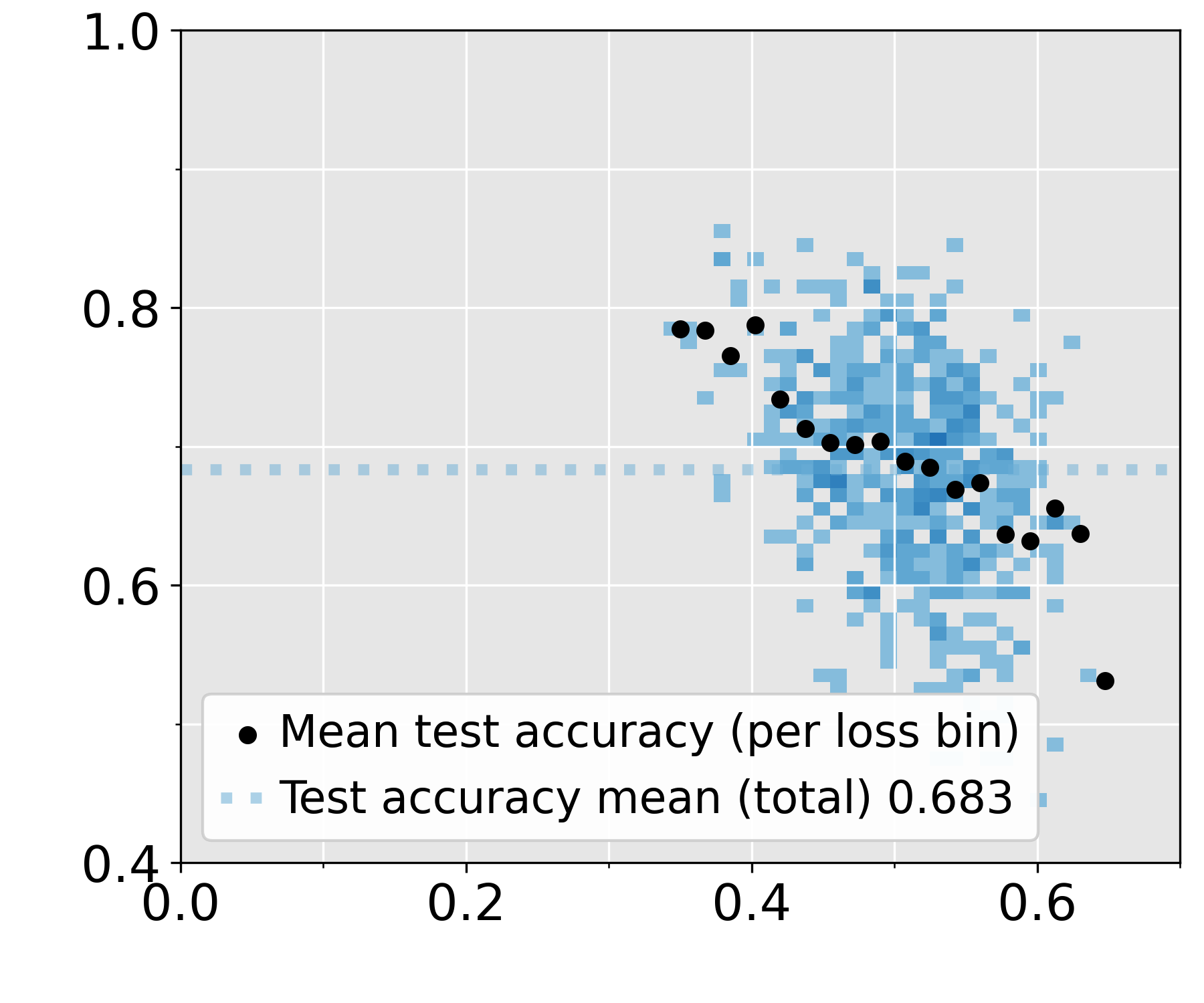}
    & \includegraphics{figures/different_loss_factor/colorbar.png} \\
    %%%%%%%%%%%%%%%%%%%%%%%%%%%%%%%%%%%%%%%%%%%%%%%%%%%%%%%%%
    % x axis labels
    %%%%%%%%%%%%%%%%%%%%%%%%%%%%%%%%%%%%%%%%%%%%%%%%%%%%%%%%%
    &
    & \multicolumn{1}{c}{\hspace{1.5em}\textbf{\makecell{Train loss\\(normalized)}}}
    & \multicolumn{1}{c}{\hspace{1.5em}\textbf{\makecell{Train loss\\(normalized)}}}
    & \multicolumn{1}{c}{\hspace{1.5em}\textbf{\makecell{Train loss\\(normalized)}}}
    & \multicolumn{1}{c}{\hspace{1.5em}\textbf{\makecell{Train loss\\(normalized)}}}
    &\\
  \end{tabularx}
\caption{\textbf{Generalization of \colorsgd{\sgd{} (optimized)} versus \colorgnc{\gnc{} (randomly sampled)} in dependency of the prior on the weights $\Pr(W)$:}  
   We ``train'' $500$ models to $100\%$ train accuracy for $16$ training samples from classes \emph{3} and \emph{5} of the MNIST dataset.
  Test accuracies for \gnc{} are similar across initializations, and the Lipschitz normalized loss (see Sec.~\ref{sec:method}) is similar across the uniform distributions.
  \textbf{Column 1:} For $\mathcal{U}[-1, 1]$ initialization (used in \citet{chiang2022loss}), the Lipschitz normalized losses and the test accuracy of \sgd{} and \gnc{} are similar, except for the convergence of \sgd{} towards more low-margin solutions. 
  The claim in \citet{chiang2022loss} that \gnc{} resembles \sgd{}, conditional on average test accuracy in the normalized loss bin (black dots), is an artifact of the suboptimal convergence of \sgd{} caused by this initialization.
  \textbf{Columns 2-4:} For other initializations, \sgd{} (first row) improves considerably both in terms of loss and accuracy.
  In contrast, \gnc{} remains unaffected, as it is independent of the weight scale in each layer.}
  \label{fig:different_initializations_appendix1}
\end{figure*}

\begin{figure*}[htb]
\centering
\setlength\tabcolsep{0pt}
\adjustboxset{width=\linewidth,valign=c}
\centering
\begin{tabularx}{0.8155\linewidth}{
@{}l 
S{p{0.04\textwidth}} 
S{p{0.05\textwidth}} 
*{3}{S{p{0.2245\textwidth}}} 
S{p{0.052\textwidth}}}
    %%%%%%%%%%%%%%%%%%%%%%
    &
    &
    & \multicolumn{1}{c}{\textbf{\quad Uniform [-1, 1]}}
    & \multicolumn{1}{c}{\textbf{\quad Kaiming Uniform}}
    & \multicolumn{1}{c}{\textbf{\quad Kaiming Gaussian}}
    & \\
    
    %%%%%%%%%%%%%%%%%%%%%%
    % SGD 2 
    %%%%%%%%%%%%%%%%%%%%%%%%
    &\multirow{2}{*}[-2em]{\rotatebox{90}{\textbf{Bird vs Ship}}}
    &\rotatebox[origin=c]{90}{\textbf{\makecell{\sgd{}\\Test accuracy}}}
    & \includegraphics{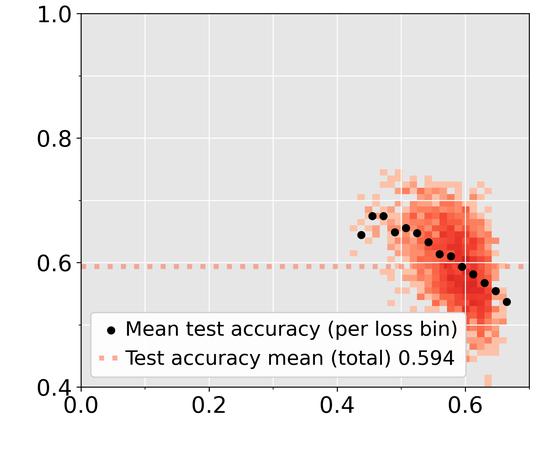}
    & \includegraphics{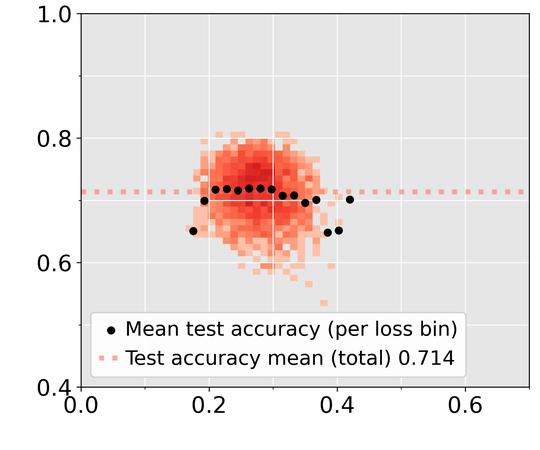}
    & \includegraphics{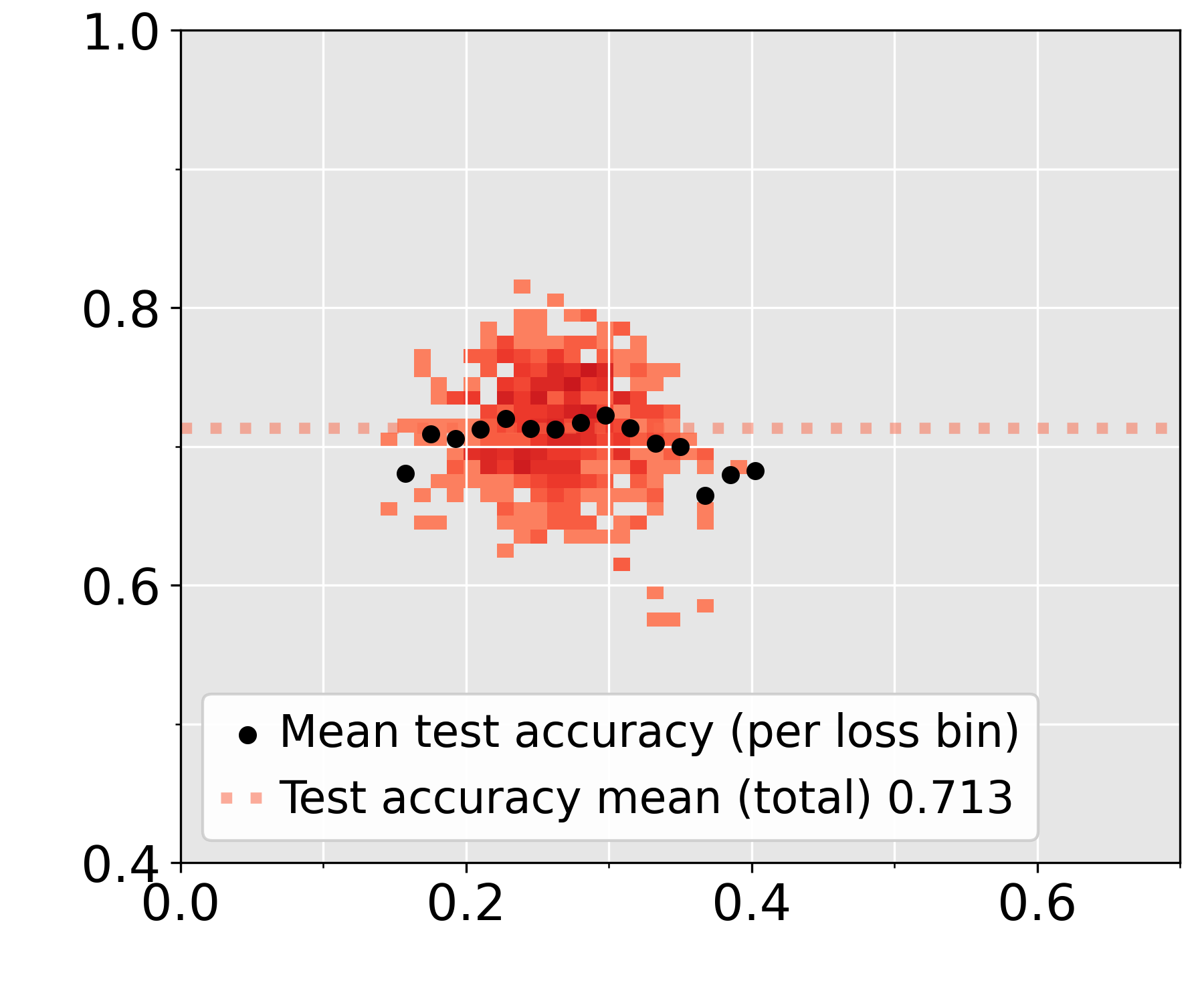}
    & \includegraphics{figures/different_loss_factor/colorbar_SGD.png} \\
    %%%%%%%%%%%%%%%%%%%%%%
    % GNC 1
    %%%%%%%%%%%%%%%%%%%%%%%%
    &
    &\rotatebox[origin=c]{90}{\textbf{\makecell{\gnc{}\\Test accuracy}}}
    &\includegraphics{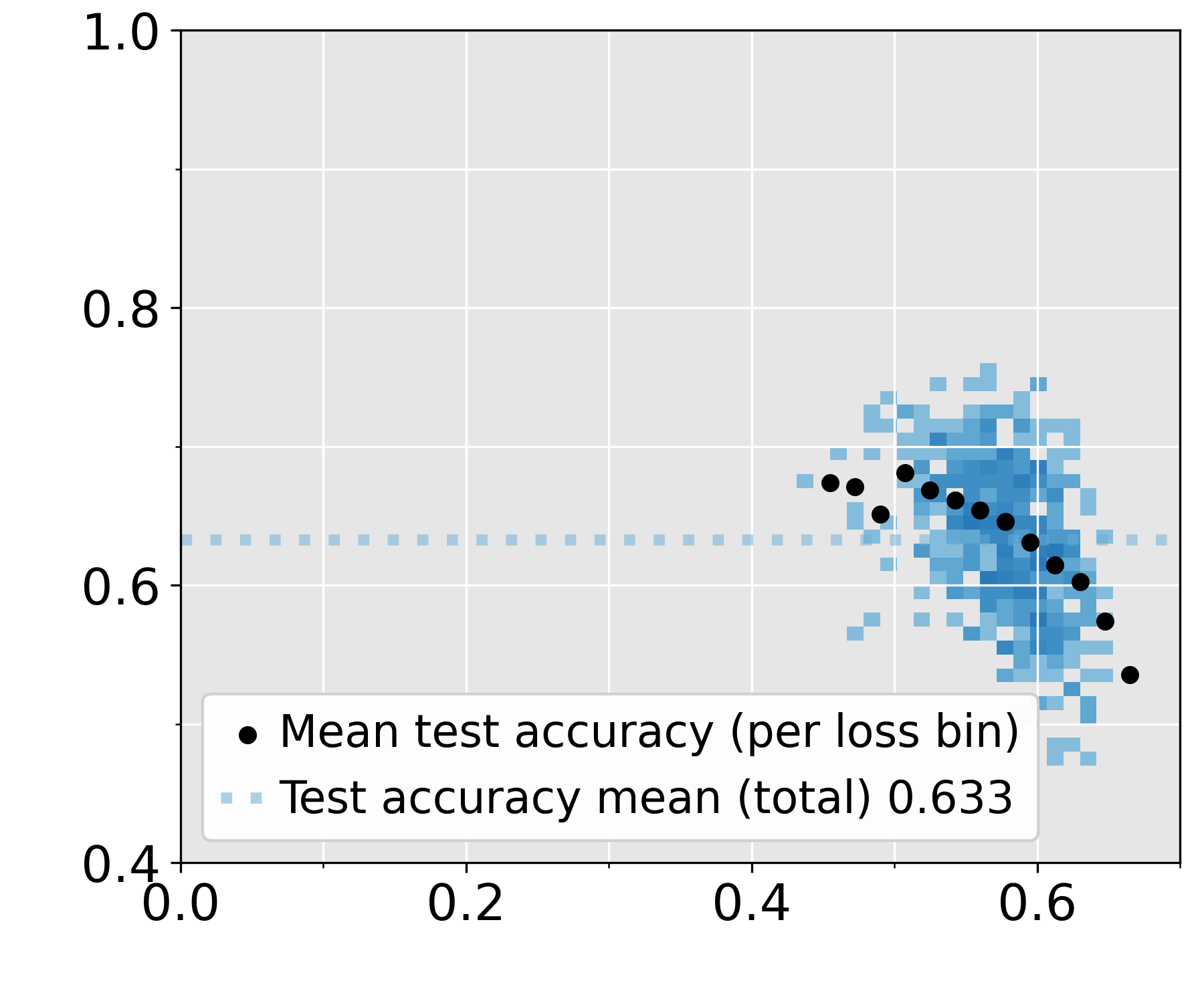}
    & \includegraphics{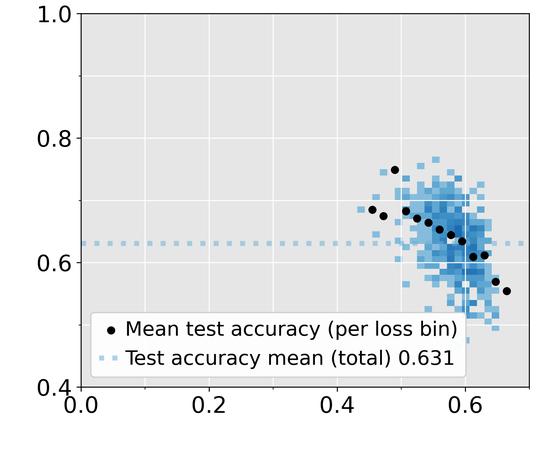}
    & \includegraphics{figures/different_loss_factor/cifar/2d_hist_train_loss_normalize_grad_input_test_acc_cifar_201_201_lenet_guess_kaiming_gaussian_width05_16samples_fixed}
    & \includegraphics{figures/different_loss_factor/colorbar.png} \\
    %%%%%%%%%%%%%%%%%%%%%%%%%%
    %%%%%%%%%%%%%%%%%%%%%%
    % SGD 
    %%%%%%%%%%%%%%%%%%%%%%%%
    &\multirow{2}{*}[-2em]{\rotatebox{90}{\textbf{Deer vs Truck}}}
    % &\rotatebox[origin=c]{90}{\textbf{Lipschitz norm}}
    &\rotatebox[origin=c]{90}{\textbf{\makecell{\sgd{}\\Test accuracy}}}
    &\includegraphics{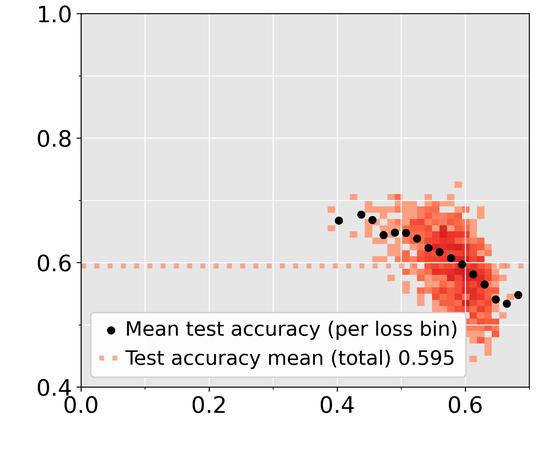}
    & \includegraphics{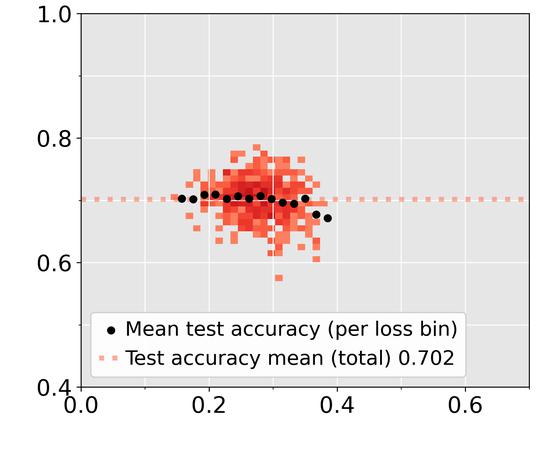}
    & \includegraphics{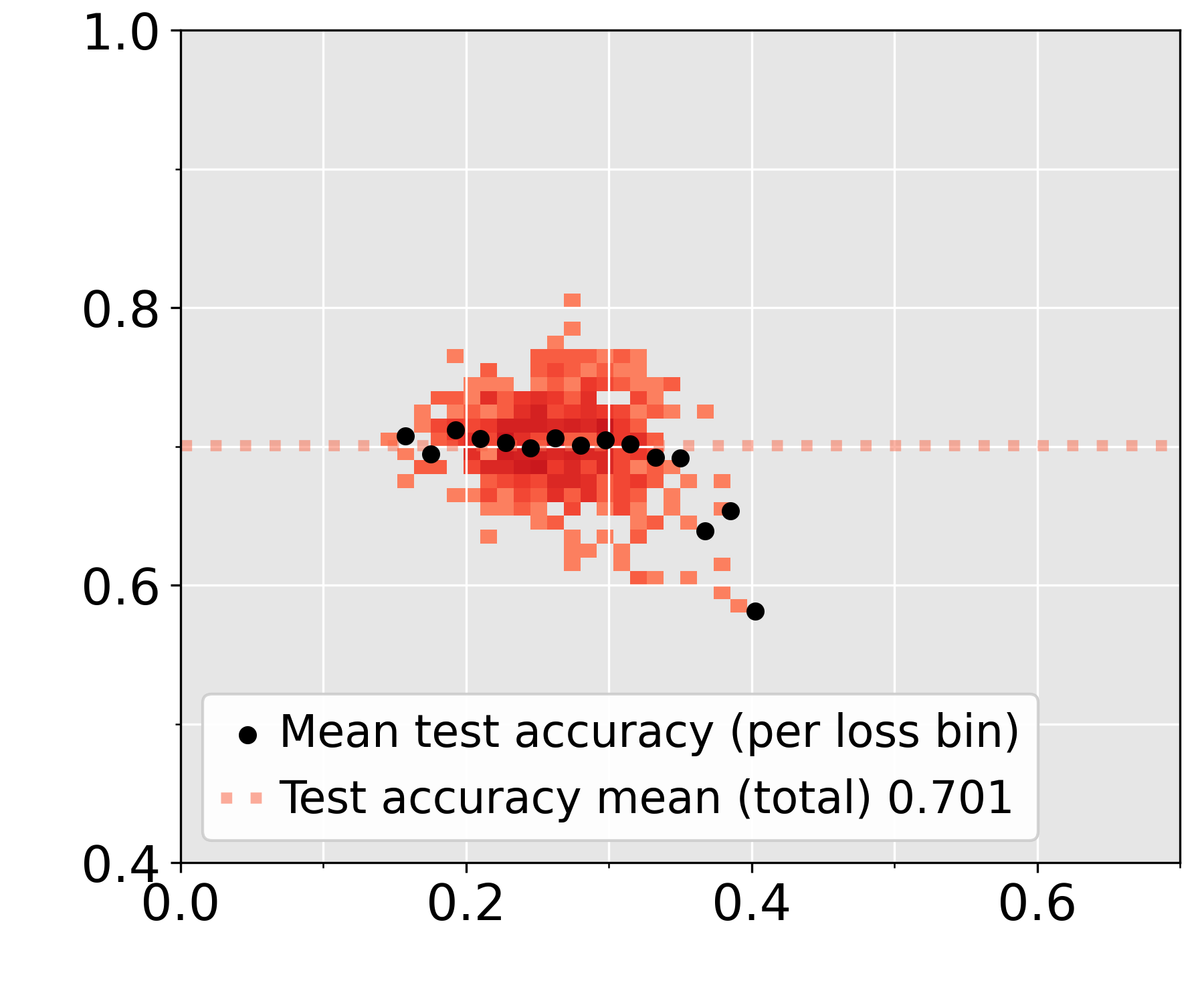}
    & \includegraphics{figures/different_loss_factor/colorbar_SGD.png} \\
    %%%%%%%%%%%%%%%%%%%%%%
    % GNC 
    %%%%%%%%%%%%%%%%%%%%%%%%
    &
    &\rotatebox[origin=c]{90}{\textbf{\makecell{\gnc{}\\Test accuracy}}}
    & \includegraphics{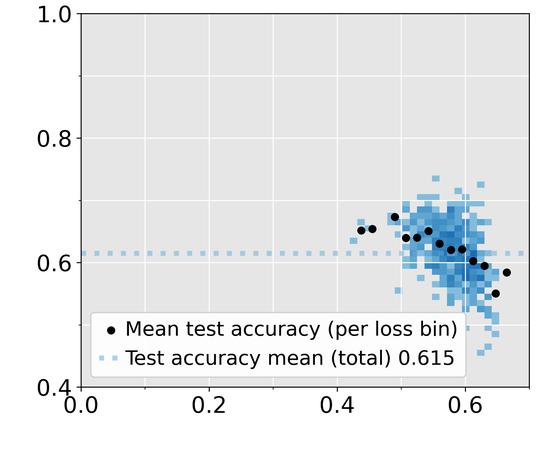}
    & \includegraphics{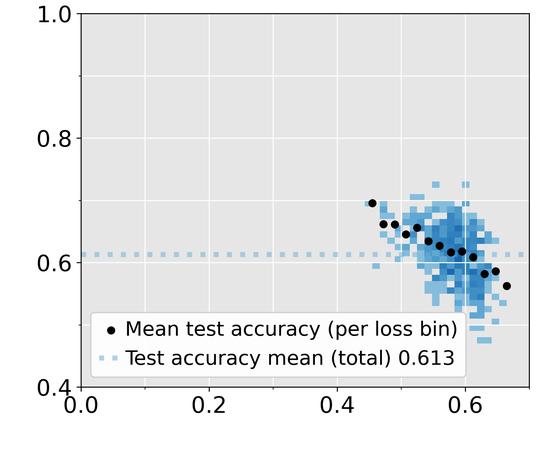}
    & \includegraphics{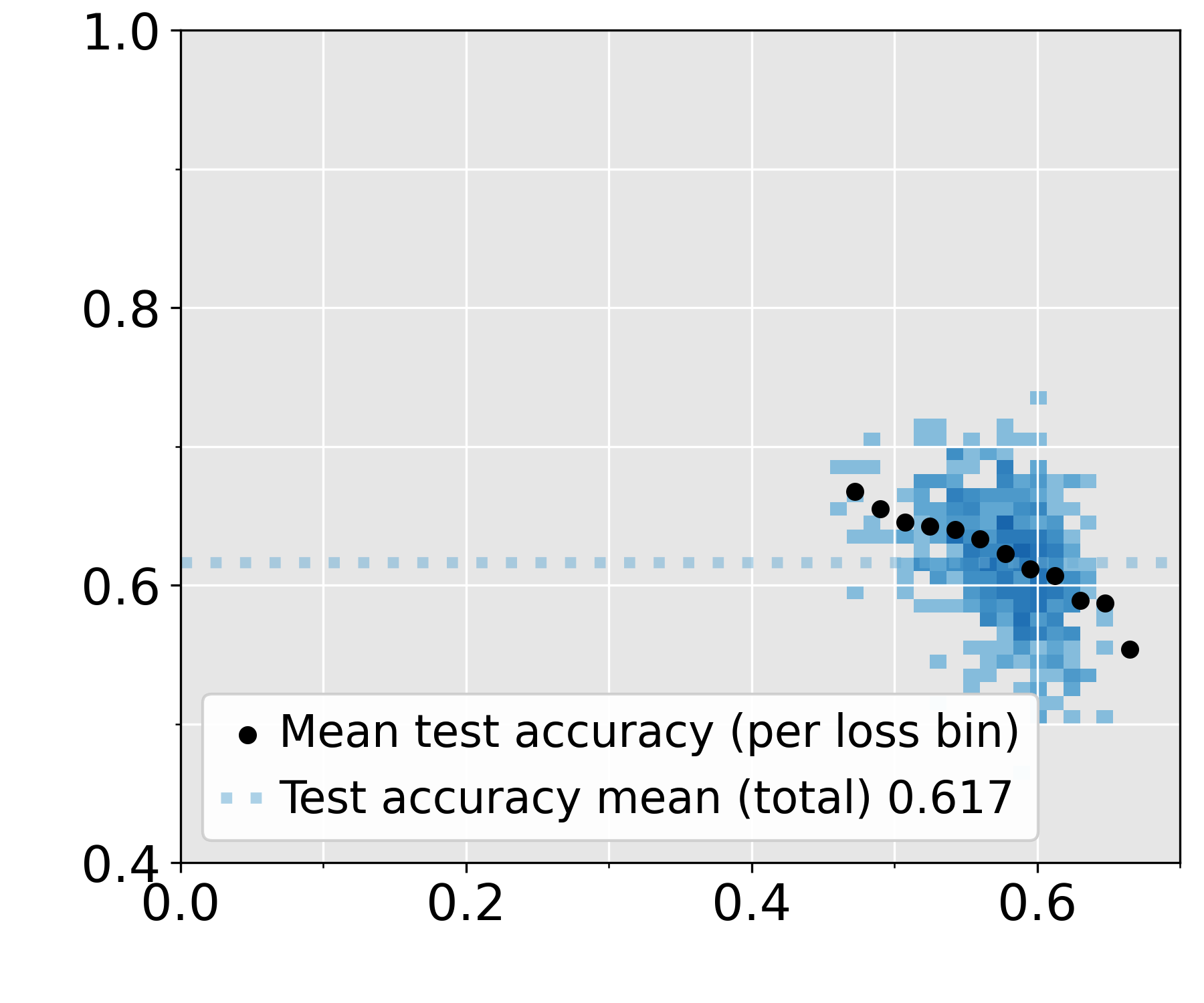}
    & \includegraphics{figures/different_loss_factor/colorbar.png} \\
    %%%%%%%%%%%%%%%%%%%%%%%%%%%%%%%%%%%%%%%%%%%%%%%%%%%%%%%%%
    % x axis labels
    %%%%%%%%%%%%%%%%%%%%%%%%%%%%%%%%%%%%%%%%%%%%%%%%%%%%%%%%%
    &
    &
    & \multicolumn{1}{c}{\hspace{1.5em}\textbf{\makecell{Train loss\\(normalized)}}}
    & \multicolumn{1}{c}{\hspace{1.5em}\textbf{\makecell{Train loss\\(normalized)}}}
    & \multicolumn{1}{c}{\hspace{1.5em}\textbf{\makecell{Train loss\\(normalized)}}}
    &
  \end{tabularx}
\caption{\textbf{Generalization of \colorsgd{\sgd{} (optimized)} versus \colorgnc{\gnc{} (randomly sampled)} in dependency of the prior on the weights $\Pr(W)$:} 
   We ``train'' $500$ models to $100\%$ train accuracy for $16$ training samples for different pairs of classes from the 
   \textbf{CIFAR10} dataset.
   \textbf{Rows~1-2:} classes \textit{bird} and \textit{ship}.
   \textbf{Rows~3-4:} classes \textit{Deer} and \textit{Truck}.
  Test accuracies and the normalized Lipschitz losses (see Sec.~\ref{sec:method}) for \gnc{} are similar across %uniform 
  initializations.
   \textbf{Column 1:} For $\mathcal{U}[-1, 1]$ initialization (used in \citet{chiang2022loss}), the Lipschitz normalized losses and the test accuracy of \sgd{} and \gnc{} are similar, except for the convergence of \sgd{} towards more low-margin solutions. 
  The claim in \citet{chiang2022loss} that \gnc{} resembles \sgd{}, conditional on average test accuracy in the normalized loss bin (black dots), is an artifact of the suboptimal convergence of \sgd{} caused by this initialization.
  % \textbf{Column 2:} For Kaiming uniform, \sgd{} (first and third rows) improves considerably both in terms of loss and accuracy.
  % In contrast, \gnc{} (second and fourth rows) remains unaffected, as it is independent of the weight scale in each layer.  
  % \textbf{Column 3:} For Kaiming Gaussian, there is an improvement for both \gnc{} and \sgd{}, but they still arrive at different losses and accuracies.
  \textbf{Columns 2-3:} For Kaiming initializations, \sgd{} (first and third rows) improves considerably both in terms of loss and accuracy.
  In contrast, \gnc{} (second and fourth rows) remains unaffected, as it is independent of the weight scale in each layer.
  }
  \label{fig:different_initializations_appendix2}
\end{figure*}

\begin{figure*}[ht]
  \setlength\tabcolsep{0pt}
  \adjustboxset{width=\linewidth,valign=c}
  \centering
  \begin{tabularx}{1.0\linewidth}{
  @{}l 
  S{p{0.00\textwidth}} 
  S{p{0.05\textwidth}} 
  *{4}{S{p{0.22245\textwidth}}} 
  S{p{0.052\textwidth}}}
    %%%%%%%%%%%%%%%%%%%%%%
    &
    &
    & \multicolumn{1}{c}{\textbf{\quad Uniform [-1, 1]}}
    & \multicolumn{1}{c}{\textbf{\quad Uniform [-0.2, 0.2]}}
    & \multicolumn{1}{c}{\textbf{\quad Kaiming Uniform}}
    & \multicolumn{1}{c}{\textbf{\quad Kaiming Gaussian}}
    & \multicolumn{1}{c}{} \\
    %%%%%%%%%%%%%%%%%%%%%%
    % SGD 1 
    %%%%%%%%%%%%%%%%%%%%%%%%
    &
    &\rotatebox[origin=c]{90}{\textbf{\makecell{\sgd{} - 4 samples\\Test accuracy}}}
    &\includegraphics{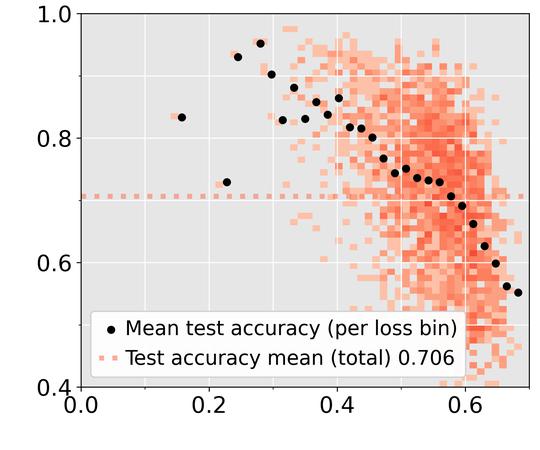}
    & \includegraphics{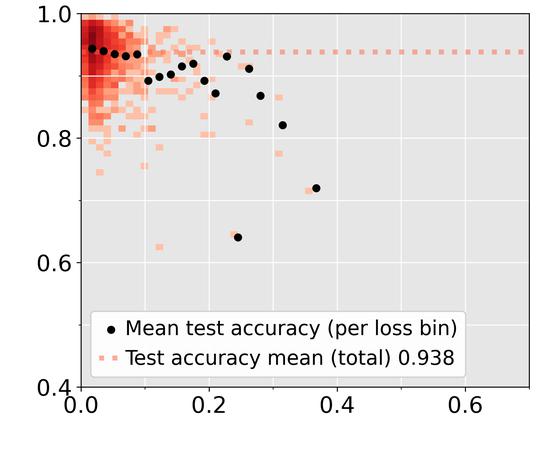}
    & \includegraphics{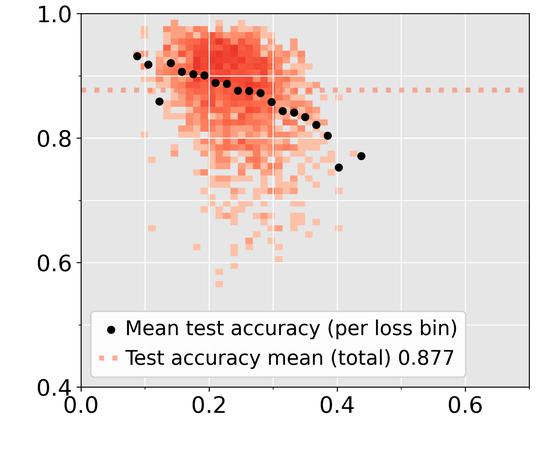}
    & \includegraphics{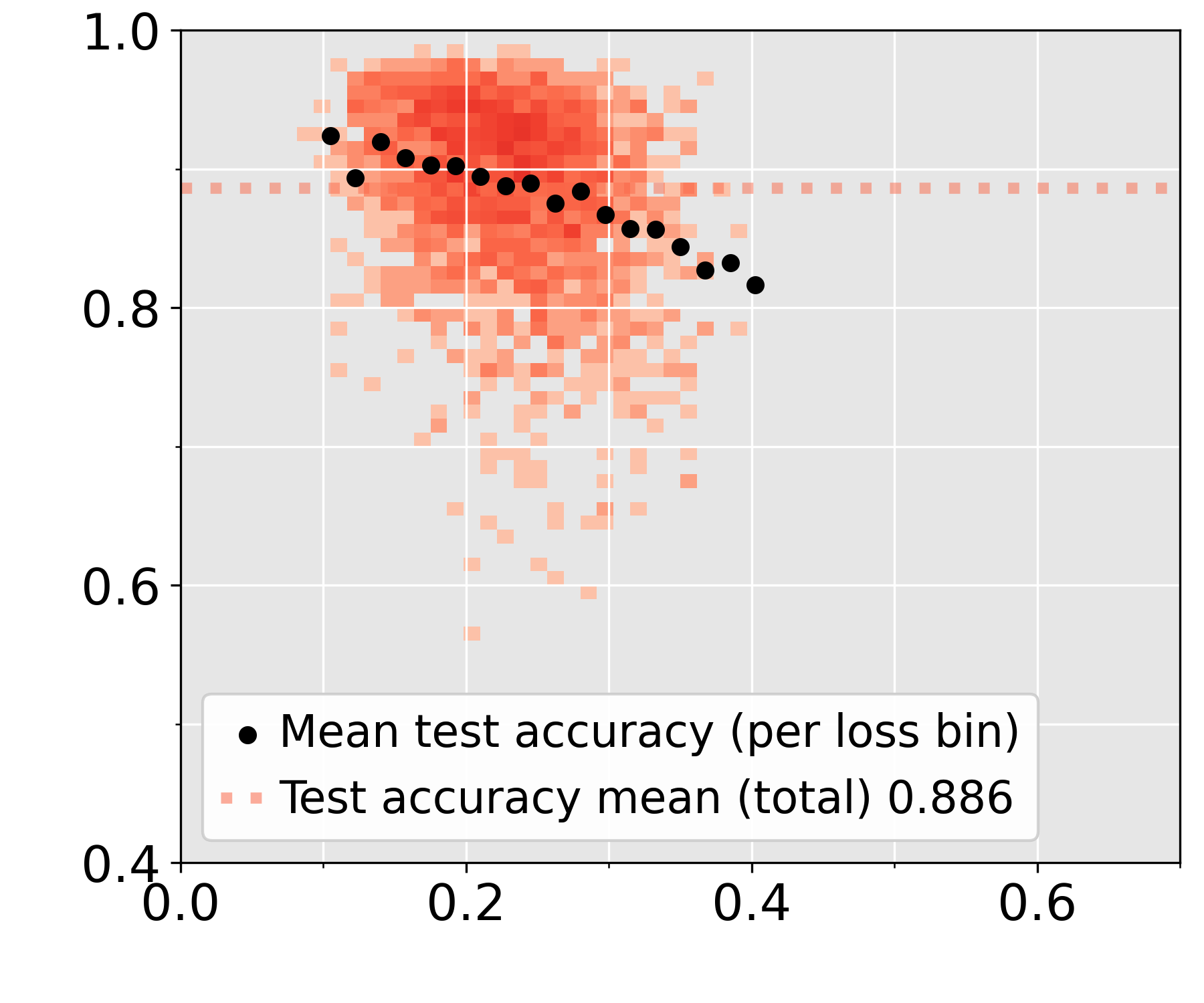}
    & \includegraphics{figures/different_loss_factor/colorbar_SGD.png} \\
    %%%%%%%%%%%%%%%%%%%%%%
    % GNC 1 
    %%%%%%%%%%%%%%%%%%%%%%%%
    &
    &\rotatebox[origin=c]{90}{\textbf{\makecell{\gnc{} - 4 samples\\Test accuracy}}}
    & \includegraphics{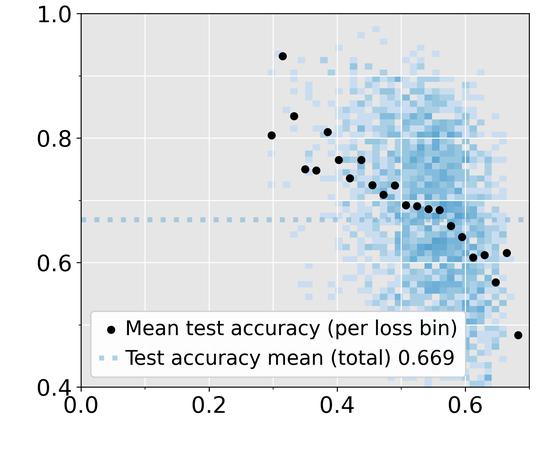}
    & \includegraphics{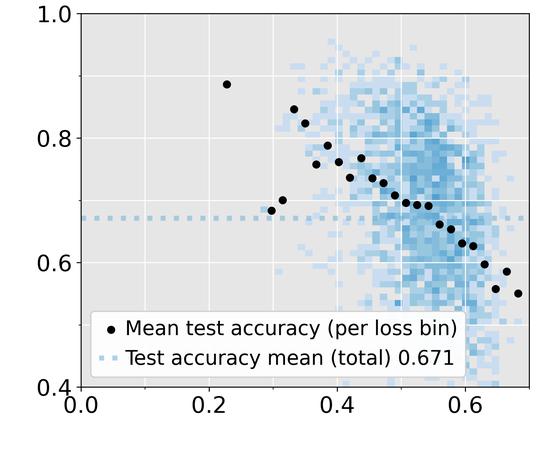}
    & \includegraphics{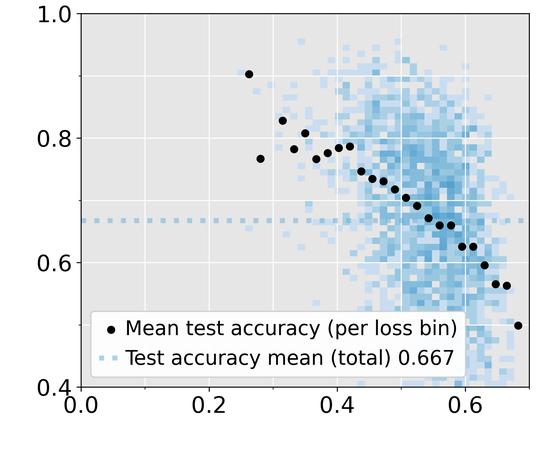}
    & \includegraphics{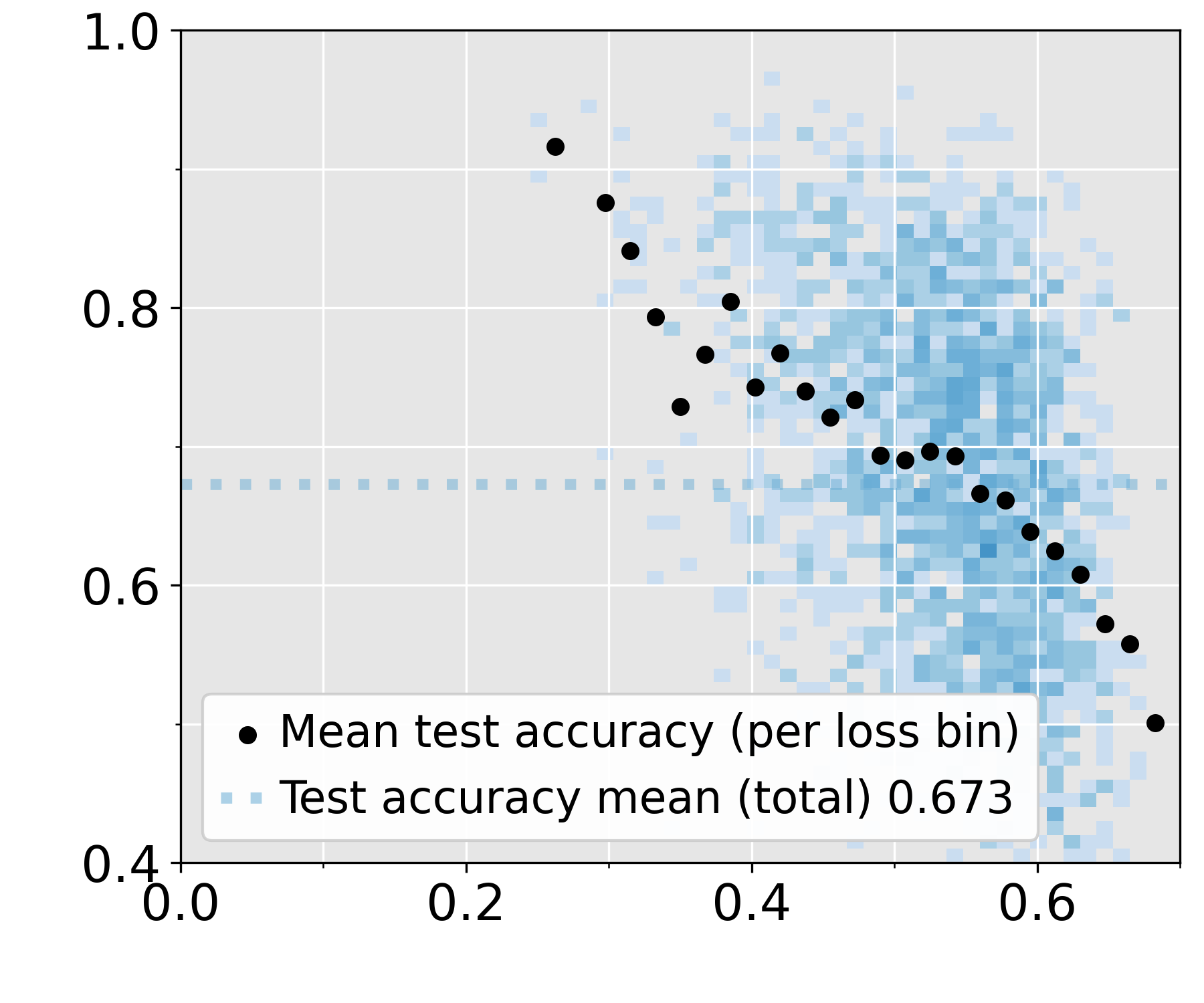}
    & \includegraphics{figures/different_loss_factor/colorbar.png} \\
        %%%%%%%%%%%%%%%%%%%%%%
    % SGD 2 
    %%%%%%%%%%%%%%%%%%%%%%%%
    &
    &\rotatebox[origin=c]{90}{\textbf{\makecell{\sgd{} - 32 samples\\Test accuracy}}}
    &\includegraphics{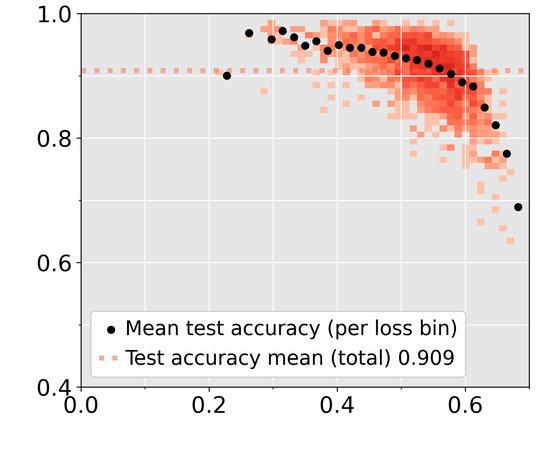}
    & \includegraphics{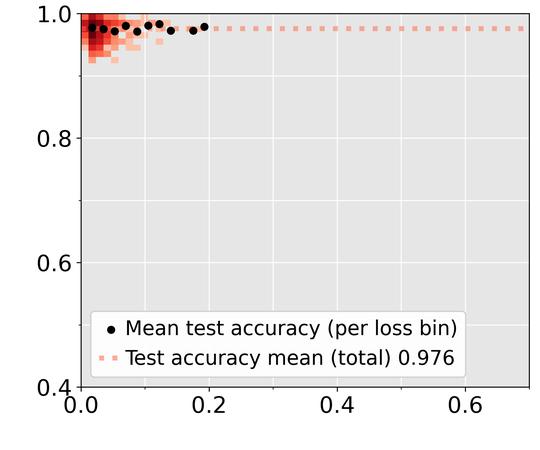}
    & \includegraphics{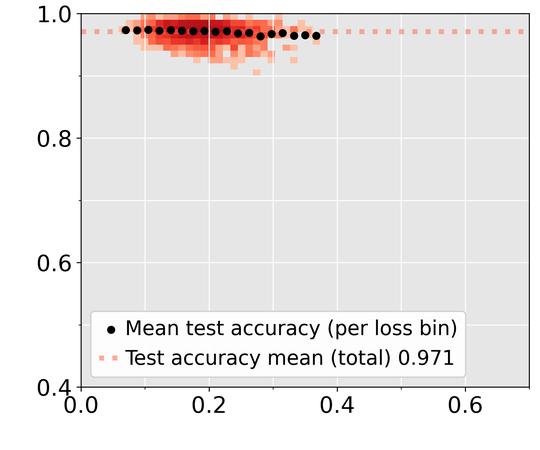}
    & \includegraphics{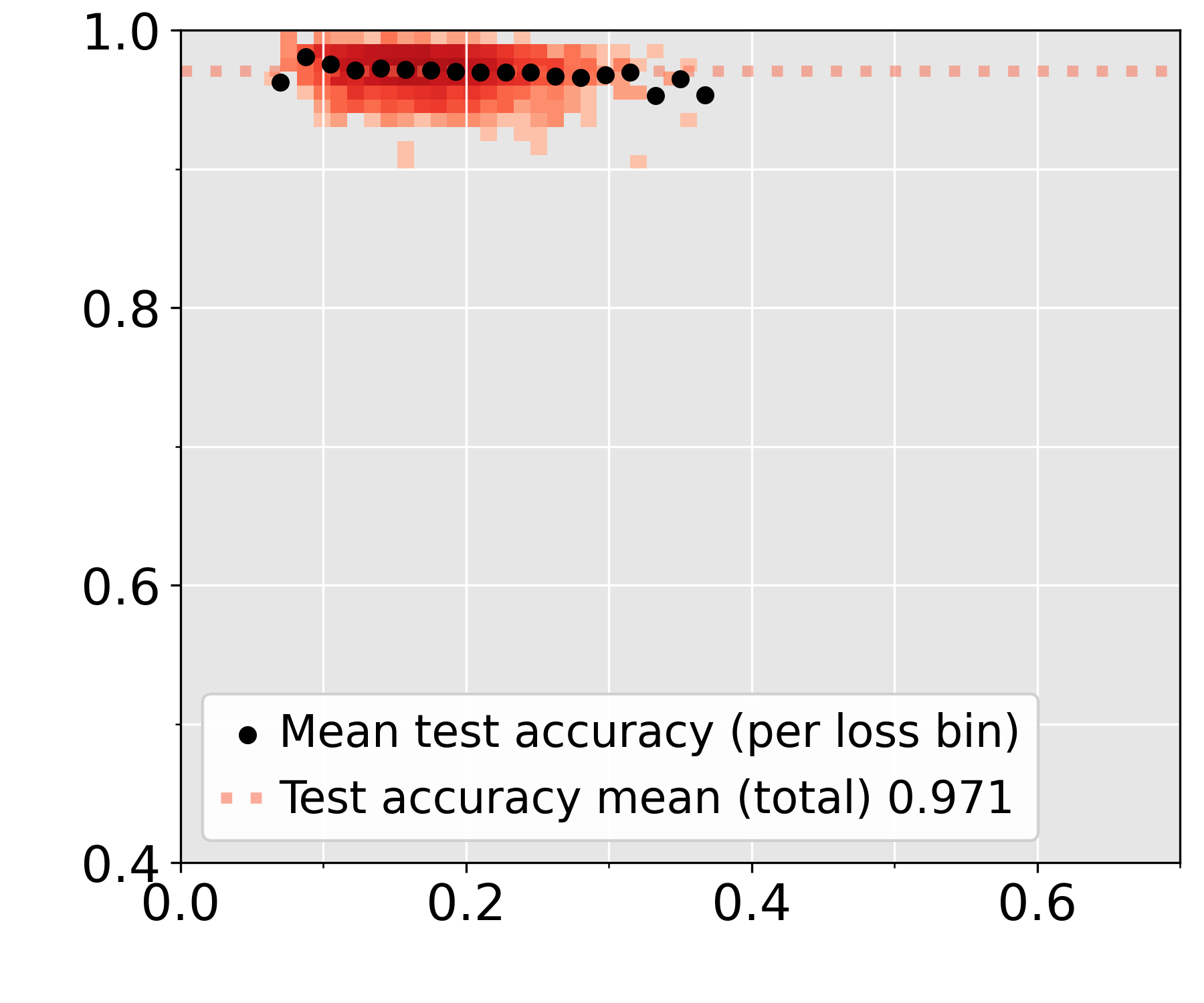}
    & \includegraphics{figures/different_loss_factor/colorbar_SGD.png} \\
    %%%%%%%%%%%%%%%%%%%%%%
    % GNC 2
    %%%%%%%%%%%%%%%%%%%%%%%%
    &
    &\rotatebox[origin=c]{90}{\textbf{\makecell{\gnc{} - 32 samples\\Test accuracy}}}
    & \includegraphics{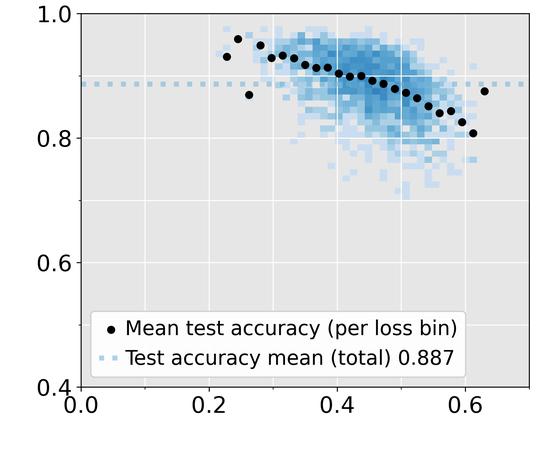}
    & \includegraphics{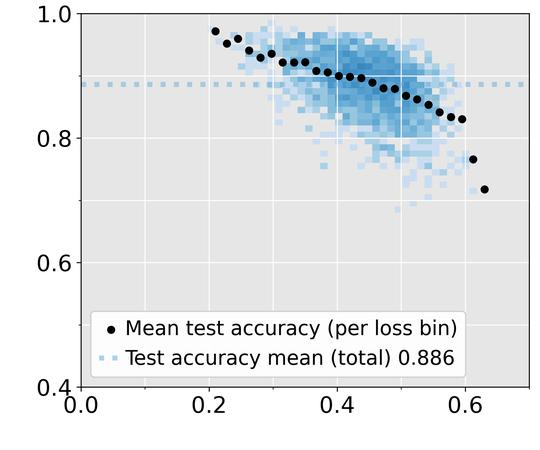}
    & \includegraphics{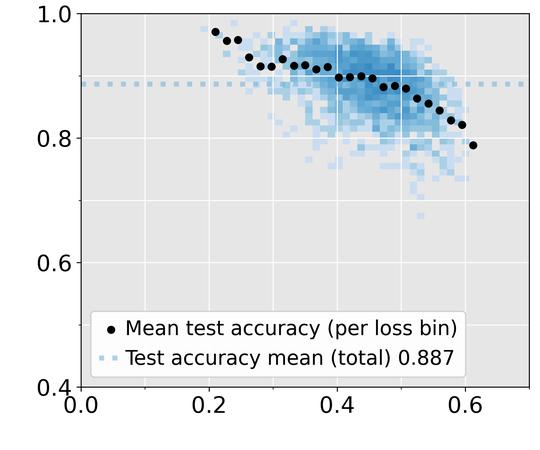}
    & \includegraphics{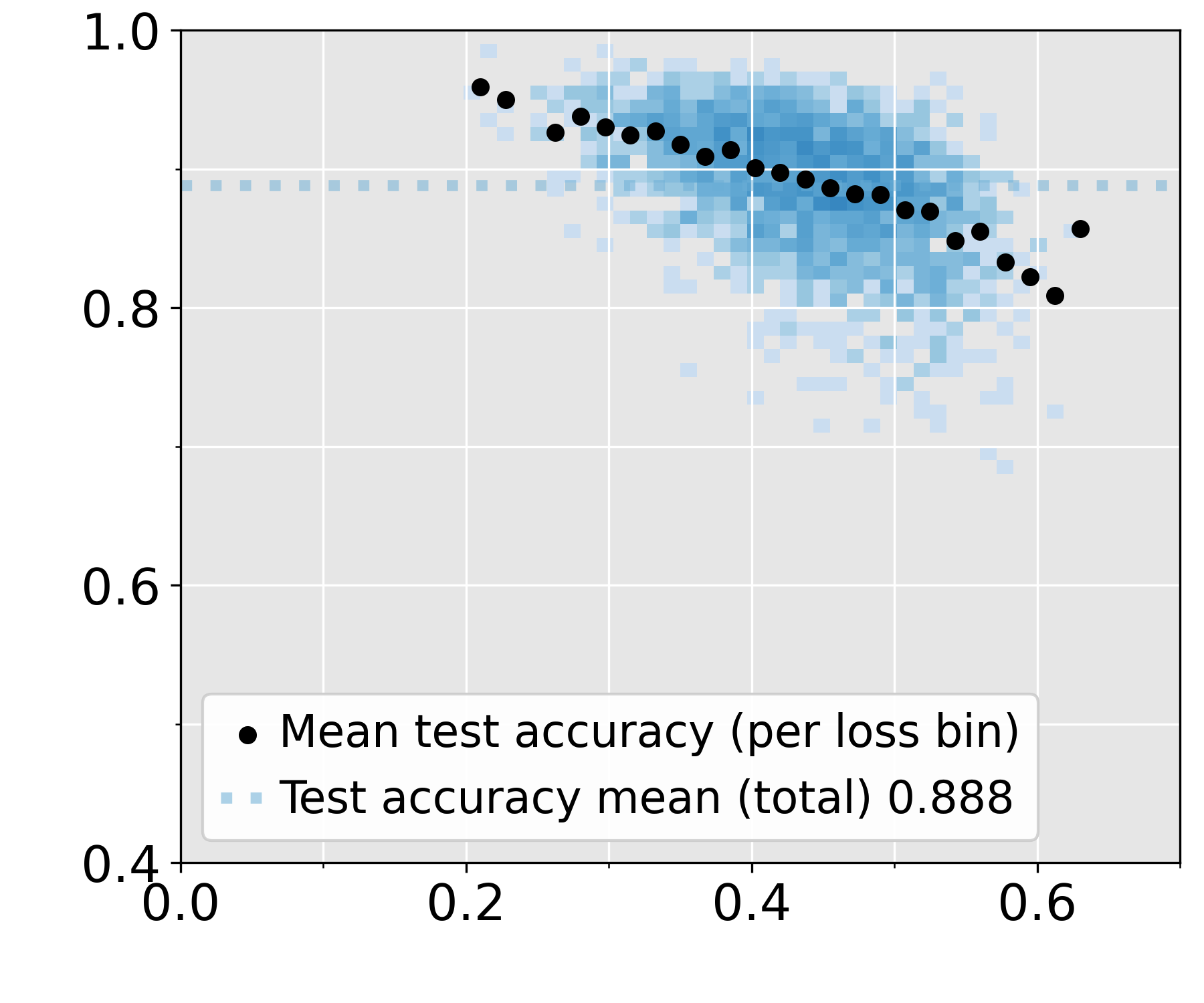}
    & \includegraphics{figures/different_loss_factor/colorbar.png} \\
    %%%%%%%%%%%%%%%%%%%%%%%%%%%%%%%%%%%%%%%%%%%%%%%%%%%%%%%%%
    % x axis labels
    %%%%%%%%%%%%%%%%%%%%%%%%%%%%%%%%%%%%%%%%%%%%%%%%%%%%%%%%%
    &
    &
    & \multicolumn{1}{c}{\hspace{1.5em} \textbf{\makecell{Train loss\\(normalized)}}}
    & \multicolumn{1}{c}{\hspace{1.5em} \textbf{\makecell{Train loss\\(normalized)}}}
    & \multicolumn{1}{c}{\hspace{1.5em} \textbf{\makecell{Train loss\\(normalized)}}}
    & \multicolumn{1}{c}{\hspace{1.5em} \textbf{\makecell{Train loss\\(normalized)}}}
    & \\
  \end{tabularx}
  \caption{\textbf{Generalization of \colorsgd{\sgd{} (optimized)} versus \colorgnc{\gnc{} (randomly sampled)} in dependency of the prior on the weights $\Pr(W)$:} 
    We ``train'' $2000$ models to $100\%$ train accuracy for $4$ and $32$ training samples from classes \emph{0} and \emph{7} of the MNIST dataset.
  Test accuracies for \gnc{} are similar across initializations, and the normalized Lipschitz loss (see Sec.~\ref{sec:method}) is similar across the uniform distributions.
  \textbf{Column 1:} For $\mathcal{U}[-1, 1]$ initialization (used in \citet{chiang2022loss}), the normalized Lipschitz losses and the test accuracy of \sgd{} and \gnc{} are similar, except for the convergence of \sgd{} towards more low margin solutions. 
  The claim in \citet{chiang2022loss} that \gnc{} resembles \sgd{}, conditional on average test accuracy in the normalized loss bin (black dots), is an artifact of the suboptimal convergence of \sgd{} caused by this initialization.
  \textbf{Columns 2-4:} For other initializations, \sgd{} (first and third rows) improves considerably both in terms of loss and accuracy.
  In contrast, \gnc{} remains unaffected, as it is independent of the weight scale in each layer.   
}
  \label{fig:different_initializations_weight_normalization_4_32_samples}
\end{figure*}

\begin{figure*}[ht]
  \setlength\tabcolsep{0pt}
  \adjustboxset{width=\linewidth,valign=c}
  \centering
  \begin{tabularx}{1.0\linewidth}{
  @{}l 
  S{p{0.00\textwidth}} 
  S{p{0.05\textwidth}} 
  *{4}{S{p{0.21196\textwidth}}} 
  S{p{0.051\textwidth}}}
    %%%%%%%%%%%%%%%%%%%%%%
    &
    &
    & \multicolumn{1}{c}{\textbf{\quad Uniform [-1, 1]}}
    & \multicolumn{1}{c}{\textbf{\quad Uniform [-0.2, 0.2]}}
    & \multicolumn{1}{c}{\textbf{\quad Kaiming Uniform}}
    & \multicolumn{1}{c}{\textbf{\quad Kaiming Gaussian}}
    & \multicolumn{1}{c}{} \\

    %%%%%%%%%%%%%%%%%%%%%%
    % SGD 2 
    %%%%%%%%%%%%%%%%%%%%%%%%
    &
    &\rotatebox[origin=c]{90}{\textbf{\makecell{\sgd{}\\Test accuracy}}}
    &\includegraphics{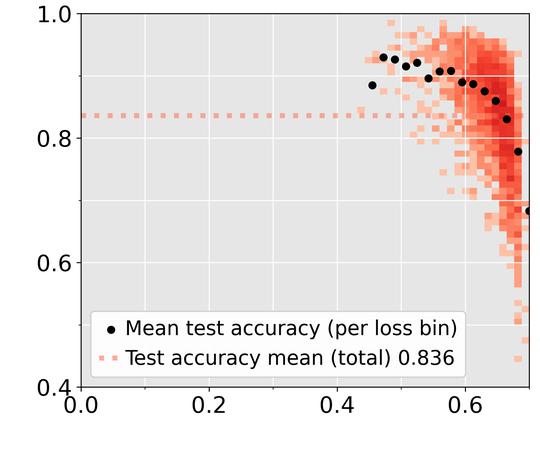}
    & \includegraphics{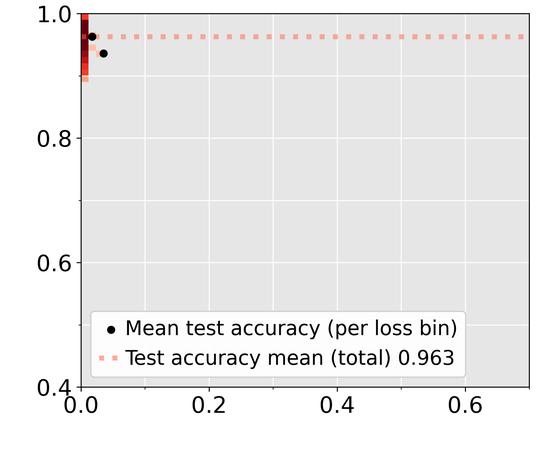}
    & \includegraphics{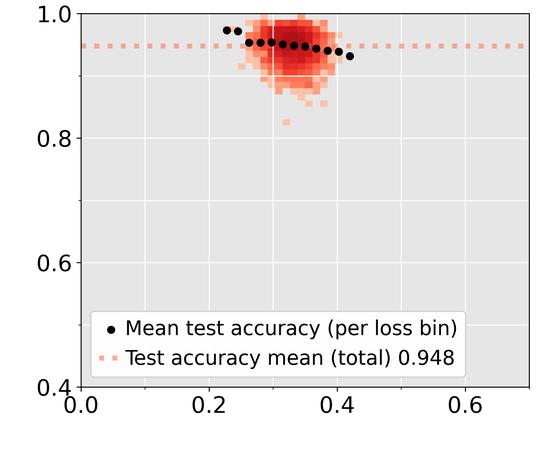}
    & \includegraphics{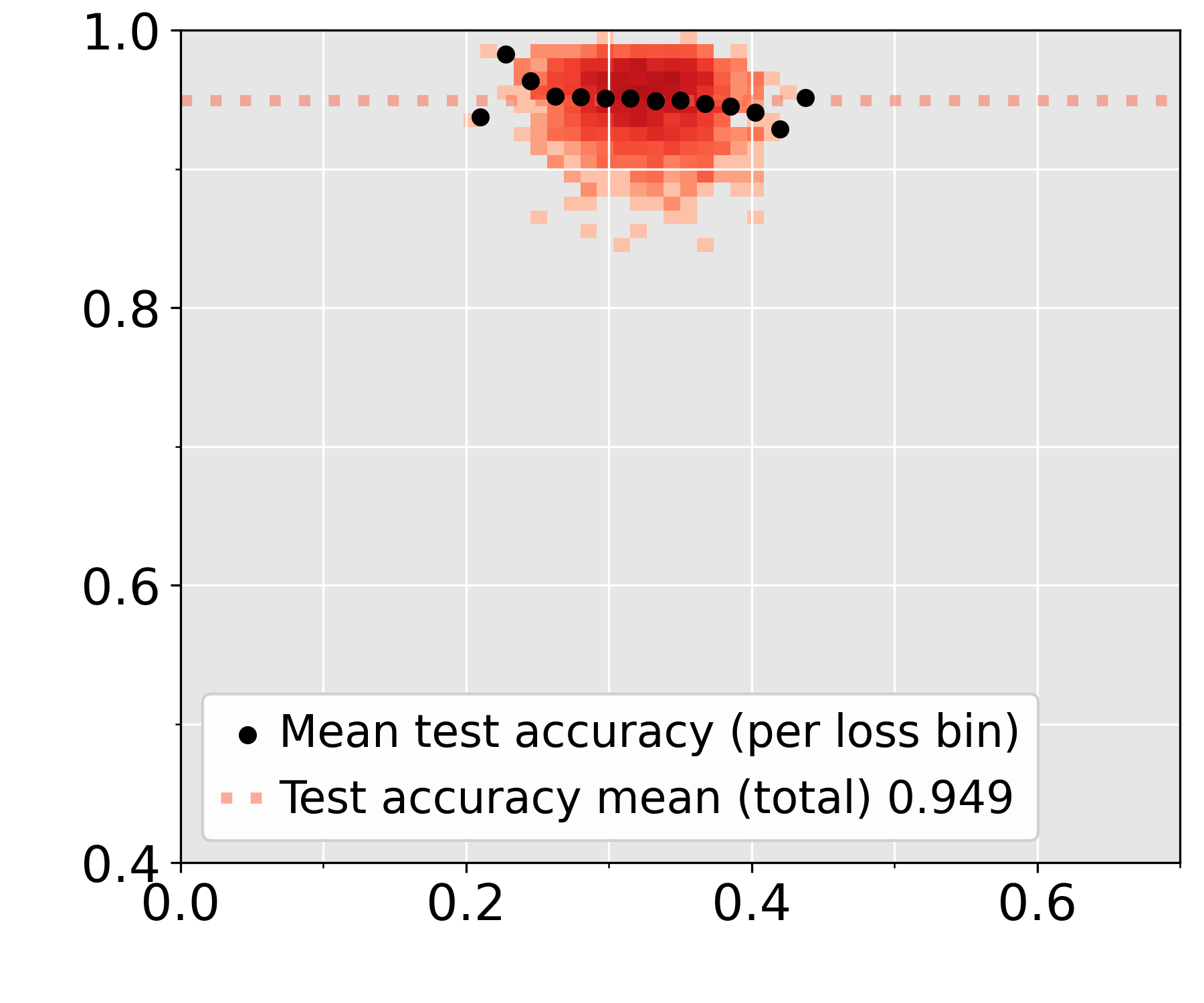}
    & \includegraphics{figures/different_loss_factor/colorbar_SGD.png} \\
    %%%%%%%%%%%%%%%%%%%%%%
    % GNC 2
    %%%%%%%%%%%%%%%%%%%%%%%%
    &
    &\rotatebox[origin=c]{90}{\textbf{\makecell{\gnc{}\\Test accuracy}}}
    & \includegraphics{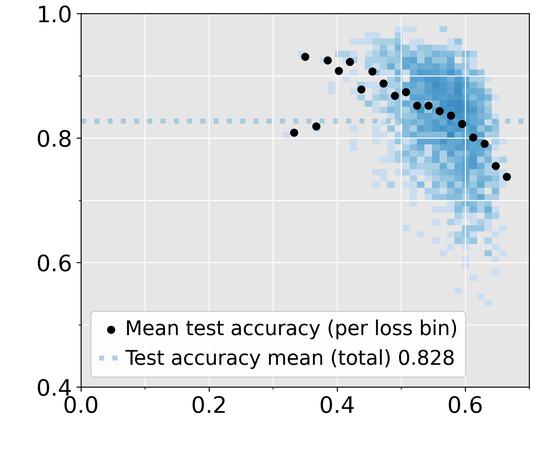}
    & \includegraphics{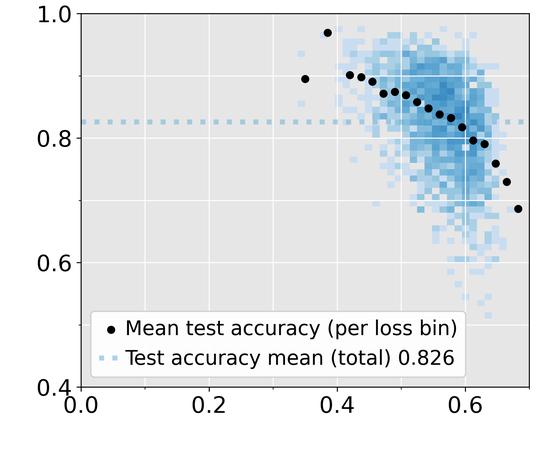}
    & \includegraphics{figures/different_loss_factor/2d_hist_train_loss_normalize_grad_input_test_acc_mnist_guess_initialization_kaiming_seed202_permseed202_16_samples.jpg}
    & \includegraphics{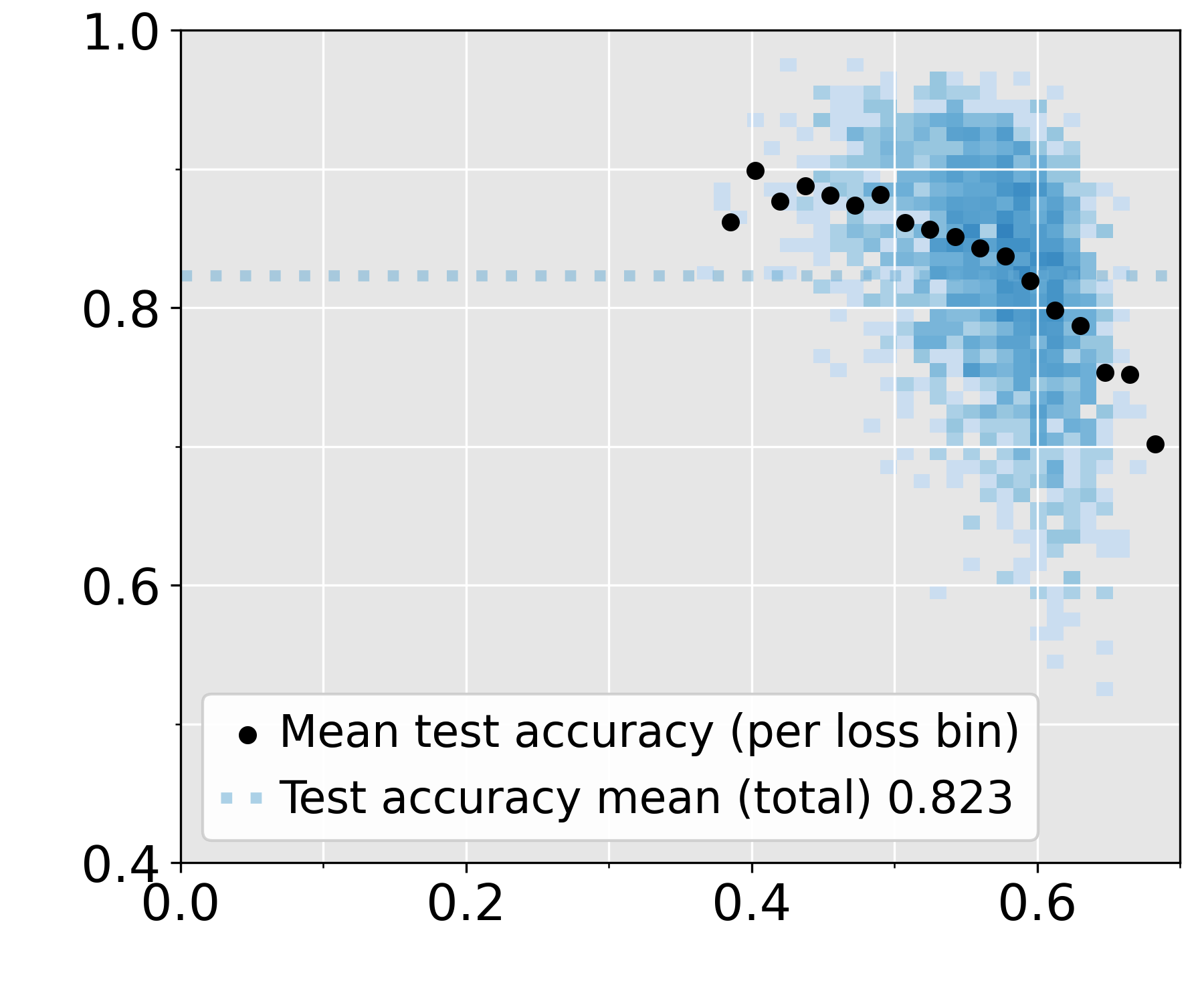}
    & \includegraphics{figures/different_loss_factor/colorbar.png} \\
    %%%%%%%%%%%%%%%%%%%%%%%%%%%%%%%%%%%%%%%%%%%%%%%%%%%%%%%%%
    % x axis labels
    %%%%%%%%%%%%%%%%%%%%%%%%%%%%%%%%%%%%%%%%%%%%%%%%%%%%%%%%%
    &
    &
    & \multicolumn{1}{c}{\hspace{0.5em} \textbf{\makecell{Train loss\\(weight norm)}}}
    & \multicolumn{1}{c}{\hspace{0.5em} \textbf{\makecell{Train loss\\(weight norm)}}}
    & \multicolumn{1}{c}{\hspace{0.5em} \textbf{\makecell{Train loss\\(weight norm)}}}
    & \multicolumn{1}{c}{\hspace{0.5em} \textbf{\makecell{Train loss\\(weight norm)}}}
    & \multicolumn{1}{c}{} \\
  \end{tabularx}
  \caption{\textbf{Generalization of \colorsgd{\sgd{} (optimized)} versus \colorgnc{\gnc{} (randomly sampled)} in dependency of the prior on the weights $\Pr(W)$:} 
    We ``train'' $2000$ models to $100\%$ train accuracy for $16$ training samples from classes \emph{0} and \emph{7} of the MNIST dataset.
  Test accuracies for \gnc{} are similar across initializations, and the weight normalized loss (see Sec.~\ref{sec:method}) is similar across the uniform distributions.
  \textbf{Column 1:} For $\mathcal{U}[-1, 1]$ initialization (used in \citet{chiang2022loss}), the weight normalized losses and the test accuracy of \sgd{} and \gnc{} are similar, except for the convergence of \sgd{} towards more low margin solutions. 
  The claim in \citet{chiang2022loss} that \gnc{} resembles \sgd{}, conditional on average test accuracy in the normalized loss bin (black dots), is an artifact of the suboptimal convergence of \sgd{} caused by this initialization.
  \textbf{Columns 2-4:} For other initializations, \sgd{} (first row) improves considerably both in terms of loss and accuracy.
  In contrast, \gnc{} remains unaffected, as it is independent of the weight scale in each layer.   
  \textbf{This graph uses the weight norm normalization \cite{chiang2022loss}.}
}
  \label{fig:different_initializations_weight_normalization}
\end{figure*}

\begin{figure}[htb]
    \centering
    \begin{subfigure}{0.3\linewidth}
    \includegraphics[width=\linewidth]{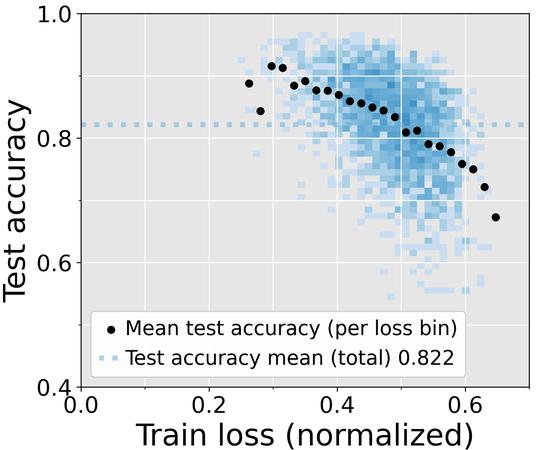}
    \caption{Train loss vs. Test accuracy of \gnc{} initialized with Kaiming uniform distribution. 
    }
    \label{fig:initialization_loss_curves}
    \end{subfigure}
        \hspace{1em}
    \begin{subfigure}{0.3\linewidth}
        \includegraphics[width=\linewidth]
        {figures/different_loss_factor/2d_hist_train_loss_normalize_grad_input_test_acc_mnist_sgd_initialization_uniform_lr001_epoch60_seed202_permseed202_16_samples}

        \caption{Train loss vs. Test accuracy of \sgd{} initialized with $\cU[-1,1]$.\\}
        
    \end{subfigure}
    \hspace{1em}
    \begin{subfigure}{0.3\linewidth}
    \includegraphics[width=\linewidth]{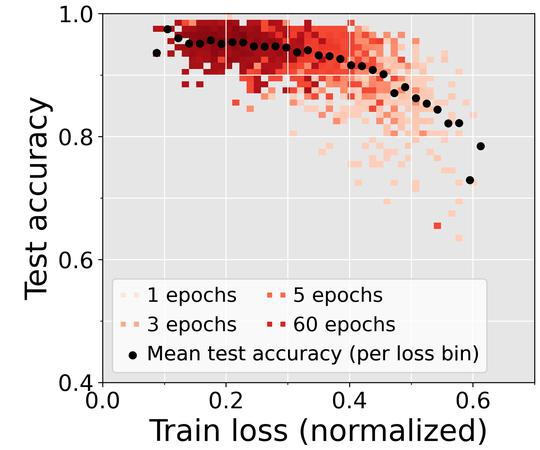}
        \caption{Train loss vs. Test accuracy of \sgd{} with Kaiming uniform initialization along different epochs.}
        
    \end{subfigure}

    \caption{Train loss vs. Test accuracy of \sgd{} with Kaiming uniform initialization along different epochs next to \gnc{}. This configuration of \sgd{} resembles the one from $\cU[-1,1]$, as models from earlier epochs still have not converged. In spite of that, the results of \sgd{} are better than the ones of \gnc{} given a loss bin, and on average.}
    \label{fig:sgd_epochs_convergence} 
\end{figure}

\begin{figure}[htb]
    \centering
    \begin{subfigure}{.4\linewidth}
        \includegraphics[width=\linewidth]{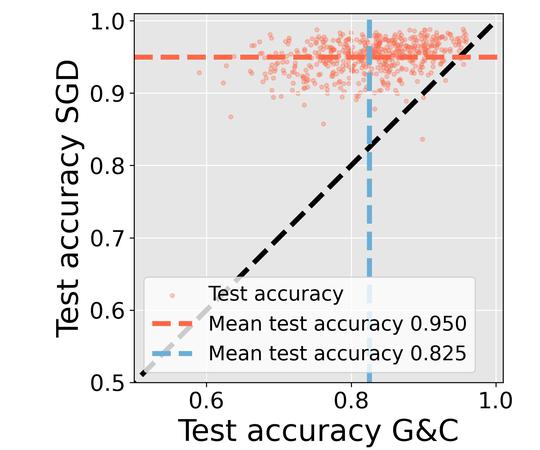}
        
        \caption{}
        \label{fig:sgd_from_gnc}
    \end{subfigure}
    \begin{subfigure}{.4\linewidth}
        \includegraphics[width=\linewidth]{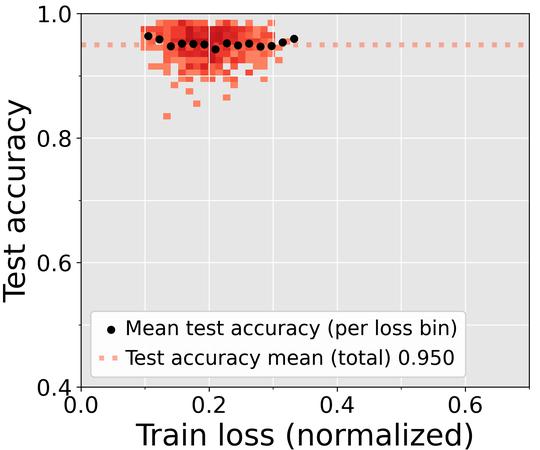}
        
        \caption{}
        \label{fig:sgd_from_gnc_hist}
    \end{subfigure}
    \caption{
    \textbf{Initialization of \sgd{} with \gnc{} networks does not improve generalization.} We initialize \sgd{} with 500 models obtained by \gnc{} (randomly sampled networks from Kaiming uniform with 100\% training accuracy) for MNIST with classes \emph{0} and \emph{7} and 16 training samples. \textbf{Left}: Most models improve through optimization (above the black line). Thus, \sgd{} converges to a smaller volume of perfectly fitting networks that generalize better, resulting in a mean test accuracy of $95\%$ for \sgd{} instead of $82.5\%$ for \gnc{}. \textbf{Right}: distribution of test accuracy versus normalized loss is similar to \sgd{} with Kaiming uniform initialization (Figure~\ref{fig:different_initializations}). Thus, initializing with \gnc{} is not better than random initialization.}
    \label{fig:sgd_from_gnc_all}
\end{figure}

First, we show in Figure~\ref{fig:initialization_gradients} the norm of the gradient of the weights along the training process for different initializations.
This figure illustrates that SGD essentially stops the optimization early in the training
process with sub-optimal initialization.
Next, we show the result of Figure~\ref{fig:different_initializations} for another pair of classes from the MNIST dataset (Figure~\ref{fig:different_initializations_appendix1}), two pairs of classes from the CIFAR10 dataset (Figure~\ref{fig:different_initializations_appendix2}), and different numbers of samples (Figure~\ref{fig:different_initializations_weight_normalization_4_32_samples}).
In Figure~\ref{fig:different_initializations_weight_normalization}, we show the same result as in Figure~\ref{fig:different_initializations}, but with the \emph{weight normalized loss} taken from \cite{chiang2022loss}.
Figure~\ref{fig:sgd_epochs_convergence} shows the loss evolution along different epochs of \sgd{}, which was initialized with Kaiming uniform distribution. 
Finally, we show in Figure~\ref{fig:sgd_from_gnc_all} the complementary experiments of the discussion in Section~\ref{sec:init} regarding the initialization of \sgd{} with already ``trained" networks of \gnc{}.
\setlength\cellspacetoplimit{0pt}
\setlength\cellspacebottomlimit{0pt}
\renewcommand\tabularxcolumn[1]{>{\centering\arraybackslash}S{p{#1}}}
% 
%%%%%%%%%%%%%%%%%%%%%%%%%
% seed 204
%%%%%%%%%%%%%%%%%%%%%%%%%
% 
% 
%%%%%%%%%%%%%%%%%%%%%%%%%
% seed 219
%%%%%%%%%%%%%%%%%%%%%%%%%
%%%%%%%%% 4 and 32 samples
\setlength\cellspacetoplimit{0pt}
\setlength\cellspacebottomlimit{0pt}
\renewcommand\tabularxcolumn[1]{>{\centering\arraybackslash}S{p{#1}}}

%%%%%%%% weight norm

\newpage
\section{Overparameterization with Increasing Width}
\label{sec:width-appendix}
In the following section, we expand the experiments of the network \emph{width} analysis from section~\ref{sec:width}.

The parameter counts for the LeNet architecture across various \emph{width} values are detailed for the MNIST dataset in Table~\ref{tab:width_parameters} and the CIFAR10 dataset in Table~\ref{tab:width_parameters_cifar10}. 
Note that the number of parameters in the linear layers depends on the dimension of the input.
The number of convolutions and neurons is determined by multiplying the number at width 1 by the width factor and rounding down the result.
The notation $\nicefrac{1}{6}^*$ indicates a width factor of $\nicefrac{1}{6}$ for the convolution layers and a width factor of $\nicefrac{1}{24}$ for the fully connected layers, given that there is already a single convolution in the first layer when using width factor $\nicefrac{1}{6}$.

\begin{figure*}[ht!]
\centering
\setlength\tabcolsep{0pt}
\adjustboxset{width=\linewidth,valign=c}
\centering
\begin{tabularx}{1.0\linewidth}{
@{}l 
S{p{0.022\textwidth}} 
S{p{0.047\textwidth}} 
*{4}{S{p{0.22\textwidth}}} 
S{p{0.051\textwidth}}}
    %%%%%%%%%%%%%%%%%%%%%%%%%%
    &
    &
    & \multicolumn{1}{c}{\textbf{\quad Width} $\nicefrac{\mathbf{1}}{\mathbf{6^*}}$}
    & \multicolumn{1}{c}{\quad $\nicefrac{\mathbf{1}}{\mathbf{6}}$}
    & \multicolumn{1}{c}{\quad $\nicefrac{\mathbf{2}}{\mathbf{6}}$}
    & \multicolumn{1}{c}{\quad $\nicefrac{\mathbf{4}}{\mathbf{6}}$}\\
    %%%%%%%%%%%%%%%%%%%%%%%%%%%%%%%%%%%
    % GNC 1
    %%%%%%%%%%%%%%%%%%%%%%%%%%%%%%%%%%%
    &\multirow{2}{*}[-3em]{\rotatebox{90}{\textbf{\gnc}}}
    &\rotatebox[origin=c]{90}{\quad     {\footnotesize \textbf{\makecell{Test accuracy vs\\ weight norm. loss}}}}
    &\includegraphics{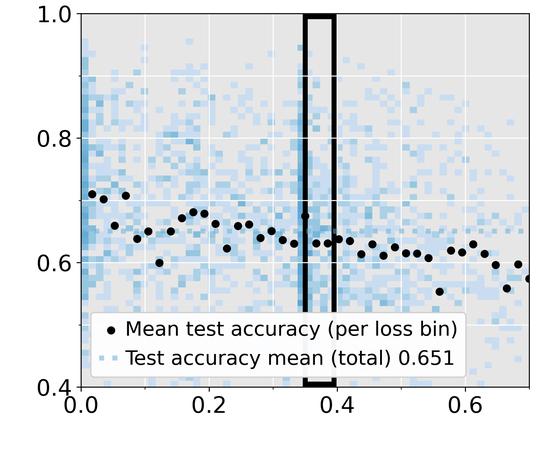}
    &\includegraphics {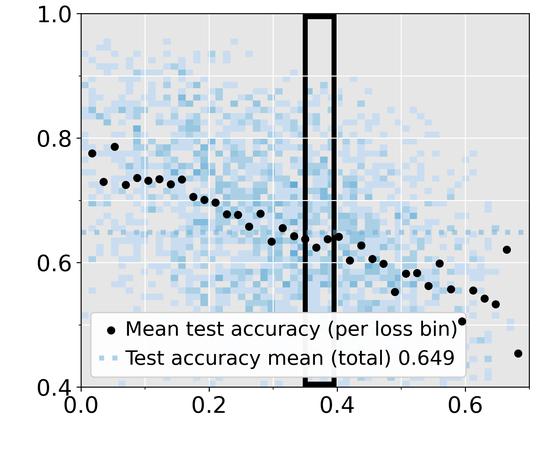}
    &\includegraphics{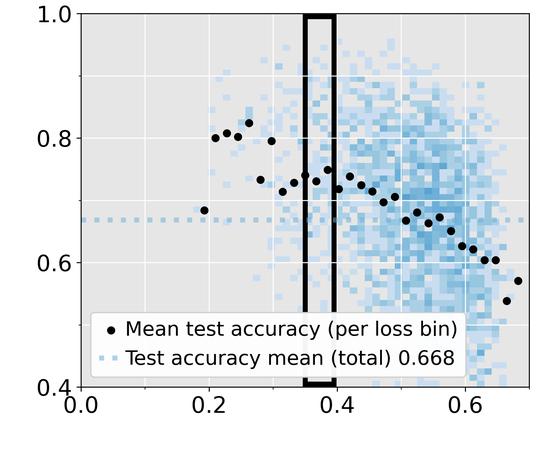}
    &\includegraphics{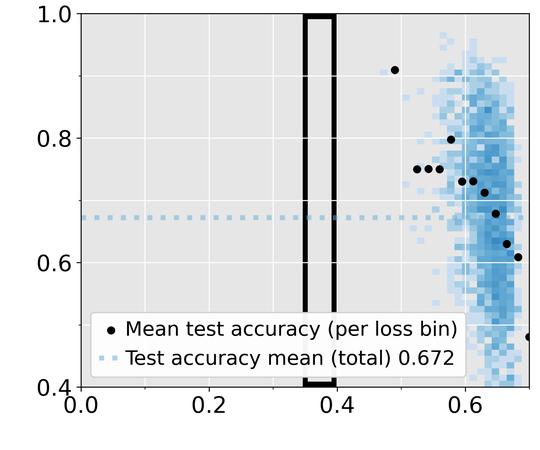}
    &\includegraphics{figures/different_loss_factor/colorbar.png} \\
     %%%%%%%%%%%%%%%%%%%%%%%%%%%%%%%%%%%
    % GNC 2
    %%%%%%%%%%%%%%%%%%%%%%%%%%%%%%%%%%%
    &
    &\rotatebox[origin=c]{90}    {\footnotesize  \enspace \textbf{\makecell{Test accuracy vs \\ Lipschitz norm. loss}}}
    &\includegraphics{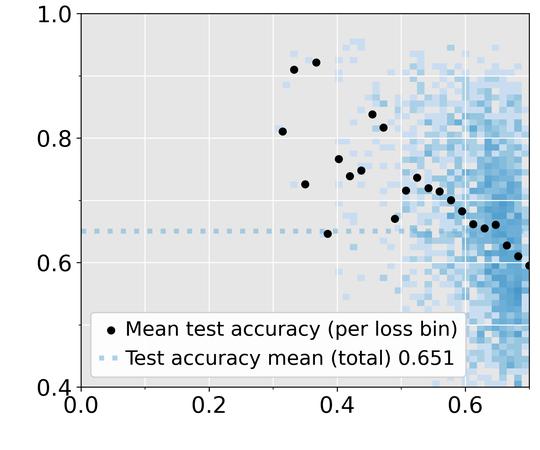}
    &\includegraphics {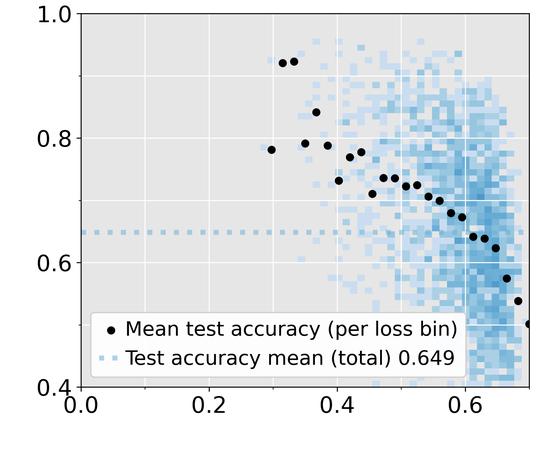}
    &\includegraphics{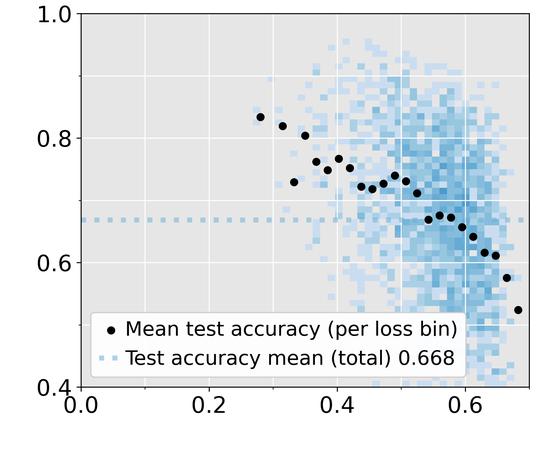}
    &\includegraphics{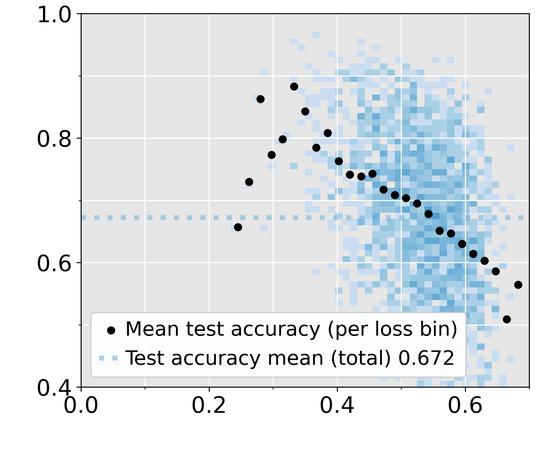}
    &\includegraphics{figures/different_loss_factor/colorbar.png} \\
    %%%%%%%%%%%%%%%%%%%%%%%%%%%%%%%%%%%
    % SGD 1
    %%%%%%%%%%%%%%%%%%%%%%%%%%%%%%%%%%%
    &\multirow{2}{*}[-3em]{\rotatebox{90}{\textbf{\sgd}}}
    &\rotatebox[origin=c]{90}{\quad     {\footnotesize \textbf{\makecell{Test accuracy vs\\ weight norm. loss}}}}
    & \includegraphics{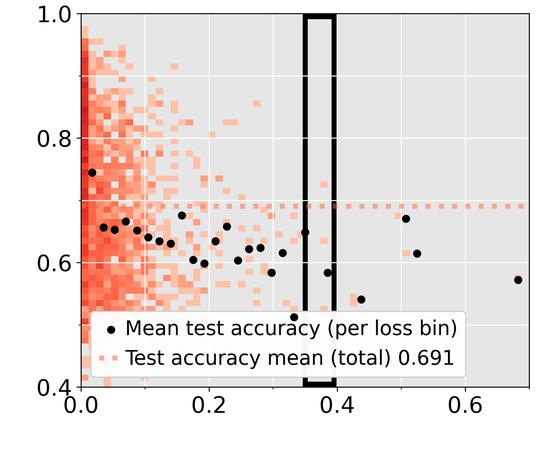}
    &\includegraphics{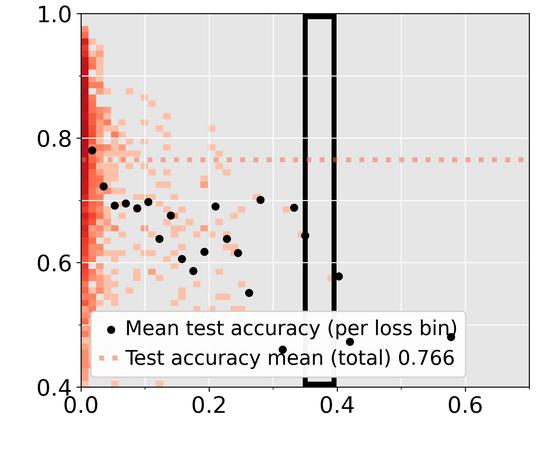}
    &\includegraphics{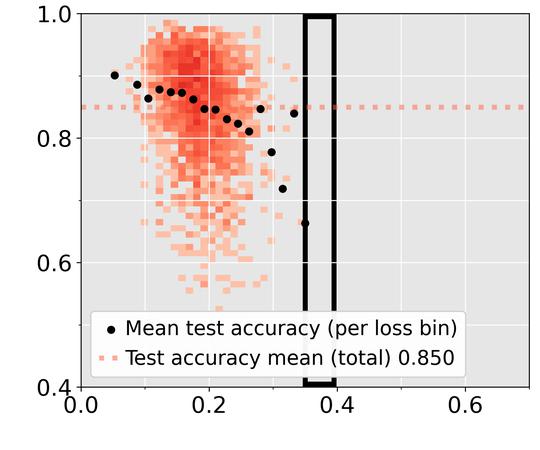}
    &\includegraphics{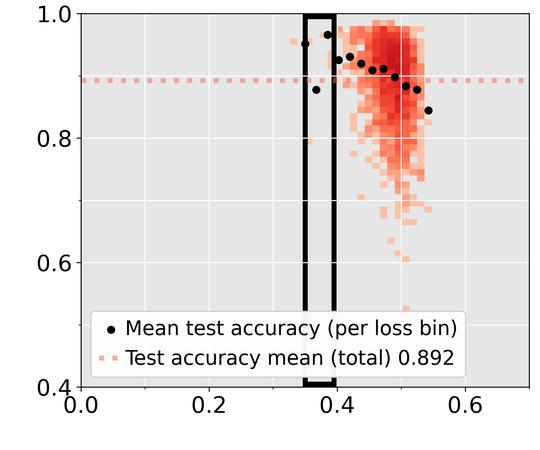}
    &\includegraphics{figures/different_loss_factor/colorbar_SGD.png} \\
    %%%%%%%%%%%%%%%%%%%%%%%%%%%%%%%%%%%
    % SGD 2
    %%%%%%%%%%%%%%%%%%%%%%%%%%%%%%%%%%%
    &
    &\rotatebox[origin=c]{90}    {\footnotesize  \enspace \textbf{\makecell{Test accuracy vs \\ Lipschitz norm. loss}}}
    &\includegraphics{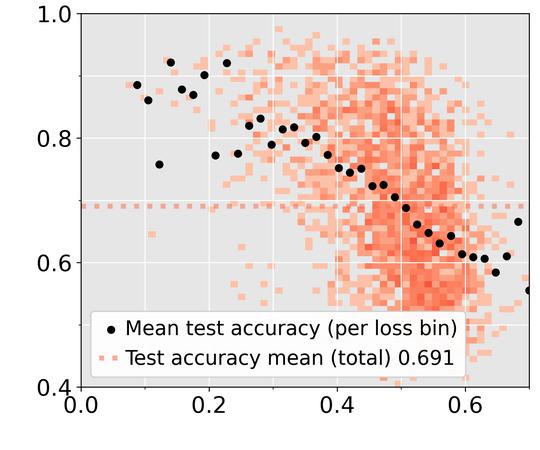}
    &\includegraphics{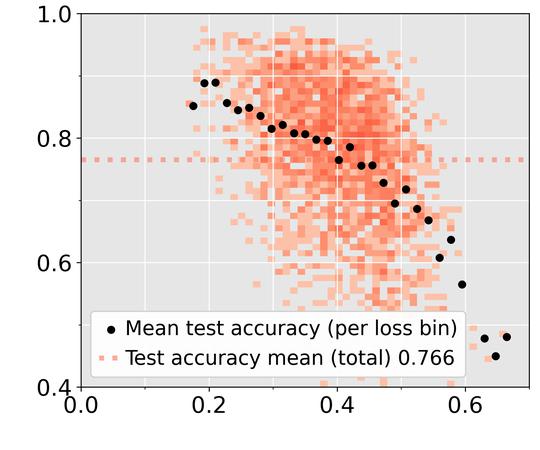}
    &\includegraphics{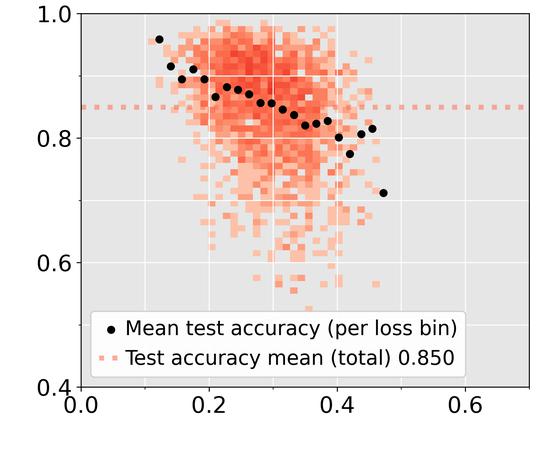}
    &\includegraphics{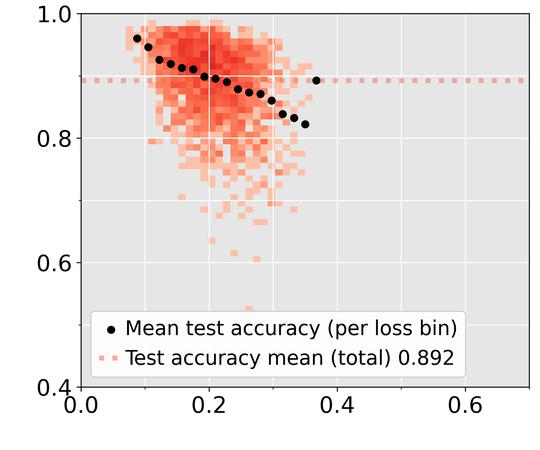}
    &\includegraphics{figures/different_loss_factor/colorbar_SGD.png} \\
    %%%%%%%%%%%%%%%%%%%%%%%%%%%%%%%%%
    &
    &
    & \multicolumn{1}{c}{\hspace{1.5em}\textbf{\makecell{Train loss\\(normalized)}}}
    & \multicolumn{1}{c}{\hspace{1.5em}\textbf{\makecell{Train loss\\(normalized)}}}
    & \multicolumn{1}{c}{\hspace{1.5em}\textbf{\makecell{Train loss\\(normalized)}}}
    & \multicolumn{1}{c}{\hspace{1.5em}\textbf{\makecell{Train loss\\(normalized)}}}
    &  \\
    %%%%%%%%%%%%%%%%%%%%%%%%%%
    \end{tabularx}
    \vspace{-0.1cm}
    \caption{
    \textbf{Analysis of overparameterization when increasing the width.}
    Test accuracy vs weight normalized loss \eqref{eq:weightnorm} of \citet{chiang2022loss} and our Lipschitz normalized loss \eqref{eq:lipschitznorm estimate} of {\color{amaranth}{\textbf{\sgd}}} and {\color{azure}{\textbf{\gnc}}} for classes \emph{0} \& \emph{7} of MNIST and 4 training samples.
    \textbf{Row 4:} Widening the networks enhances 
    geometric margin and average test accuracy for \sgd{}, while for \gnc{} \textbf{(Row 2)}, the margin and average test accuracy improve only slightly. This suggests that the improvement is mainly due to the bias of \sgd{} and not due to an architectural bias (see Fig.~\ref{fig:different_widths}). \textbf{Rows 1 and 3:} \citet{chiang2022loss} compare networks conditional on the (weight) normalized loss bin (illustrated by black boxes), which led them to conclude that \gnc{} improves with increasing width.
    With our Lischitz normalized loss, one would arrive at the opposite conclusion, which shows the problem of normalization.
    }
    \label{fig:different_widths_loss_with_nornalization_4_samples}
    \vspace{-0.1cm}
\end{figure*}

\begin{figure*}[ht!]
\centering
\setlength\tabcolsep{0pt}
\adjustboxset{width=\linewidth,valign=c}
\centering
\begin{tabularx}{1.0\linewidth}{
@{}l 
S{p{0.022\textwidth}} 
S{p{0.047\textwidth}} 
*{4}{S{p{0.22\textwidth}}} 
S{p{0.051\textwidth}}}
    %%%%%%%%%%%%%%%%%%%%%%%%%%
    &
    &
    & \multicolumn{1}{c}{\textbf{\quad Width} $\nicefrac{\mathbf{1}}{\mathbf{6^*}}$}
    & \multicolumn{1}{c}{\quad $\nicefrac{\mathbf{1}}{\mathbf{6}}$}
    & \multicolumn{1}{c}{\quad $\nicefrac{\mathbf{2}}{\mathbf{6}}$}
    & \multicolumn{1}{c}{\quad $\nicefrac{\mathbf{4}}{\mathbf{6}}$}\\    %%%%%%%%%%%%%%%%%%%%%%%%%%%%%%%%%%%
    % GNC 1
    %%%%%%%%%%%%%%%%%%%%%%%%%%%%%%%%%%%
    &\multirow{2}{*}[-3em]{\rotatebox{90}{\textbf{\gnc}}}
    &\rotatebox[origin=c]{90}{\quad     {\footnotesize
    \textbf{\makecell{Test accuracy vs\\ weight norm. loss}}}}
    &\includegraphics{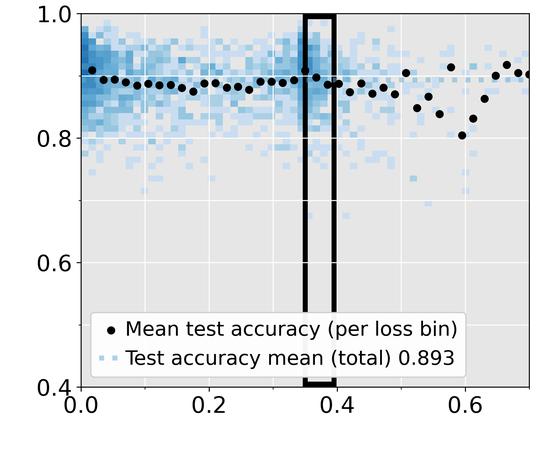}
    &\includegraphics {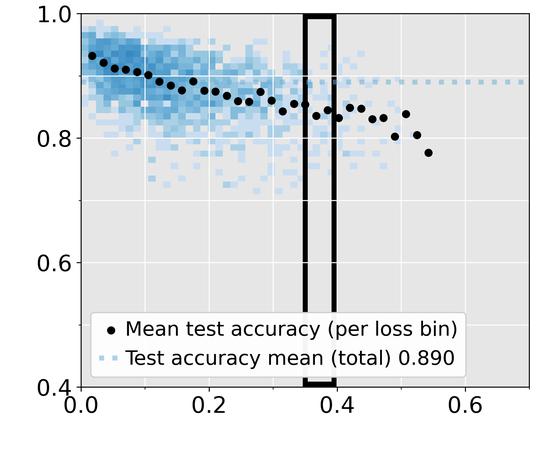}
    &\includegraphics{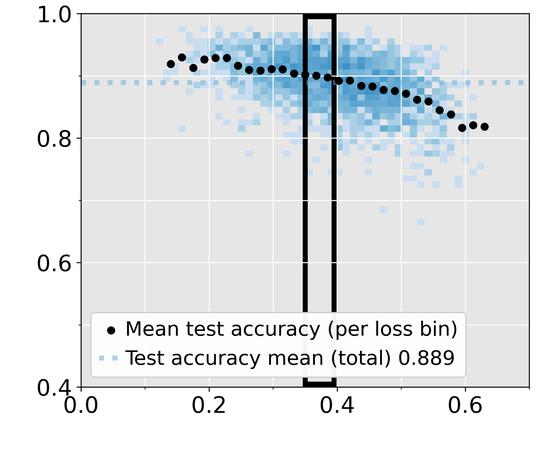}
    &\includegraphics{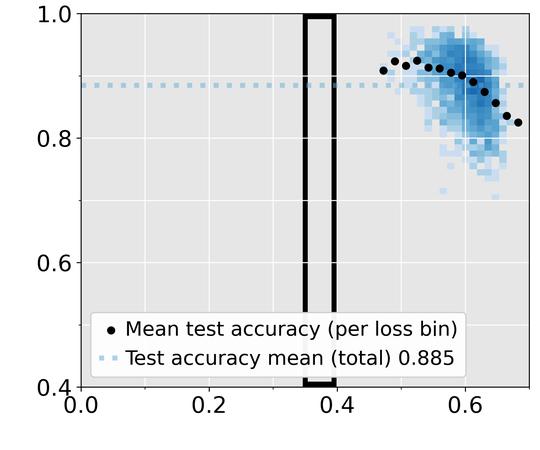}
    &\includegraphics{figures/different_loss_factor/colorbar.png} \\
     %%%%%%%%%%%%%%%%%%%%%%%%%%%%%%%%%%%
    % GNC 2
    %%%%%%%%%%%%%%%%%%%%%%%%%%%%%%%%%%%
    &
    &\rotatebox[origin=c]{90}    {\footnotesize  \enspace \textbf{\makecell{Test accuracy vs \\ Lipschitz norm. loss}}}
    
    &\includegraphics{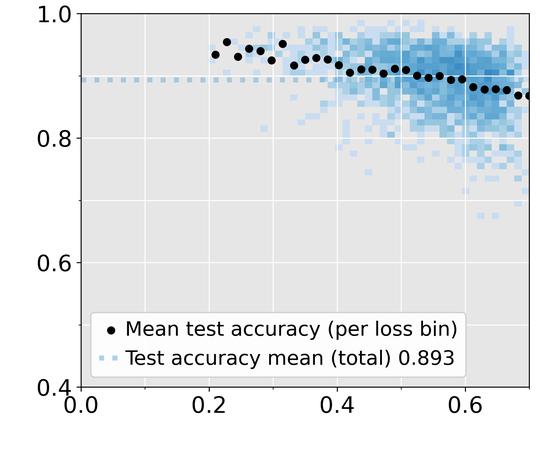}
    &\includegraphics {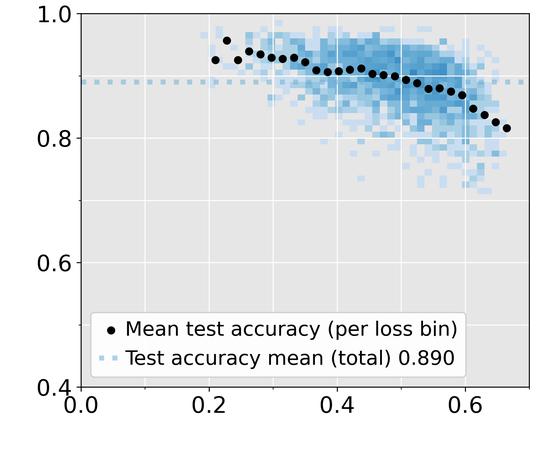}
    &\includegraphics{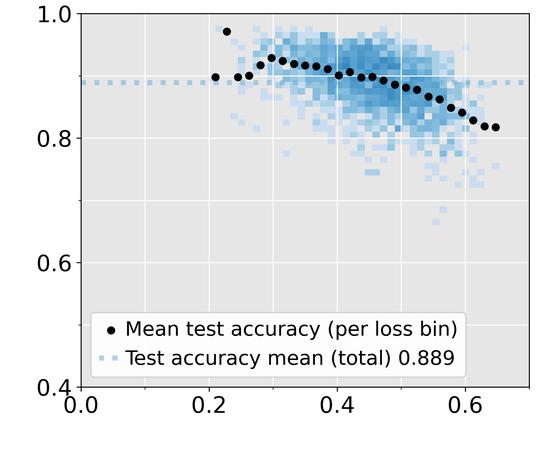}
    &\includegraphics{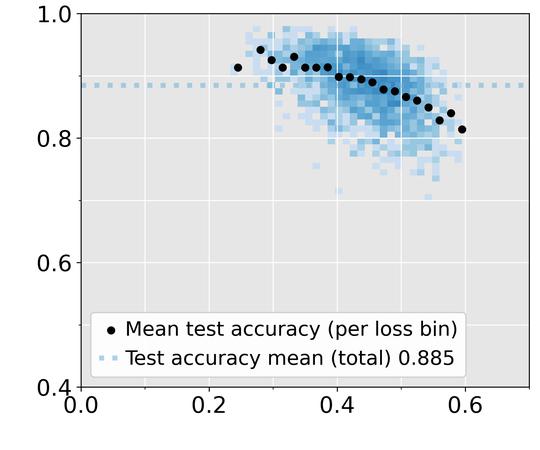}
    &\includegraphics{figures/different_loss_factor/colorbar.png} \\
    %%%%%%%%%%%%%%%%%%%%%%%%%%%%%%%%%%%
    % SGD 1
    %%%%%%%%%%%%%%%%%%%%%%%%%%%%%%%%%%%
    &\multirow{2}{*}[-3em]{\rotatebox{90}{\textbf{\sgd}}}
    &\rotatebox[origin=c]{90}{\quad     {\footnotesize \textbf{\makecell{Test accuracy vs\\ weight norm. loss}}}}
    & \includegraphics{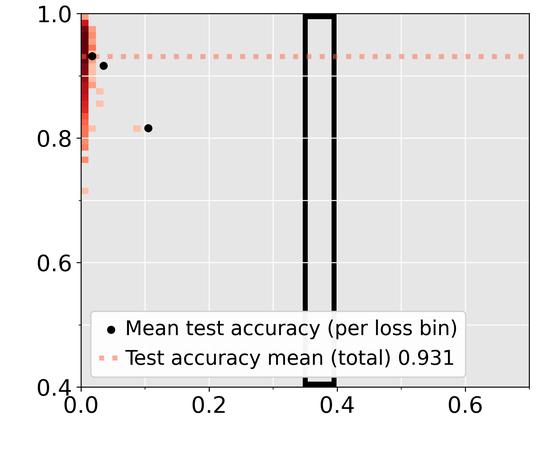}
    &\includegraphics{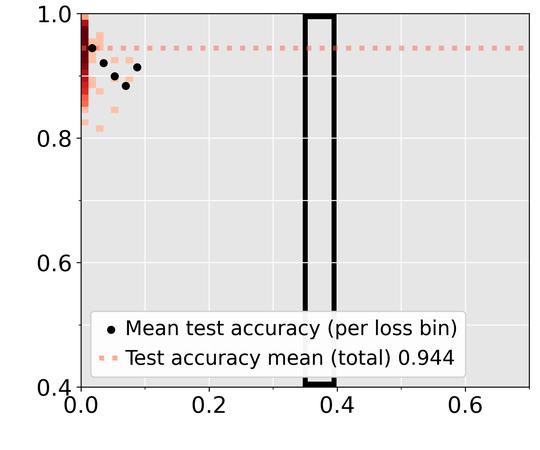}
    &\includegraphics{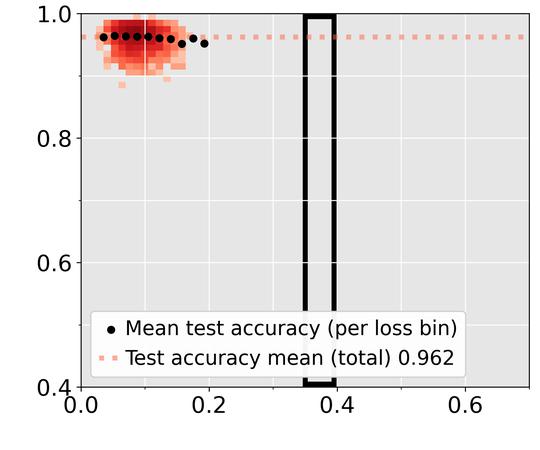}
    &\includegraphics{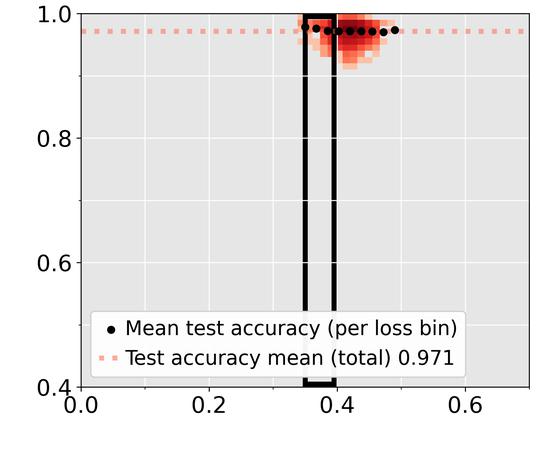}
    &\includegraphics{figures/different_loss_factor/colorbar_SGD.png} \\
    %%%%%%%%%%%%%%%%%%%%%%%%%%%%%%%%%%%
    % SGD 2
    %%%%%%%%%%%%%%%%%%%%%%%%%%%%%%%%%%%
    &
    &\rotatebox[origin=c]{90}    {\footnotesize  \enspace \textbf{\makecell{Test accuracy vs \\ Lipschitz norm. loss}}}
    &\includegraphics{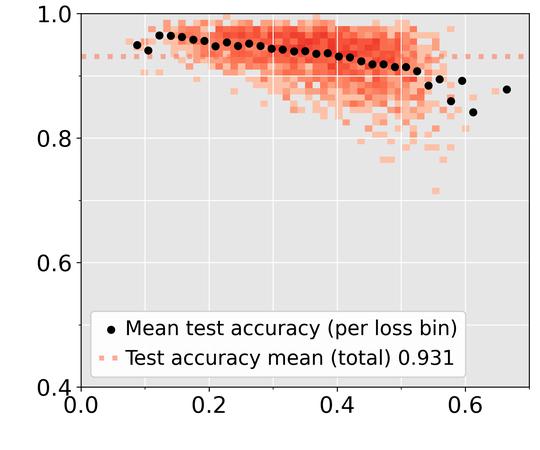}
    &\includegraphics{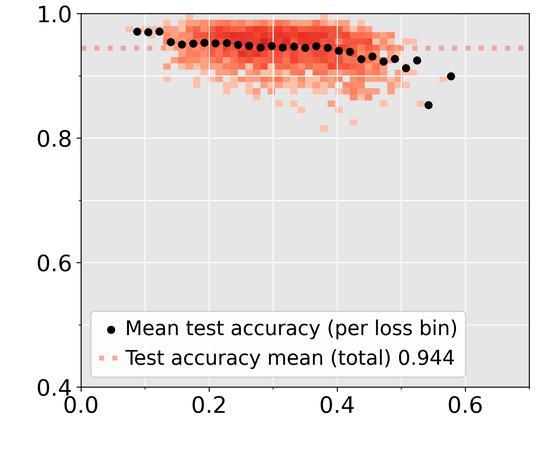}
    &\includegraphics{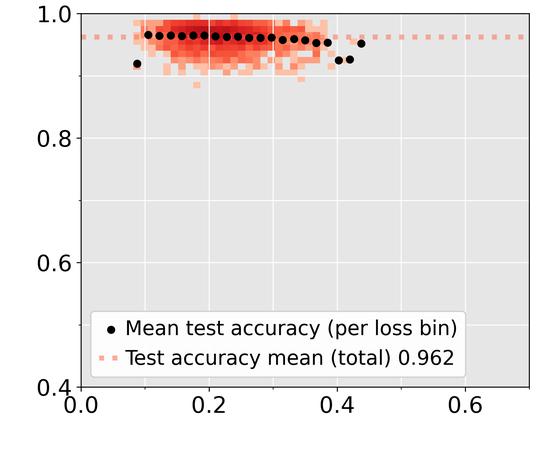}
    &\includegraphics{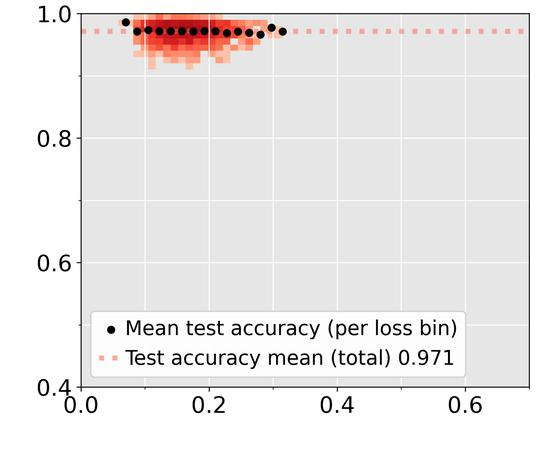}
    &\includegraphics{figures/different_loss_factor/colorbar_SGD.png} \\
    %%%%%%%%%%%%%%%%%%%%%%%%%%%%%%%%%
    &
    &
    & \multicolumn{1}{c}{\hspace{1.5em}\textbf{\makecell{Train loss\\(normalized)}}}
    & \multicolumn{1}{c}{\hspace{1.5em}\textbf{\makecell{Train loss\\(normalized)}}}
    & \multicolumn{1}{c}{\hspace{1.5em}\textbf{\makecell{Train loss\\(normalized)}}}
    & \multicolumn{1}{c}{\hspace{1.5em}\textbf{\makecell{Train loss\\(normalized)}}}
    &  \\
    %%%%%%%%%%%%%%%%%%%%%%%%%%
    \end{tabularx}
    \caption{
    \textbf{Analysis of overparameterization when increasing the width.}
    Test accuracy vs weight normalized loss \eqref{eq:weightnorm} of \citet{chiang2022loss} and our Lipschitz normalized loss \eqref{eq:lipschitznorm estimate} of {\color{amaranth}{\textbf{\sgd}}} and {\color{azure}{\textbf{\gnc}}} for classes \emph{0} \& \emph{7} of MNIST and 32 training samples.
    \textbf{Row 4:} Widening the networks enhances 
    geometric margin and average test accuracy for \sgd{}, while for \gnc{} \textbf{(Row 2)}, the margin improves only slightly, and average test accuracy remains the same. This suggests that the improvement is mainly due to the bias of \sgd{} and not due to an architectural bias (see Fig.~\ref{fig:different_widths}). \textbf{Rows 1 and 3:} \citet{chiang2022loss} compare networks conditional on the (weight) normalized loss bin (illustrated by black boxes), which led them to conclude that \gnc{} improves with increasing width.
    With our Lischitz normalized loss, one would arrive at the opposite conclusion, which shows the problem of normalization.
    }
    \label{fig:different_widths_loss_with_nornalization_32_samples}
    \vspace{-0.1cm}
\end{figure*}

\begin{figure*}[htb]
\centering
\setlength\tabcolsep{0pt}
\adjustboxset{width=\linewidth,valign=c}
\centering
\begin{tabularx}{1.0\linewidth}{
@{}l 
S{p{0.022\textwidth}} 
S{p{0.05\textwidth}} 
*{4}{S{p{0.22\textwidth}}} 
S{p{0.051\textwidth}}}
    %%%%%%%%%%%%%%%%%%%%%%%%%%
    &
    &
    & \multicolumn{1}{c}{\quad \textbf{Width} $\nicefrac{\mathbf{1}}{\mathbf{6^*}}$}
    & \multicolumn{1}{c}{\quad $\nicefrac{\mathbf{1}}{\mathbf{6}}$}
    & \multicolumn{1}{c}{\quad $\nicefrac{\mathbf{2}}{\mathbf{6}}$}
    & \multicolumn{1}{c}{\quad $\nicefrac{\mathbf{4}}{\mathbf{6}}$}
    & \multicolumn{1}{c}{} \\
    %%%%%%%%%%%%%%%%%%%%%%%%%%%%%%%%%%%
    % GNC 1
    %%%%%%%%%%%%%%%%%%%%%%%%%%%%%%%%%%%
    &\multirow{2}{*}[-4em]{\rotatebox{90}{\textbf{\gnc}}}
    &\rotatebox[origin=c]{90}{\quad     {\footnotesize  \textbf{\makecell{Test accuracy vs\\ weight norm. loss}}}}
    &\includegraphics{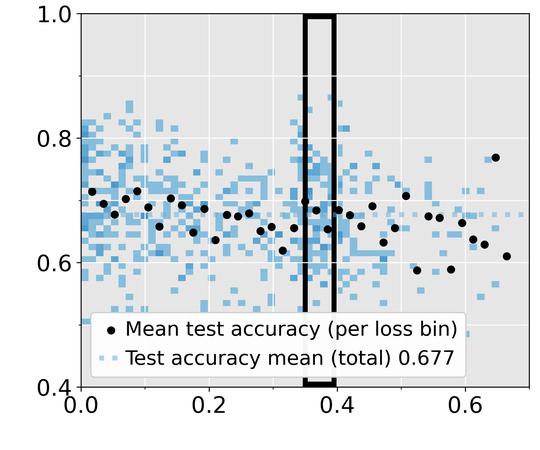}
    &\includegraphics{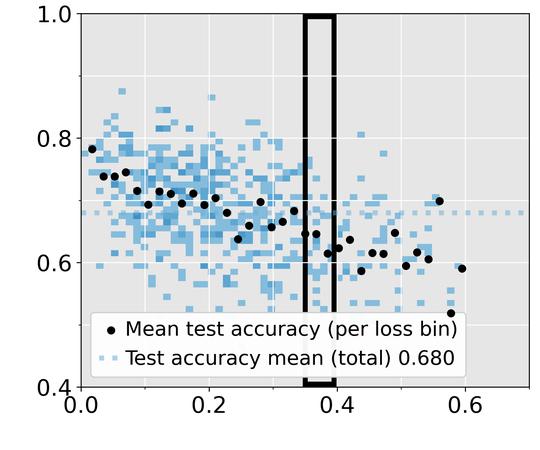}
    &\includegraphics{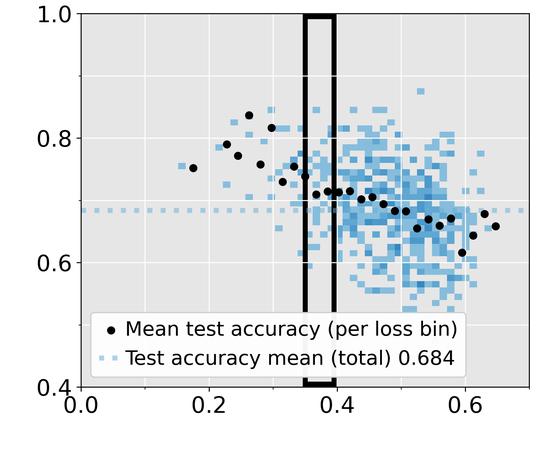}
    &\includegraphics{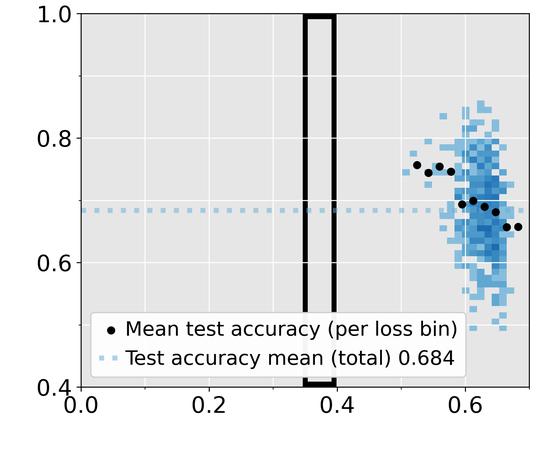}
    &\includegraphics{figures/different_loss_factor/colorbar.png} \\
     %%%%%%%%%%%%%%%%%%%%%%%%%%%%%%%%%%%
    % GNC 2
    %%%%%%%%%%%%%%%%%%%%%%%%%%%%%%%%%%%
    &
    &\rotatebox[origin=c]{90}    {\footnotesize  \enspace \textbf{\makecell{Test accuracy vs \\ Lipschitz norm. loss}}}
    &\includegraphics{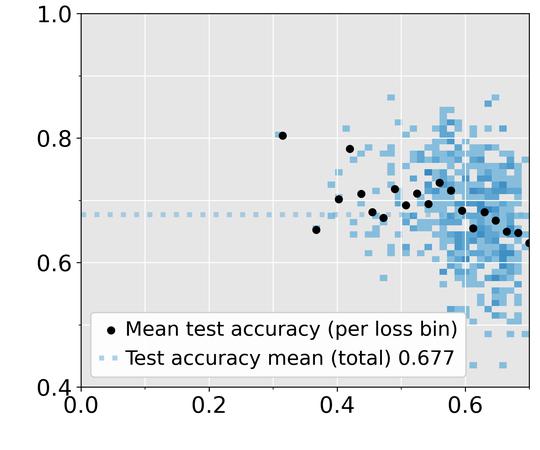}
    &\includegraphics {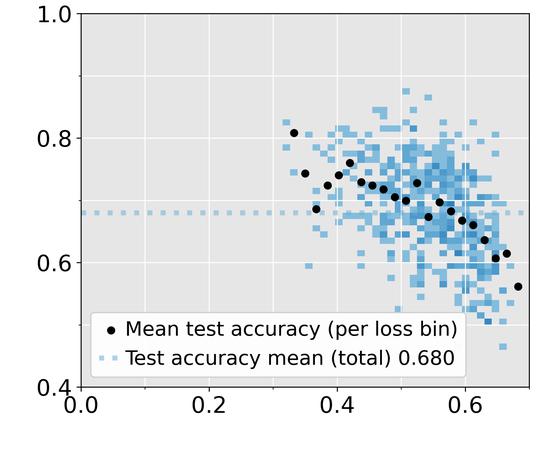}
    &\includegraphics{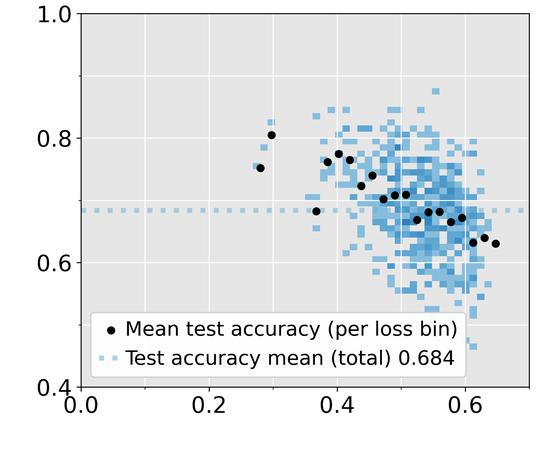}
    &\includegraphics{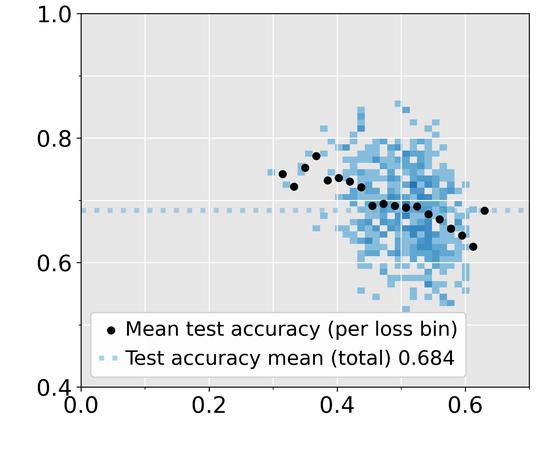}
    &\includegraphics{figures/different_loss_factor/colorbar.png} \\
    %%%%%%%%%%%%%%%%%%%%%%%%%%%%%%%%%%%
    % SGD 1
    %%%%%%%%%%%%%%%%%%%%%%%%%%%%%%%%%%%
    &\multirow{2}{*}[-4em]{\rotatebox{90}{\textbf{\sgd}}}
    &\rotatebox[origin=c]{90}{\quad     {\footnotesize  \textbf{\makecell{Test accuracy vs\\ weight norm. loss}}}}
    & \includegraphics{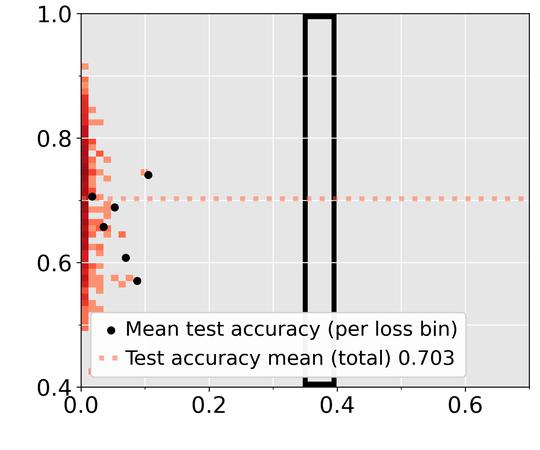}
    &\includegraphics{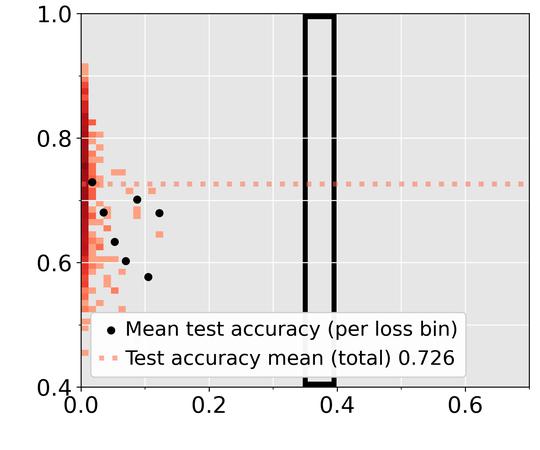}
    &\includegraphics{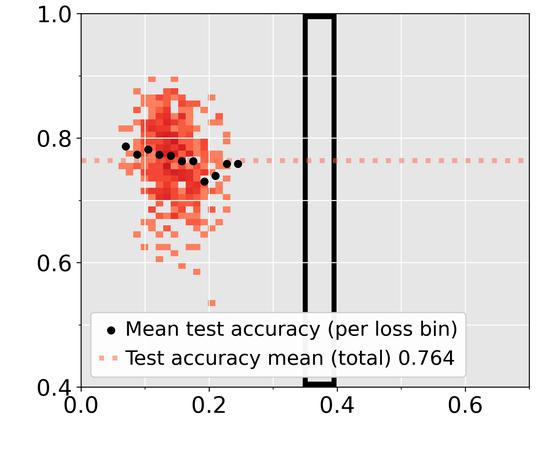}
    &\includegraphics{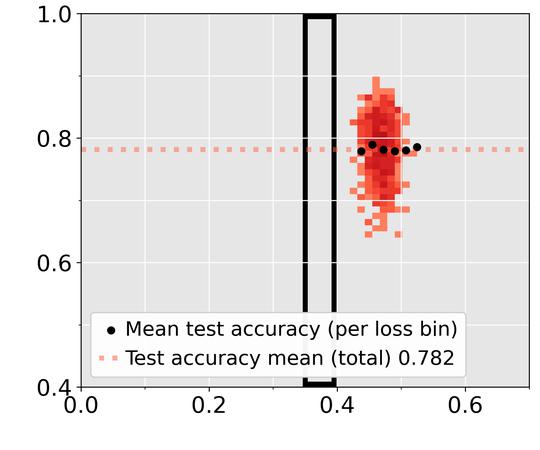}
    &\includegraphics{figures/different_loss_factor/colorbar_SGD.png} \\
    %%%%%%%%%%%%%%%%%%%%%%%%%%%%%%%%%%%
    % SGD 2
    %%%%%%%%%%%%%%%%%%%%%%%%%%%%%%%%%%%
    &
    &\rotatebox[origin=c]{90}    {\footnotesize  \enspace \textbf{\makecell{Test accuracy vs \\ Lipschitz norm. loss}}}
    &\includegraphics{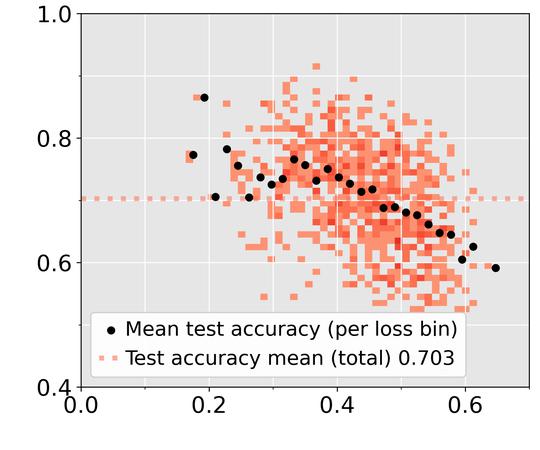}
    &\includegraphics{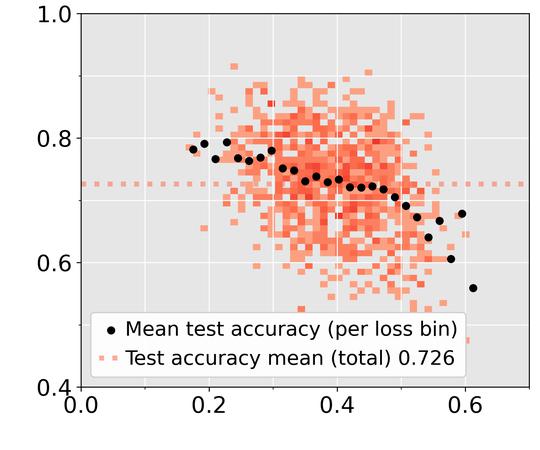}
    &\includegraphics{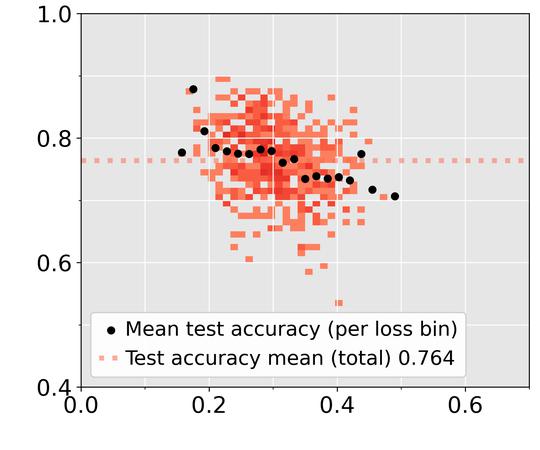}
    &\includegraphics{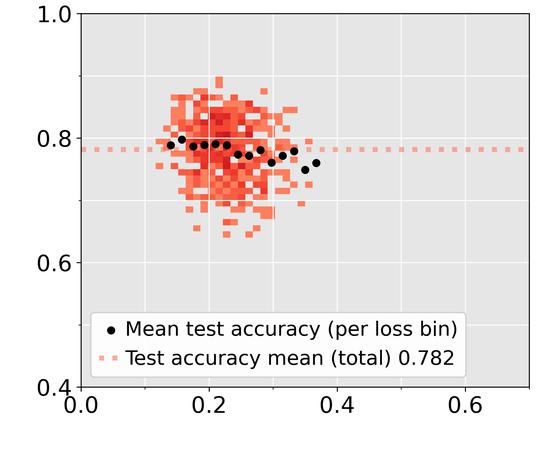}
    &\includegraphics{figures/different_loss_factor/colorbar_SGD.png} \\
    %%%%%%%%%%%%%%%%%%%%%%%%%%%%%%%%%
    &
    &
    & \multicolumn{1}{c}{\hspace{1.5em}\textbf{\makecell{Train loss\\(normalized)}}}
    & \multicolumn{1}{c}{\hspace{1.5em}\textbf{\makecell{Train loss\\(normalized)}}}
    & \multicolumn{1}{c}{\hspace{1.5em}\textbf{\makecell{Train loss\\(normalized)}}}
    & \multicolumn{1}{c}{\hspace{1.5em}\textbf{\makecell{Train loss\\(normalized)}}}
    &  \\
    %%%%%%%%%%%%%%%%%%%%%%%%%%
    \end{tabularx}
    \caption{\textbf{Qualitative analysis of overparameterization when widening the networks.}
    Test accuracy vs weight normalized loss \eqref{eq:weightnorm} of \citet{chiang2022loss} and our Lipschitz normalized loss \eqref{eq:lipschitznorm} of {\color{amaranth}{\textbf{\sgd}}} and {\color{azure}{\textbf{\gnc}}} for classes \emph{3} \& \emph{5} of MNIST and 16 training samples.
    \textbf{Bottom row:} Widening the networks enhances both geometric margin and average test accuracy for \sgd{}, while for \gnc{} \textbf{(second row)}, margin and average test accuracy improve only slightly. This suggests that the improvement is mainly due to the bias of \sgd{} and not due to an architectural bias. \textbf{Rows 1 and 3:} \citet{chiang2022loss} compare networks conditional on the (weight) normalized loss bin (illustrated by the black boxes), which led them to the conclusion of an increased volume of good minima for wider networks. However, with our Lischitz normalized loss, one would arrive at the opposite conclusion, which shows the problems of comparisons across network architectures. 
    }
    \label{fig:different_widths_loss_with_normalization_appendix_seed204}
\end{figure*}

\begin{figure*}[htb]
\centering
\setlength\tabcolsep{0pt}
\adjustboxset{width=\linewidth,valign=c}
\centering
\begin{tabularx}{1.0\linewidth}{
@{}l 
S{p{0.022\textwidth}} 
S{p{0.05\textwidth}} 
*{4}{S{p{0.22\textwidth}}} 
S{p{0.051\textwidth}}}
    %%%%%%%%%%%%%%%%%%%%%%%%%%
    &
    &
    & \multicolumn{1}{c}{\quad \textbf{Width} $\nicefrac{\mathbf{1}}{\mathbf{6^*}}$}
    & \multicolumn{1}{c}{\quad $\nicefrac{\mathbf{1}}{\mathbf{6}}$}
    & \multicolumn{1}{c}{\quad $\nicefrac{\mathbf{2}}{\mathbf{6}}$}
    & \multicolumn{1}{c}{\quad $\nicefrac{\mathbf{4}}{\mathbf{6}}$}
    & \multicolumn{1}{c}{} \\
    %%%%%%%%%%%%%%%%%%%%%%%%%%%%%%%%%%%
    % GNC 1
    %%%%%%%%%%%%%%%%%%%%%%%%%%%%%%%%%%%
    &\multirow{2}{*}[-4em]{\rotatebox{90}{\textbf{\gnc}}}
    &\rotatebox[origin=c]{90}{\quad     {\footnotesize  \textbf{\makecell{Test accuracy vs\\ weight norm. loss}}}}
    &\includegraphics{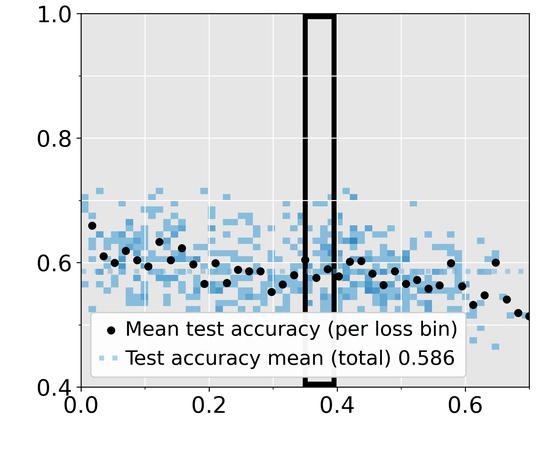}
    &\includegraphics{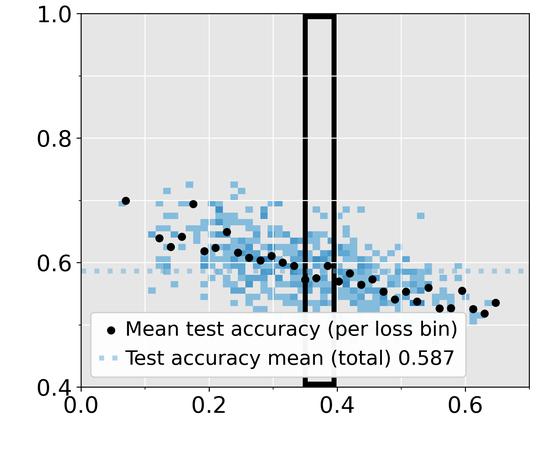}
    &\includegraphics{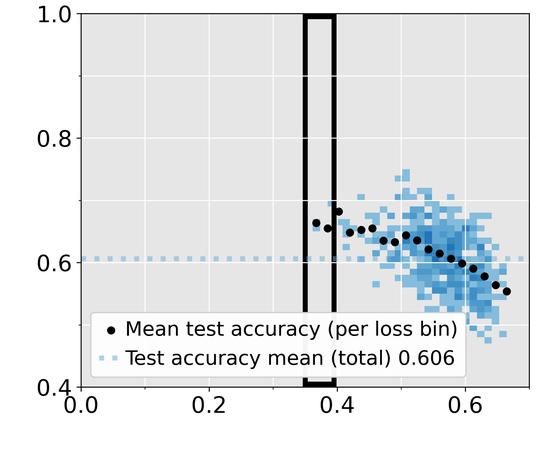}
    &\includegraphics{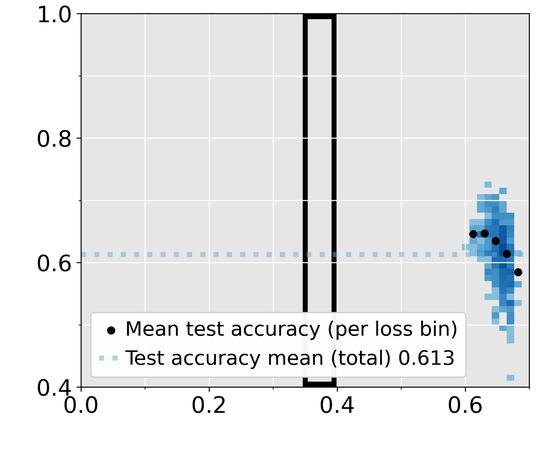}
    &\includegraphics{figures/different_loss_factor/colorbar.png} \\
     %%%%%%%%%%%%%%%%%%%%%%%%%%%%%%%%%%%
    % GNC 2
    %%%%%%%%%%%%%%%%%%%%%%%%%%%%%%%%%%%
    &
    &\rotatebox[origin=c]{90}    {\footnotesize  \enspace \textbf{\makecell{Test accuracy vs \\ Lipschitz norm. loss}}}
    
    &\includegraphics{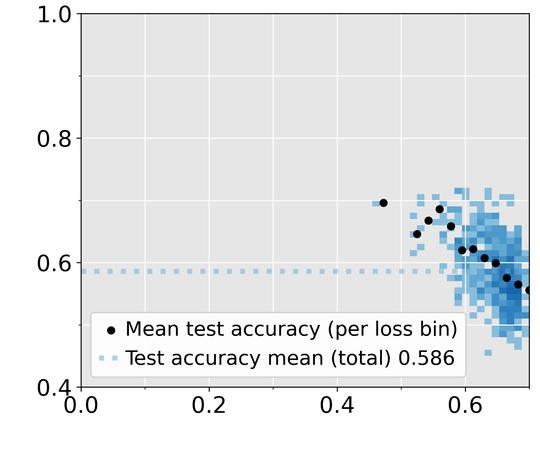}
    &\includegraphics {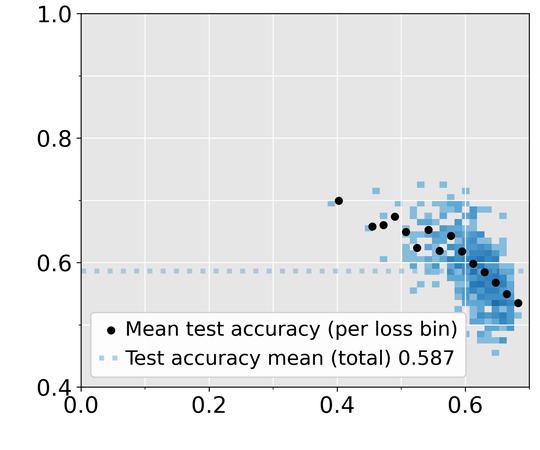}
    &\includegraphics{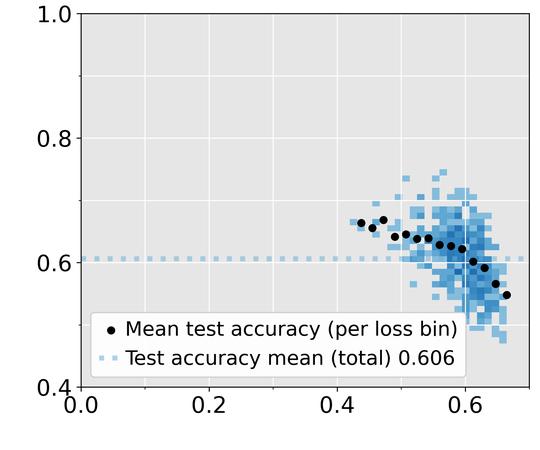}
    &\includegraphics{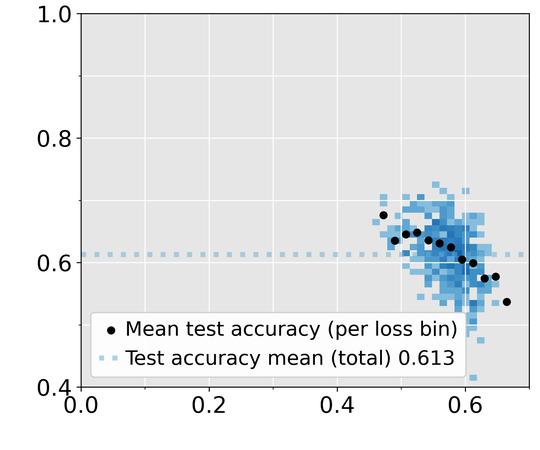}
    &\includegraphics{figures/different_loss_factor/colorbar.png} \\
    %%%%%%%%%%%%%%%%%%%%%%%%%%%%%%%%%%%
    % SGD 1
    %%%%%%%%%%%%%%%%%%%%%%%%%%%%%%%%%%%
    &\multirow{2}{*}[-4em]{\rotatebox{90}{\textbf{\sgd}}}
    &\rotatebox[origin=c]{90}{\quad     {\footnotesize  \textbf{\makecell{Test accuracy vs\\ weight norm. loss}}}}
    &\includegraphics{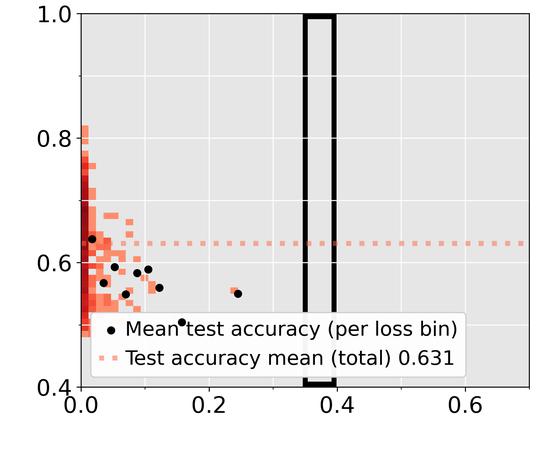}
    &\includegraphics{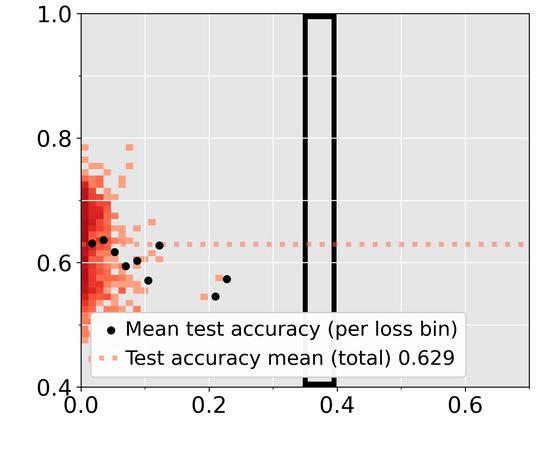}
    &\includegraphics{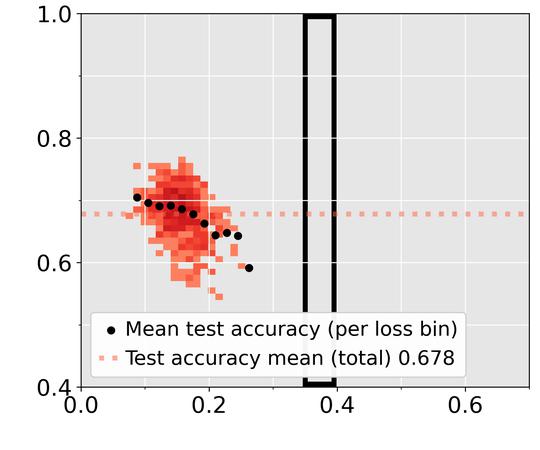}
     &\includegraphics{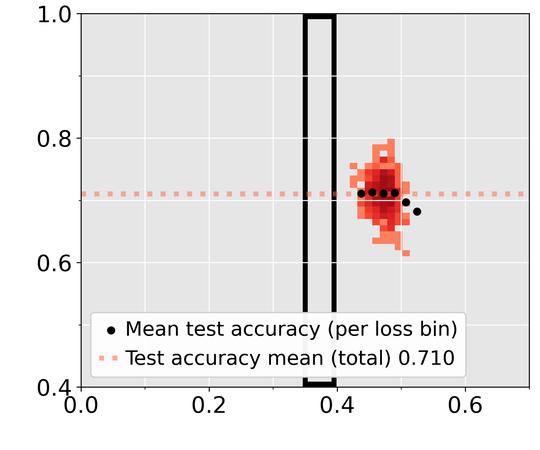}
    &\includegraphics{figures/different_loss_factor/colorbar_SGD.png} \\
    %%%%%%%%%%%%%%%%%%%%%%%%%%%%%%%%%%%
    % SGD 2
    %%%%%%%%%%%%%%%%%%%%%%%%%%%%%%%%%%%
    &
    &\rotatebox[origin=c]{90}    {\footnotesize  \enspace \textbf{\makecell{Test accuracy vs \\ Lipschitz norm. loss}}}
    
    &\includegraphics{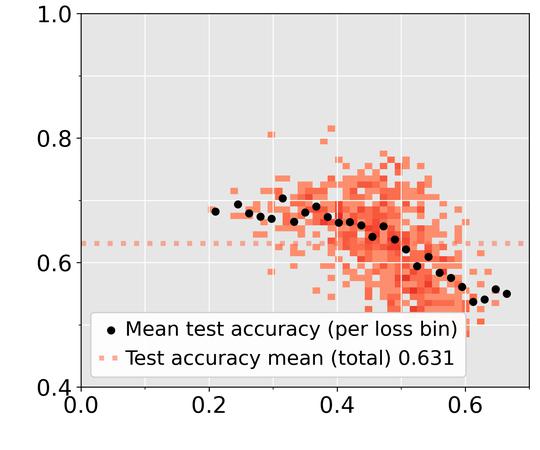}
    &\includegraphics{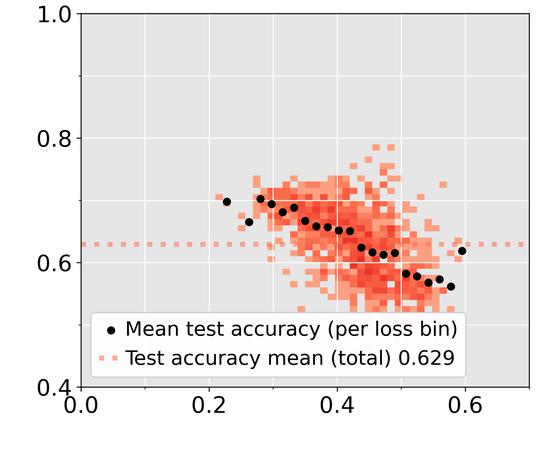}
    &\includegraphics{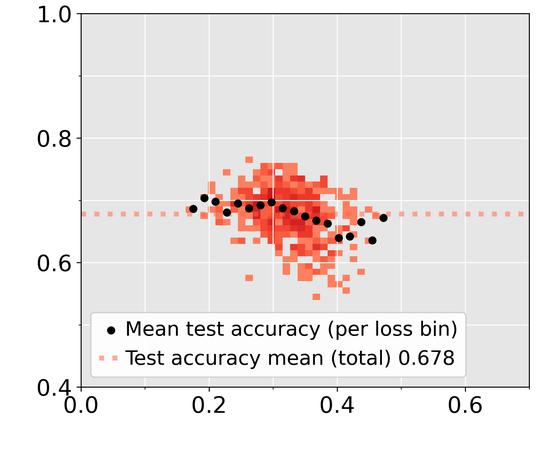}
    &\includegraphics{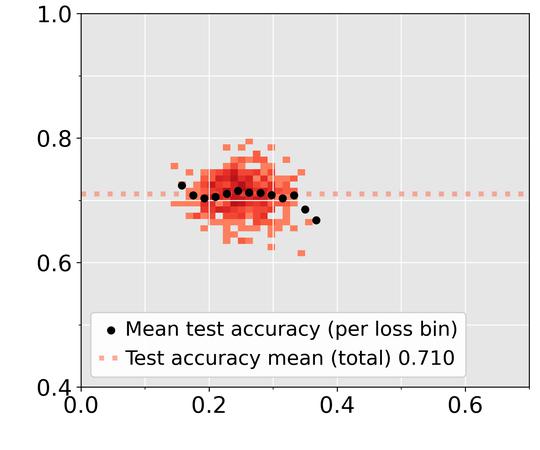}
    &\includegraphics{figures/different_loss_factor/colorbar_SGD.png} \\
    %%%%%%%%%%%%%%%%%%%%%%%%%%%%%%%%%
    &
    &
    & \multicolumn{1}{c}{\hspace{1.5em}\textbf{\makecell{Train loss\\(normalized)}}}
    & \multicolumn{1}{c}{\hspace{1.5em}\textbf{\makecell{Train loss\\(normalized)}}}
    & \multicolumn{1}{c}{\hspace{1.5em}\textbf{\makecell{Train loss\\(normalized)}}}
    & \multicolumn{1}{c}{\hspace{1.5em}\textbf{\makecell{Train loss\\(normalized)}}}
    &  \\
    %%%%%%%%%%%%%%%%%%%%%%%%%%
    \end{tabularx}
    \caption{\textbf{Qualitative analysis of overparameterization when widening the networks.}
    Test accuracy vs weight normalized loss \eqref{eq:weightnorm} of \citet{chiang2022loss} and our Lipschitz normalized loss \eqref{eq:lipschitznorm} of {\color{amaranth}{\textbf{\sgd}}} and {\color{azure}{\textbf{\gnc}}} for classes \emph{Deer} \& \emph{Truck} of CIFAR10 and 16 training samples.
    \textbf{Bottom row:} Widening the networks enhances both geometric margin and average test accuracy for \sgd{}, while for \gnc{} \textbf{(second row)}, margin and average test accuracy improve up to a width of $\nicefrac{2}{6}$ and then only slightly. This suggests that the improvement is mainly due to the bias of \sgd{} and not due to an architectural bias. \textbf{Rows 1 and 3:} \citet{chiang2022loss} compare networks conditional on the (weight) normalized loss bin (illustrated by the black boxes), which led them to the conclusion of an increased volume of good minima for wider networks. However, with our Lischitz normalized loss, one would arrive at the opposite conclusion, which shows the problems of comparisons across network architectures. 
    }
    \label{fig:different_widths_loss_with_normalization_appendix_seed219}
\end{figure*}

\begin{figure*}[htb]
\centering
\setlength\tabcolsep{0pt}
\adjustboxset{width=\linewidth,valign=c}
\centering
\begin{tabularx}{1.0\linewidth}{
@{}l 
S{p{0.022\textwidth}} 
S{p{0.05\textwidth}} 
*{4}{S{p{0.22\textwidth}}} 
S{p{0.051\textwidth}}}
    %%%%%%%%%%%%%%%%%%%%%%%%%%
    &
    &
    & \multicolumn{1}{c}{\quad \textbf{Width} $\nicefrac{\mathbf{1}}{\mathbf{6^*}}$}
    & \multicolumn{1}{c}{\quad $\nicefrac{\mathbf{1}}{\mathbf{6}}$}
    & \multicolumn{1}{c}{\quad $\nicefrac{\mathbf{2}}{\mathbf{6}}$}
    & \multicolumn{1}{c}{\quad $\nicefrac{\mathbf{4}}{\mathbf{6}}$}
    & \multicolumn{1}{c}{} \\
    %%%%%%%%%%%%%%%%%%%%%%%%%%%%%%%%%%%
    % GNC 1
    %%%%%%%%%%%%%%%%%%%%%%%%%%%%%%%%%%%
    &\multirow{2}{*}[-4em]{\rotatebox{90}{\textbf{\gnc}}}
    &\rotatebox[origin=c]{90}{\quad     {\footnotesize  \textbf{\makecell{Test accuracy vs\\ weight norm. loss}}}}
    &\includegraphics{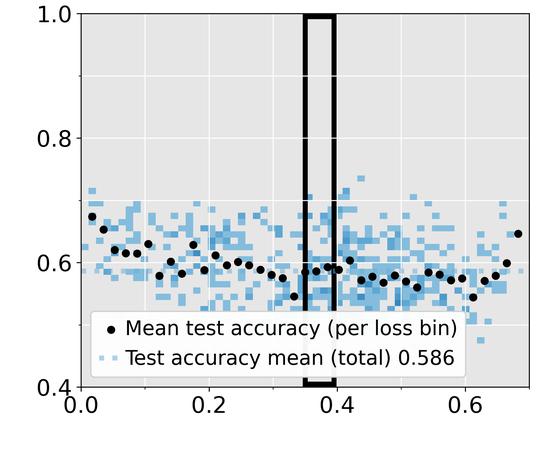}
    &\includegraphics{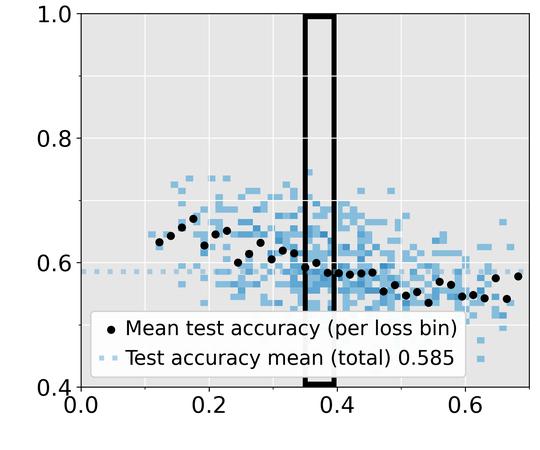}
    &\includegraphics{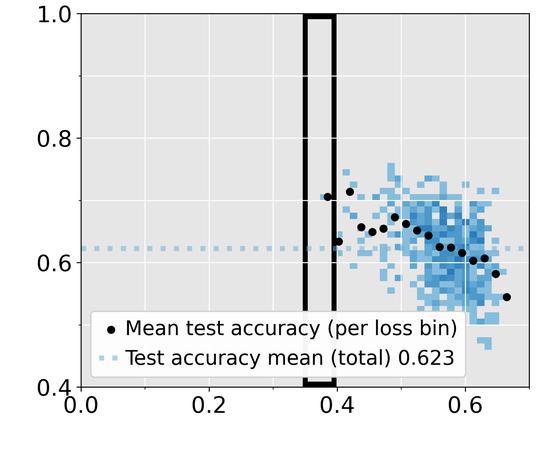}
    &\includegraphics{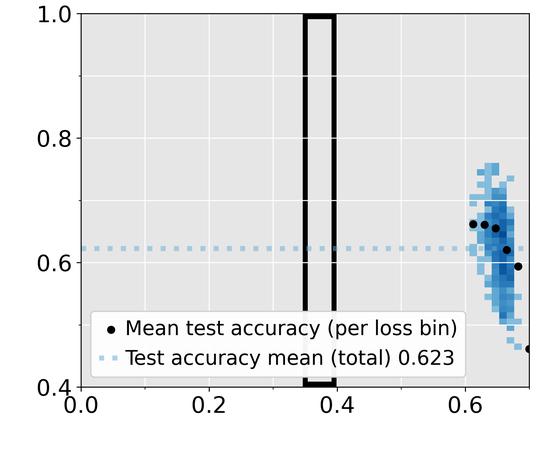}
    &\includegraphics{figures/different_loss_factor/colorbar.png} \\
     %%%%%%%%%%%%%%%%%%%%%%%%%%%%%%%%%%%
    % GNC 2
    %%%%%%%%%%%%%%%%%%%%%%%%%%%%%%%%%%%
    &
    &\rotatebox[origin=c]{90}    {\footnotesize  \enspace \textbf{\makecell{Test accuracy vs \\ Lipschitz norm. loss}}}
    
    &\includegraphics{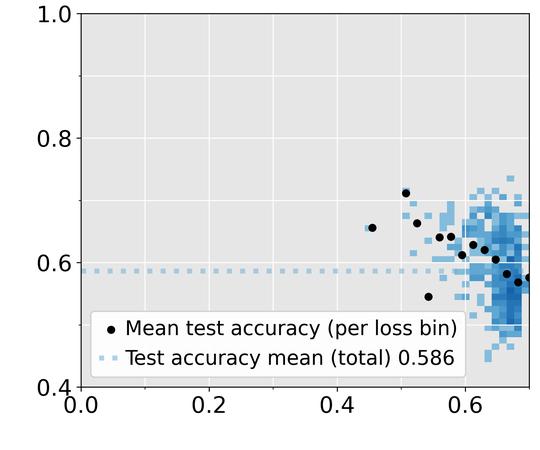}
    &\includegraphics {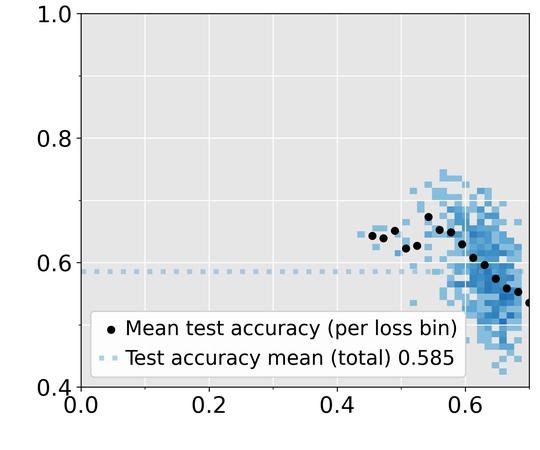}
    &\includegraphics{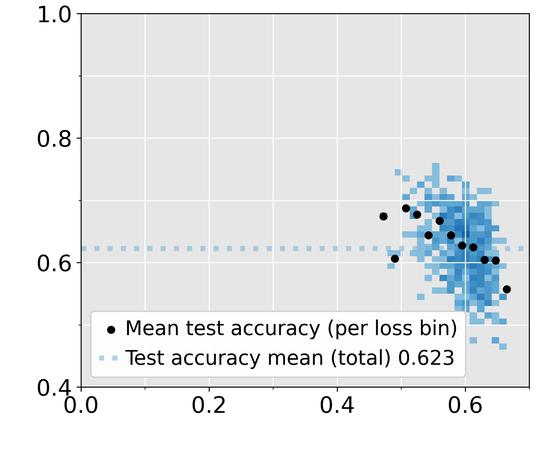}
    &\includegraphics{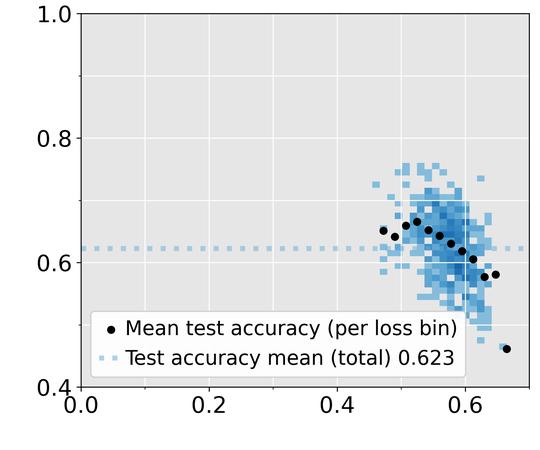}
    &\includegraphics{figures/different_loss_factor/colorbar.png} \\
    %%%%%%%%%%%%%%%%%%%%%%%%%%%%%%%%%%%
    % SGD 1
    %%%%%%%%%%%%%%%%%%%%%%%%%%%%%%%%%%%
    &\multirow{2}{*}[-4em]{\rotatebox{90}{\textbf{\sgd}}}
    &\rotatebox[origin=c]{90}{\quad     {\footnotesize  \textbf{\makecell{Test accuracy vs\\ weight norm. loss}}}}
    & \includegraphics{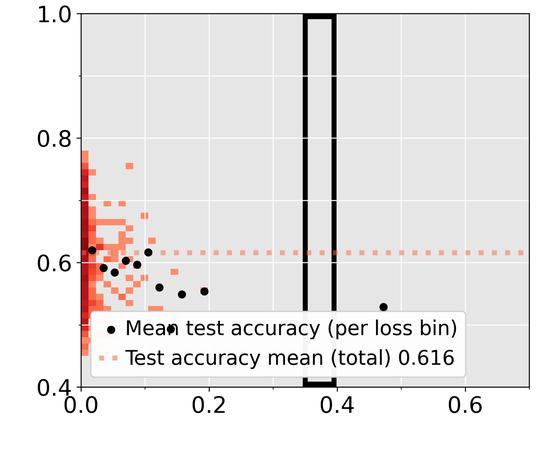}
    &\includegraphics{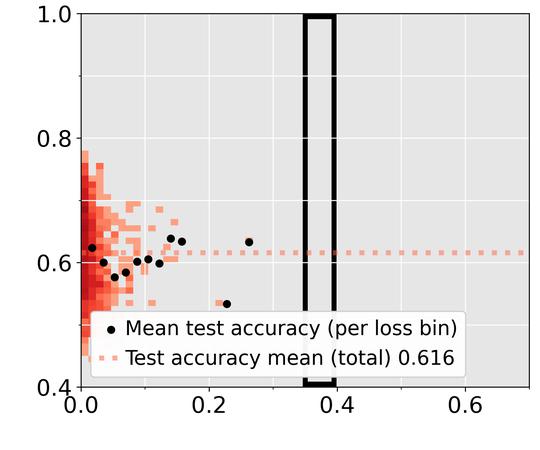}
    &\includegraphics{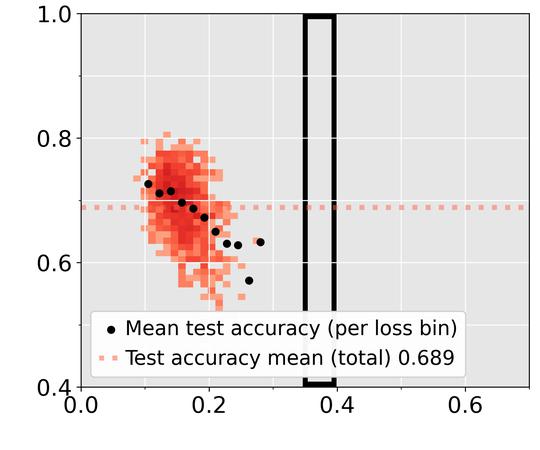}
    &\includegraphics{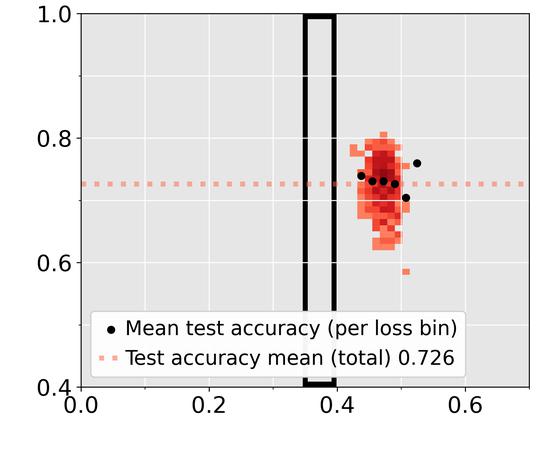}
    &\includegraphics{figures/different_loss_factor/colorbar_SGD.png} \\
    %%%%%%%%%%%%%%%%%%%%%%%%%%%%%%%%%%%
    % SGD 2
    %%%%%%%%%%%%%%%%%%%%%%%%%%%%%%%%%%%
    &
    &\rotatebox[origin=c]{90}    {\footnotesize  \enspace \textbf{\makecell{Test accuracy vs \\ Lipschitz norm. loss}}}
    
    &\includegraphics{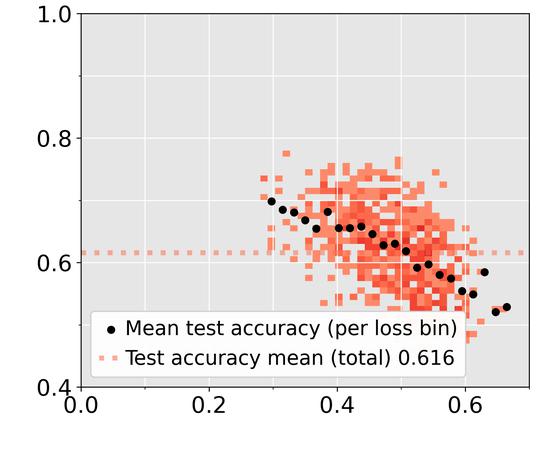}
    &\includegraphics{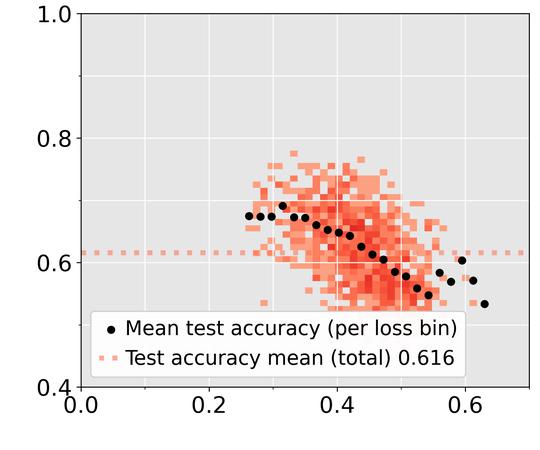}
    &\includegraphics{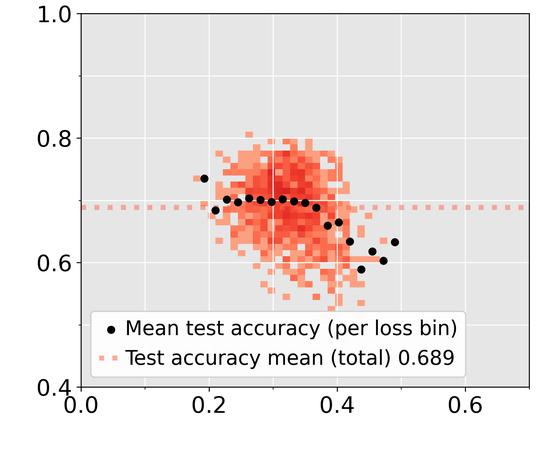}
    &\includegraphics{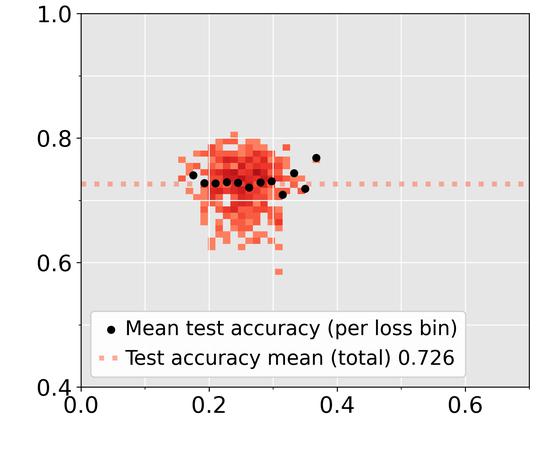}
    &\includegraphics{figures/different_loss_factor/colorbar_SGD.png} \\
    %%%%%%%%%%%%%%%%%%%%%%%%%%%%%%%%%
    &
    &
    & \multicolumn{1}{c}{\hspace{1.5em}\textbf{\makecell{Train loss\\(normalized)}}}
    & \multicolumn{1}{c}{\hspace{1.5em}\textbf{\makecell{Train loss\\(normalized)}}}
    & \multicolumn{1}{c}{\hspace{1.5em}\textbf{\makecell{Train loss\\(normalized)}}}
    & \multicolumn{1}{c}{\hspace{1.5em}\textbf{\makecell{Train loss\\(normalized)}}}
    &  \\
    %%%%%%%%%%%%%%%%%%%%%%%%%%
    \end{tabularx}
    \caption{\textbf{Qualitative analysis of overparameterization when widening the networks.}
    Test accuracy vs weight normalized loss \eqref{eq:weightnorm} of \citet{chiang2022loss} and our Lipschitz normalized loss \eqref{eq:lipschitznorm} of {\color{amaranth}{\textbf{\sgd}}} and {\color{azure}{\textbf{\gnc}}} for classes \emph{Bird} \& \emph{Ship} of CIFAR10 and 16 training samples.
    \textbf{Bottom row:} Widening the networks enhances both geometric margin and average test accuracy for \sgd{}, while for \gnc{} \textbf{(second row)}, margin and average test accuracy improve up to a width of $\nicefrac{2}{6}$ and then only slightly. This suggests that the improvement is mainly due to the bias of \sgd{} and not due to an architectural bias. \textbf{Rows 1 and 3:} \citet{chiang2022loss} compare networks conditional on the (weight) normalized loss bin (illustrated by the black boxes), which led them to the conclusion of an increased volume of good minima for wider networks. However, with our Lischitz normalized loss, one would arrive at the opposite conclusion, which shows the problems of comparisons across network architectures. 
    }
    \label{fig:different_widths_loss_with_nornalization_cifar}
\end{figure*}

Figure~\ref{fig:different_widths_loss_with_nornalization_4_samples} and Figure~\ref{fig:different_widths_loss_with_nornalization_32_samples} extend the results shown in Figure~\ref{fig:different_widths_loss_with_nornalization} for different sample sizes.
Figure~\ref{fig:different_widths_loss_with_normalization_appendix_seed204} provides results for another pair of classes from MNIST.
Similarly, in Figure~\ref{fig:different_widths_loss_with_normalization_appendix_seed219} and Figure~\ref{fig:different_widths_loss_with_nornalization_cifar} we follow the analysis of Figure~\ref{fig:different_widths_loss_with_nornalization}
for two additional classes of CIFAR10.

\begin{figure*}
\setlength\tabcolsep{0pt}
\adjustboxset{width=\linewidth,valign=c}
\centering
\begin{tabularx}{0.75\linewidth}{
@{}l 
S{p{0.02\textwidth}} 
*{3}{S{p{0.245\textwidth}}}}
    %%%%%%%%%%%%%%%%%%%%%%%%%%
    &
    & \multicolumn{1}{c}{\quad \enspace \textbf{Mean test accuracy}}
    & \multicolumn{1}{c}
    {
    \textbf{\quad \enspace \makecell{Probability to sample\\ 100\% train accuracy}}}
    & \multicolumn{1}{c}{\textbf{\makecell{\quad Distribution of\\ \quad test accuracies}}}\\
    %%%%%%%%%%%%%%%%%%%%%%%%%%%%%%%%%%%
    % 2 vs 5 (200)
    %%%%%%%%%%%%%%%%%%%%%%%%%%%%%%%%%%%
    &\rotatebox[origin=c]{90}{\textbf{\quad 2 vs 5}}
    &\includegraphics{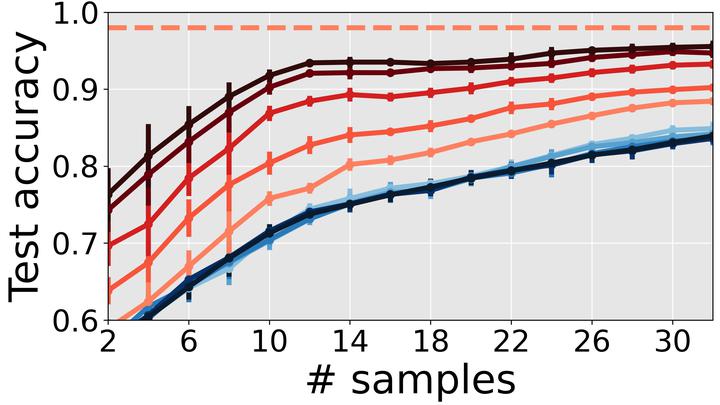}
    &\includegraphics{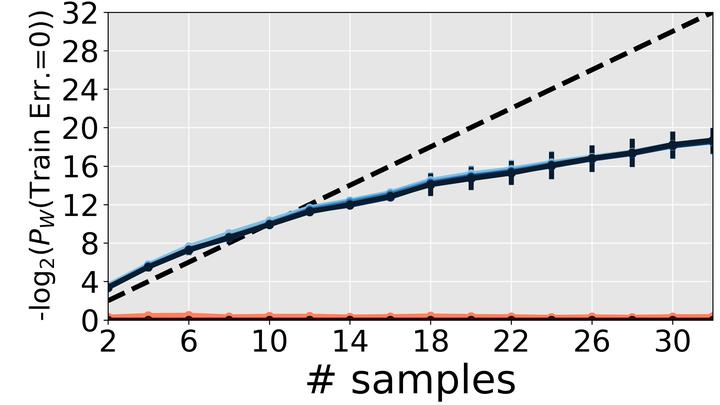}
    &\includegraphics{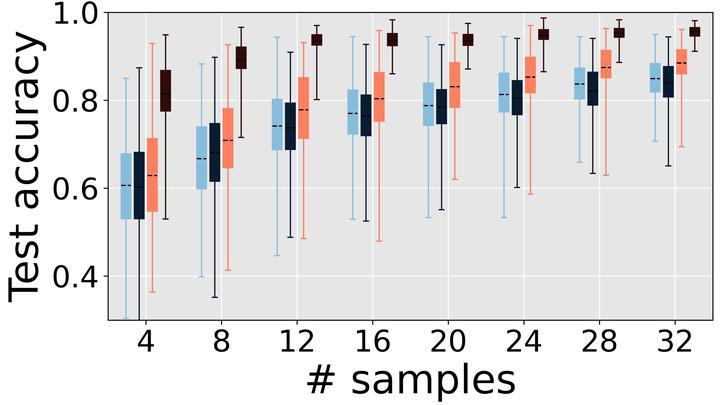}\\
   %%%%%%%%%%%%%%%%%%%%%%%%%%%%%%%%%%%
    % 2 vs 8 (201)
    %%%%%%%%%%%%%%%%%%%%%%%%%%%%%%%%%%%
    &\rotatebox[origin=c]{90}{\textbf{\quad 2 vs 8}}
    &\includegraphics{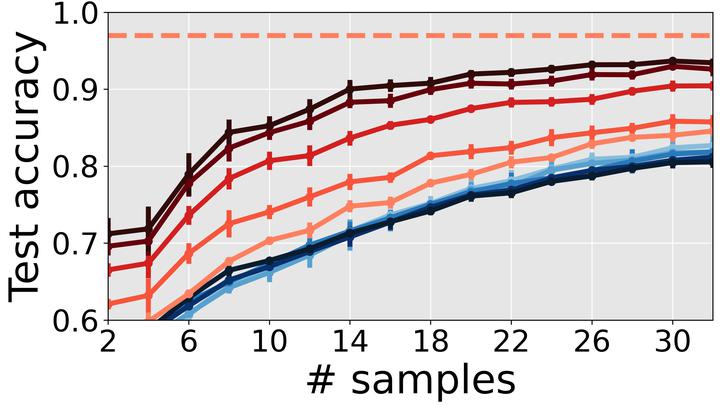}
    &\includegraphics{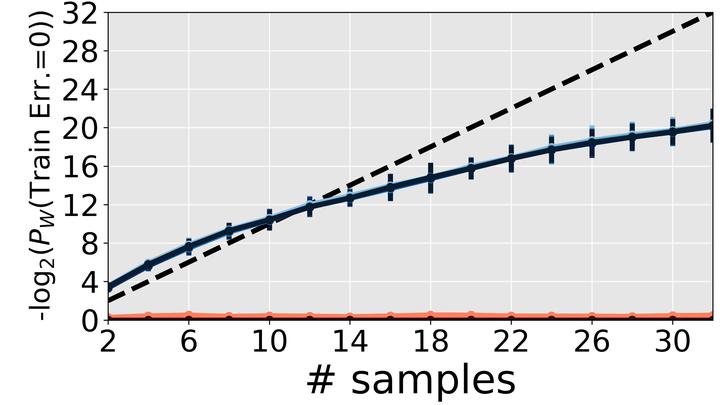}
    &\includegraphics{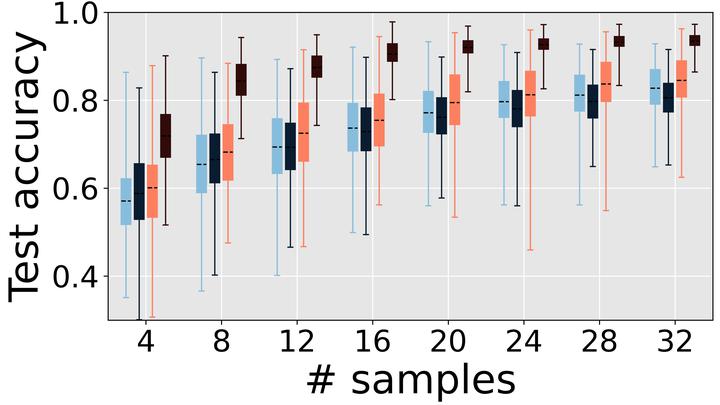}\\

    %%%%%%%%%%%%%%%%%%%%%%%%%%%%%%%%%%%
    % 0 vs 7 (202)
    %%%%%%%%%%%%%%%%%%%%%%%%%%%%%%%%%%%
    &\rotatebox[origin=c]{90}{\textbf{\quad 0 vs 7}}
    &\includegraphics{figures/different_widths/acc_plot_mnist_202.jpg}
    &\includegraphics{figures/different_widths/num_models_plot_mnist_202_lenet2.jpg}
    &\includegraphics{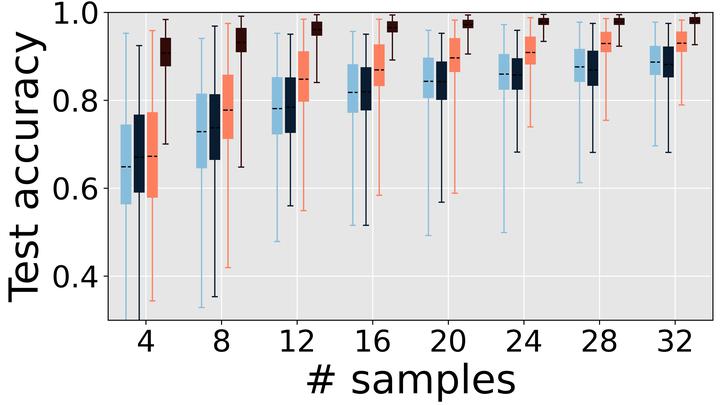}\\
    %%%%%%%%%%%%%%%%%%%%%%%%%%%%%%%%%%%
    % 0 vs 8 (203)
    %%%%%%%%%%%%%%%%%%%%%%%%%%%%%%%%%%%
    &\rotatebox[origin=c]{90}{\textbf{\quad 0 vs 8}}
    &\includegraphics{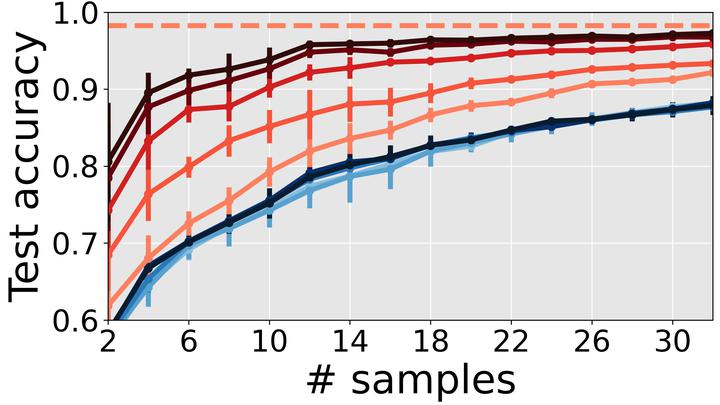}
    &\includegraphics{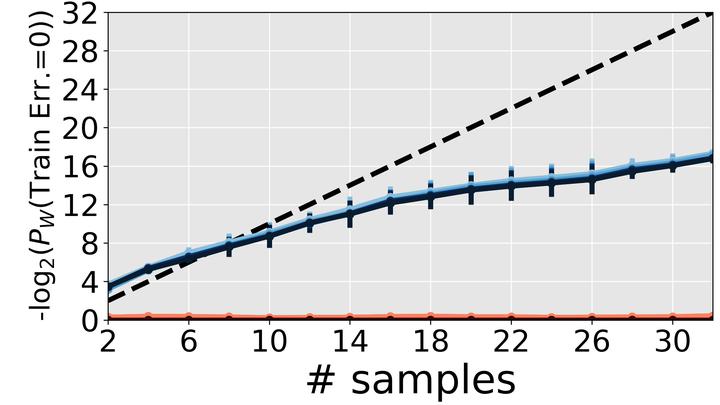}
    &\includegraphics{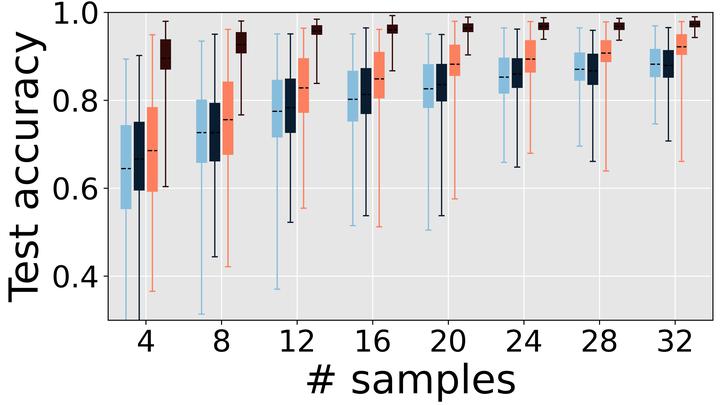}\\
    %%%%%%%%%%%%%%%%%%%%%%%%%%%%%%%%%%%
    % 3 vs 5 (204)
    %%%%%%%%%%%%%%%%%%%%%%%%%%%%%%%%%%%
    &\rotatebox[origin=c]{90}{\textbf{\quad 3 vs 5}}
    &\includegraphics{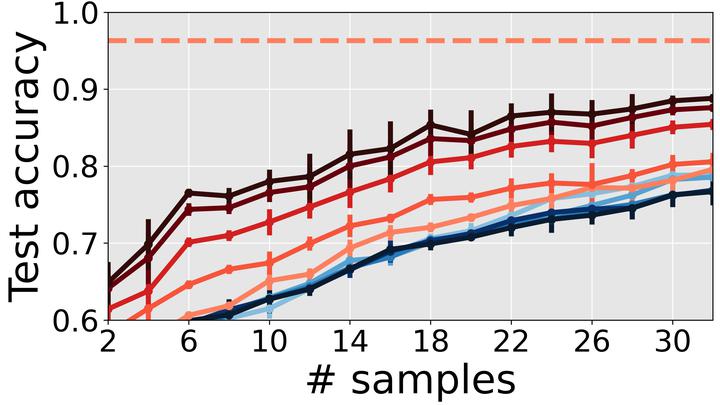}
    &\includegraphics{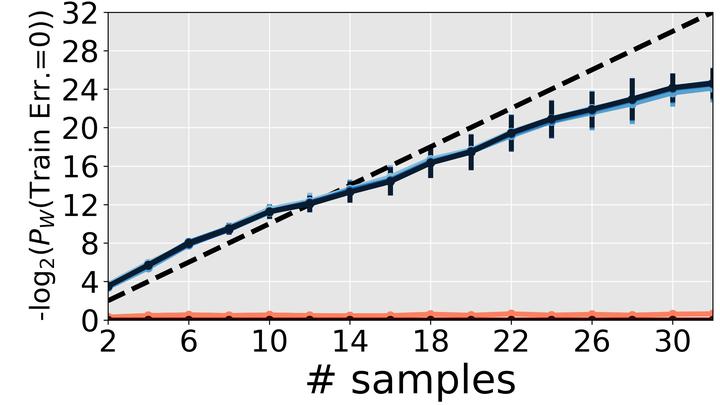}
    &\includegraphics{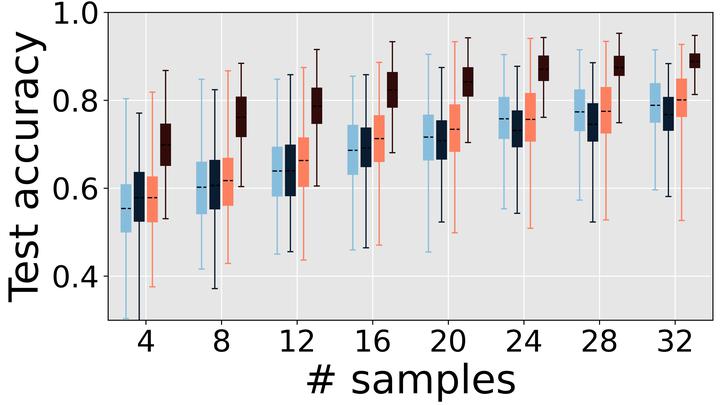}\\
    & \multicolumn{4}{c}{\includegraphics[width=0.75\linewidth]
    {figures/different_widths/legend_widths_lenet.png}
    }
    \end{tabularx}
    \caption{\textbf{Increasing width is a positive optimization bias.} 
    The rows represent different pairs of classes from the MNIST dataset using the LeNet architecture.
    \textbf{Column 1:} Mean test accuracy vs number
    of training samples across network widths. 
    Wider networks result in improvement for \sgd{}, whereas \gnc{} shows no change.
    Thus, for increasing width, there is a bias of \sgd{} towards better generalizing networks independent of an architectural bias. 
    However, overfitting does not occur for \gnc{}, implying that overparameterization does not hurt generalization. 
    \textbf{Column 2:} 
    We report the negative log probability of \gnc{} to find a network fitting the training data (${\Pr_W(\TrainError=0)}$).
    Note that this does not change with the width, which indicates that the pool of ``fitting networks'' does not change when the width increases.  
    \textbf{Column 3:} We report the combined distribution of the four subsets for the extreme width factors.
    While for \gnc{}, the entire distribution of models aligns, for \sgd{}, the wider networks are considerably better than the narrow networks (and the \gnc{} networks).
    For more details, please refer to Section~\ref{sec:width}.}
\label{fig:different_widths_mnist_seeds}
\end{figure*}

\begin{figure*}
\centering
\setlength\tabcolsep{0pt}
\adjustboxset{width=\linewidth,valign=c}
\centering
\begin{tabularx}{0.75\linewidth}{
@{}l 
S{p{0.02\textwidth}} 
*{3}{S{p{0.24\textwidth}}}}
    %%%%%%%%%%%%%%%%%%%%%%%%%%
    &
    & \multicolumn{1}{c}{\textbf{\quad \enspace Mean test accuracy}}
    & \multicolumn{1}{c}
    {\textbf{\quad \enspace \makecell{Probability to sample\\ 100\% train accuracy}}}
    & \multicolumn{1}{c}{\textbf{\makecell{\quad Distribution of\\ \quad test accuracies}}}\\
    %%%%%%%%%%%%%%%%%%%%%%%%%%%%%%%%%%%
    % Bird vs Ship (201)
    %%%%%%%%%%%%%%%%%%%%%%%%%%%%%%%%%%%
    &\rotatebox[origin=c]{90}{\textbf{\quad Bird vs Ship}}
    &\includegraphics{figures/different_widths/CIFAR10/seed201/acc_plot_cifar_201.jpg}
    &\includegraphics{figures/different_widths/CIFAR10/seed201/num_models_plot_cifar_201_lenet2.jpg}
    &\includegraphics{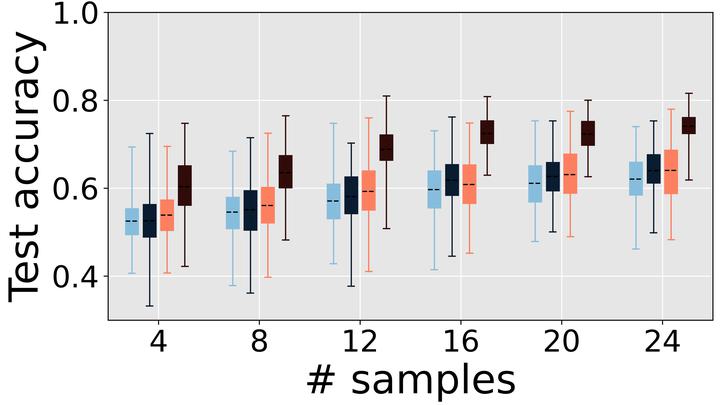}\\
   %%%%%%%%%%%%%%%%%%%%%%%%%%%%%%%%%%%
    % Deer vs Truck (219)
    %%%%%%%%%%%%%%%%%%%%%%%%%%%%%%%%%%%
    &\rotatebox[origin=c]{90}{\textbf{\quad Deer vs Truck}}
    &\includegraphics{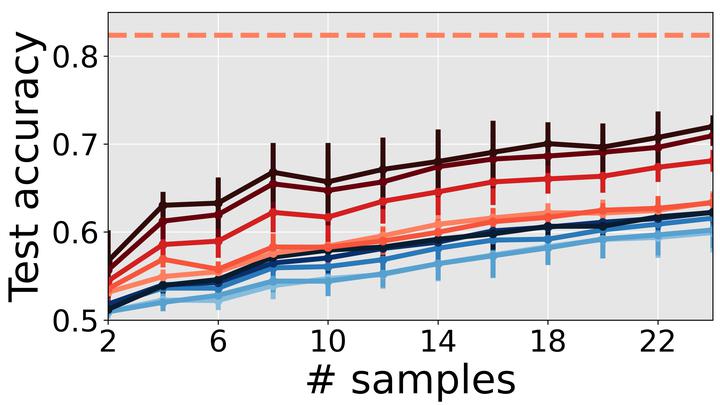}
    &\includegraphics{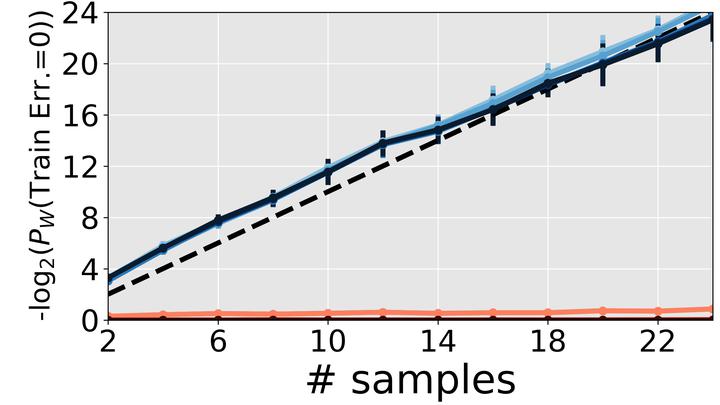}
    &\includegraphics{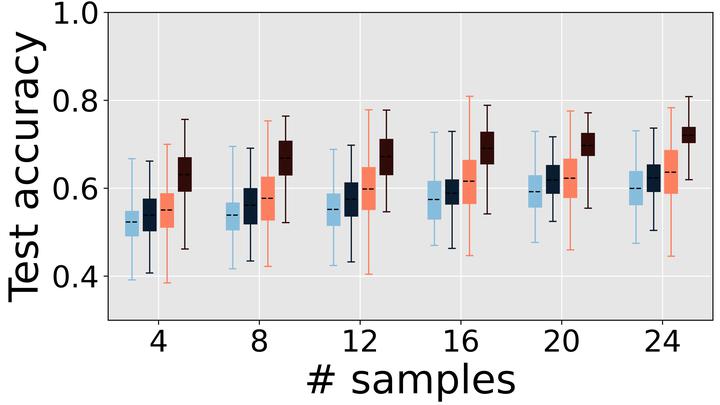}\\
    %%%%%%%%%%%%%%%%%%%%%%%%%%%%%%%%%%%
    % Plane vs Frog (224)
    %%%%%%%%%%%%%%%%%%%%%%%%%%%%%%%%%%%
    &\rotatebox[origin=c]{90}{\textbf{\quad Plane vs Frog}}
    &\includegraphics{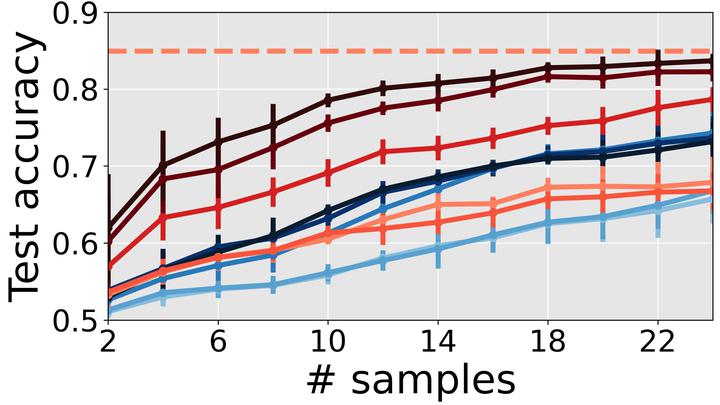}
    &\includegraphics{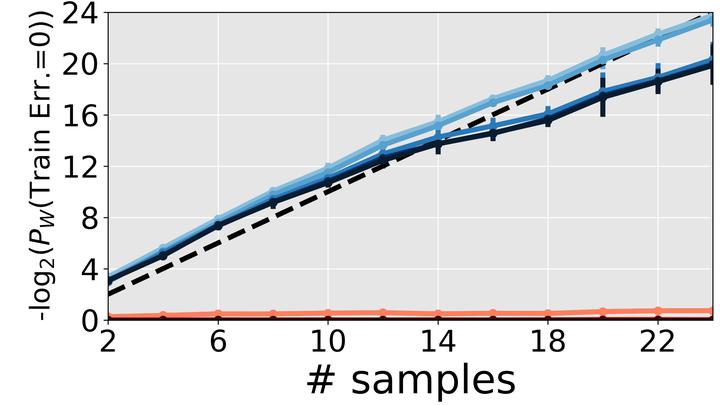}
    &\includegraphics{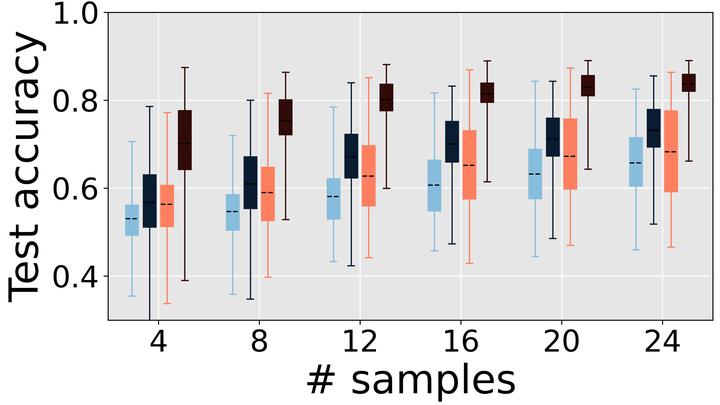}\\
        %%%%%%%%%%%%%%%%%%%%%%%%%%%%%%%%%%%
    % Car vs Ship (230) 
    %%%%%%%%%%%%%%%%%%%%%%%%%%%%%%%%%%%
    &\rotatebox[origin=c]{90}{\textbf{\quad Car vs Ship}}
    &\includegraphics{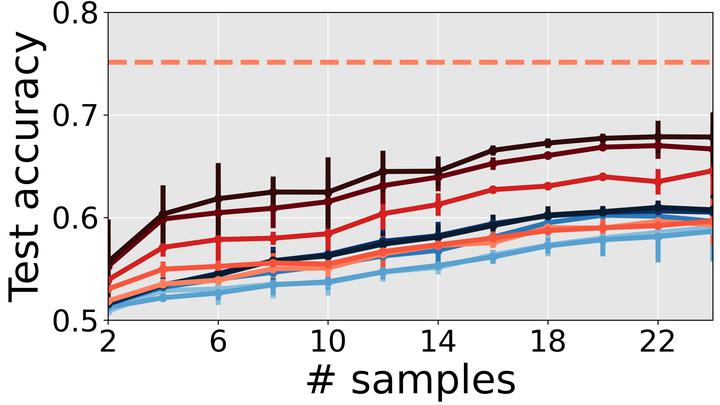}
    &\includegraphics{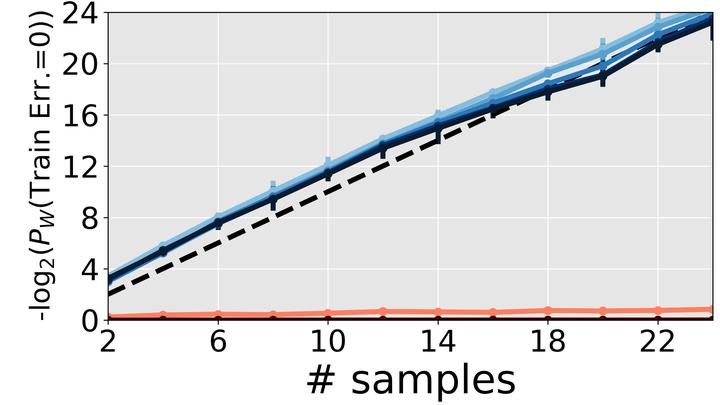}
    &\includegraphics{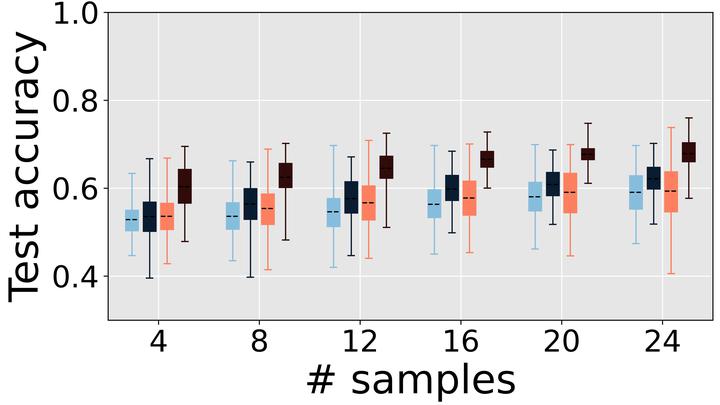}\\
            %%%%%%%%%%%%%%%%%%%%%%%%%%%%%%%%%%%
    % Horse vs Truck (243) 
    %%%%%%%%%%%%%%%%%%%%%%%%%%%%%%%%%%%
    &\rotatebox[origin=c]{90}{\textbf{\quad Horse vs Truck}}
    &\includegraphics{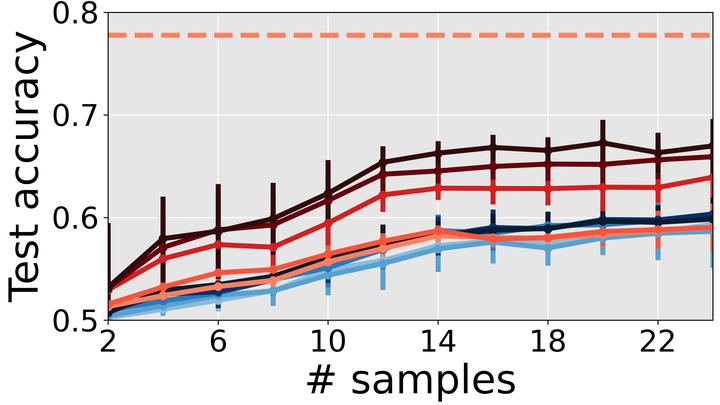}
    &\includegraphics{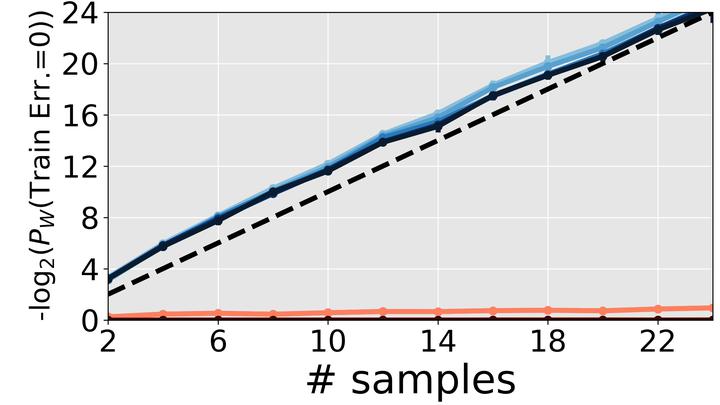}
    &\includegraphics{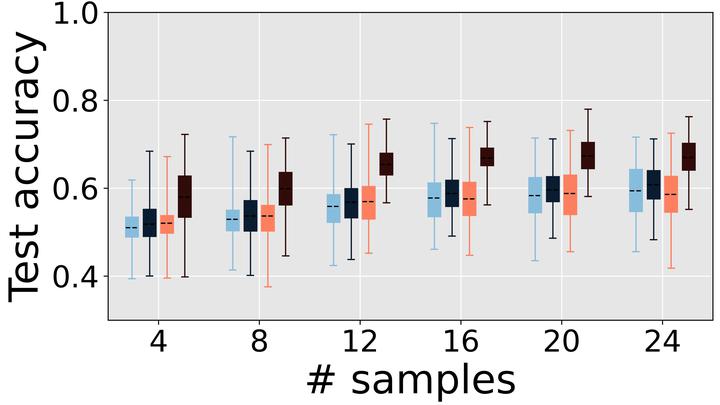}
    \\
    &\multicolumn{4}{c}{
    \includegraphics[width=0.75\linewidth]{figures/different_widths/legend_widths_lenet.png}
    }
    \end{tabularx}
    \caption{\textbf{Increasing width is a positive optimization bias.} 
    The rows represent different pairs of classes from the CIFAR10 dataset using the LeNet architecture.
    \textbf{Column 1:} Mean test accuracy vs number
    of training samples across network widths.
    Wider networks result in improvement for \sgd{}, whereas \gnc{} improves for width up to 2/6 and then stagnates.
    Thus, for increasing width, \sgd{} has a bias towards better generalizing networks independent of an architectural bias.
    However, overfitting does not occur for \gnc{}, implying that overparameterization does not hurt generalization. 
    \textbf{Column 2:} 
    We report the negative log probability of \gnc{} to find a network fitting the training data (${\Pr_W(\TrainError=0)}$).
    Note that this is constant or improves slightly with the width, which indicates that the pool of ``fitting networks'' does not change when the width increases.  
    \textbf{Column 3:} We report the combined distribution of the four subsets for the extreme width factors.
    While for \gnc{}, the entire distribution of models aligns, for \sgd{}, the wider networks are considerably better than the narrow networks (and the \gnc{} networks).
    For more details, please refer to Section~\ref{sec:width}.}
    \label{fig:different_widths_cifar}
\end{figure*}

\begin{figure*}
\centering
\setlength\tabcolsep{0pt}
\adjustboxset{width=\linewidth,valign=c}
\centering
\begin{tabularx}{0.5\linewidth}{
@{}l 
S{p{0.02\textwidth}} 
*{2}{S{p{0.24\textwidth}}}}
    %%%%%%%%%%%%%%%%%%%%%%%%%%
    &
    & \multicolumn{1}{c}{\textbf{\quad \enspace Mean test accuracy}}
    & \multicolumn{1}{c}
    {\textbf{\quad \enspace \makecell{Probability to sample\\ 100\% train accuracy}}}
    \\
    %%%%%%%%%%%%%%%%%%%%%%%%%%%%%%%%%%%
    % 3 vs 5 (204)
    %%%%%%%%%%%%%%%%%%%%%%%%%%%%%%%%%%%
    &\rotatebox[origin=c]{90}{\textbf{\quad 3 vs 5}}
    &\includegraphics{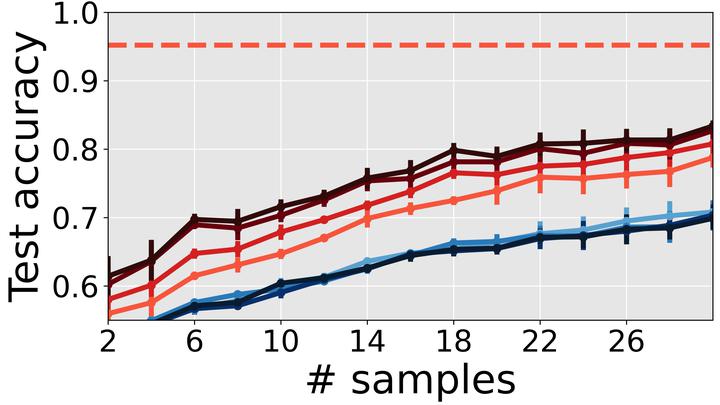}
    &\includegraphics{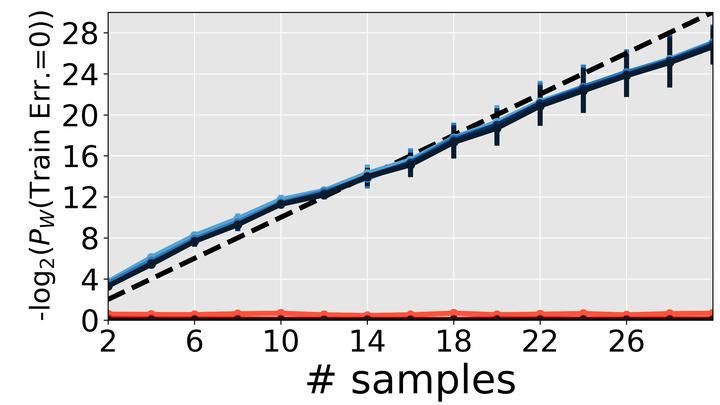}
    \\
    %%%%%%%%%%%%%%%%%%%%%%%%%%%%%%%%%%%
    % Plane vs Frog (224)
    %%%%%%%%%%%%%%%%%%%%%%%%%%%%%%%%%%%
    &\rotatebox[origin=c]{90}{\textbf{\quad Plane vs Frog}}
    &\includegraphics{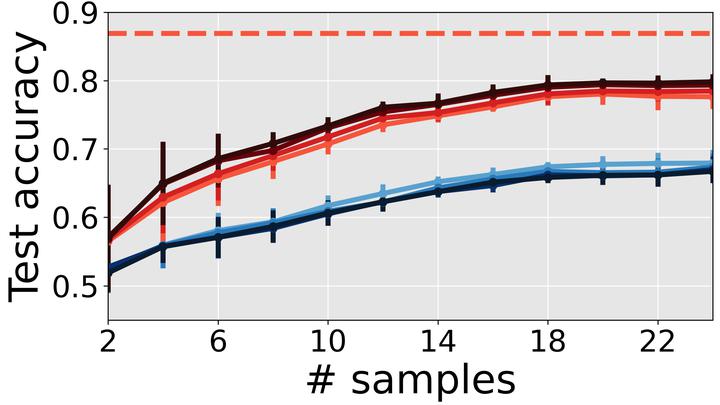}
    &\includegraphics{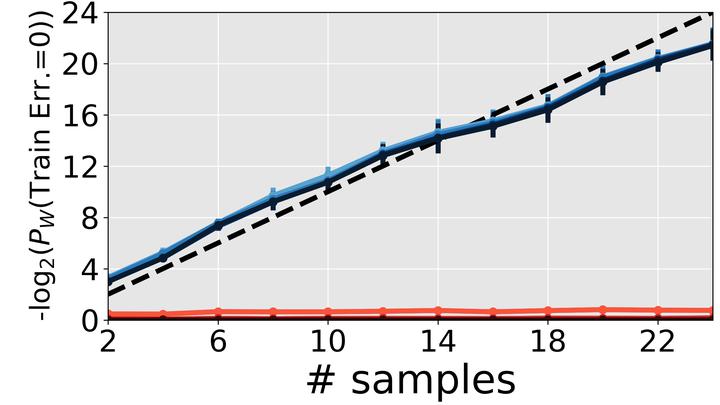}
    \\
    %%%%%%%%%%%%%%%%%%%%%%%%%%%%%%%%%%%
    &\multicolumn{3}{c}{
    \includegraphics[width=0.5\linewidth]{figures/mlp/legend_widths_mlp.png}
    }
    \end{tabularx}
    \caption{\textbf{Increasing width is a positive optimization bias.} 
    The rows represent classes 3 vs 5 from MNIST and Plane vs Frog from CIFAR10 using the MLP architecture.
    \textbf{Left:} Mean test accuracy vs number
    of training samples across network widths.
    Wider networks result in improvement for \sgd{}, whereas \gnc{} stagnates.
    Thus, for increasing width, \sgd{} has a bias towards better generalizing networks independent of an architectural bias. 
    However, overfitting does not occur for \gnc{}, implying that overparameterization does not hurt generalization. 
    \textbf{Right:} 
    We report the negative log probability of \gnc{} to find a network fitting the training data (${\Pr_W(\TrainError=0)}$).
    Note that this does not change with the width, which indicates that the pool of ``fitting networks'' does not change when the width increases.
    For more details, please refer to Section~\ref{sec:width}.}
    \label{fig:different_widths_appendix_mlp}
\end{figure*}

\begin{figure*}
\setlength\tabcolsep{0pt}
\adjustboxset{width=\linewidth,valign=c}
\centering
\begin{tabularx}{0.75\linewidth}{
@{}l 
S{p{0.02\textwidth}} 
*{2}{S{p{0.35\textwidth}}}}
    %%%%%%%%%%%%%%%%%%%%%%%%%%
    &
    & \multicolumn{1}{c}{\quad \enspace \textbf{Mean test accuracy}}
    & \multicolumn{1}{c}
    {\textbf{\quad \enspace \makecell{Probability to sample\\ 100\% train accuracy}}}\\
    %%%%%%%%%%%%%%%%%%%%%%%%%%%%%%%%%%%
    % 0 vs 7 (202)
    %%%%%%%%%%%%%%%%%%%%%%%%%%%%%%%%%%%
    &\rotatebox[origin=c]{90}{\textbf{\quad 0 vs 7}}
    &\includegraphics{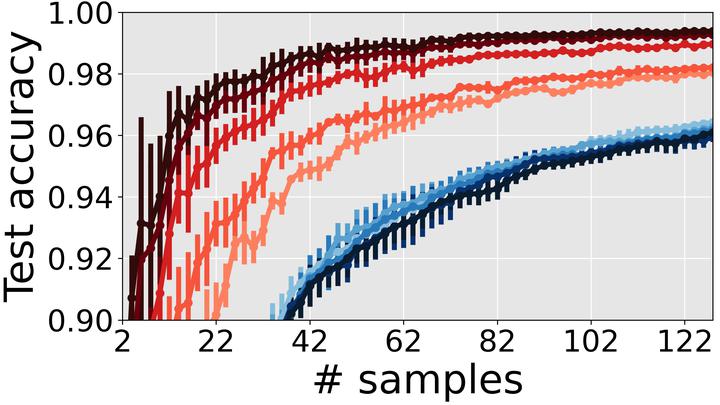}
    &\includegraphics{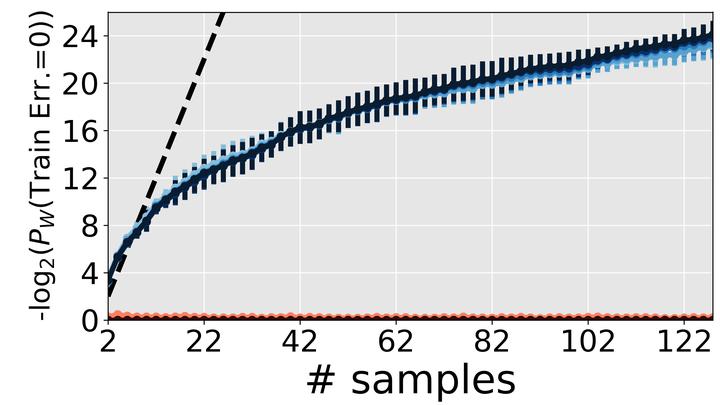}\\
    %%%%%
    & \multicolumn{3}{c}{\includegraphics[width=0.75\linewidth]
    {figures/different_widths/legend_widths_lenet.png}
    }
    \end{tabularx}
    \caption{\textbf{Trend of \sgd{} vs \gnc{} when increasing width remains consistent with larger training set.}
    \textbf{Left:} Mean test accuracy vs number of training samples across network widths.
    Wider networks result in improvement for \sgd{}, whereas \gnc{} stagnates.
    Thus, for increasing width, \sgd{} has a bias towards better generalizing networks independent of an architectural bias.
    However, overfitting does not occur for \gnc{}, implying that overparameterization does not hurt generalization. 
    \textbf{Right:} 
    We report the negative log probability of \gnc{} to find a network fitting the training data (${\Pr_W(\TrainError=0)}$).
    This number remains the same for different widths, indicating that
    the pool of “fitting networks” does not change with increasing
    width.
    The fact that we see the same trend for 128 samples as we saw for 32 samples (MNIST) and 24 samples (CIFAR10) suggests that the results stated in the previous parts are not confined only to the low sample regime, at least for simpler problems like 0 vs 7.}
\label{fig:widths_07_128samples}
\end{figure*}

We further examine the performance of various network widths along different training set sizes. Figures~\ref{fig:different_widths_mnist_seeds} and~\ref{fig:different_widths_cifar} replicate the analysis from Figure~\ref{fig:different_widths} for the LeNet architecture but for different class pairs from the MNIST and CIFAR10 datasets, Figure~\ref{fig:different_widths_appendix_mlp} replicates the same for MLP. 
Moreover, Figure~\ref{fig:widths_07_128samples} presents results for classes 0 and 7 from MNIST with a training size of 128 to illustrate the consistency of our findings for larger training sets.

\setlength\cellspacetoplimit{0pt}
\setlength\cellspacebottomlimit{0pt}
\renewcommand\tabularxcolumn[1]{>{\centering\arraybackslash}S{p{#1}}}

%%%%%%% figure 4

%%%%%%% figure 32

\setlength\cellspacetoplimit{0pt}
\setlength\cellspacebottomlimit{0pt}
\renewcommand\tabularxcolumn[1]{>{\centering\arraybackslash}S{p{#1}}}

%%%%%%%%%%%%
%%%%seed 219
%%%%%%%%%%%%%%%

\setlength\cellspacetoplimit{0pt}
\setlength\cellspacebottomlimit{0pt}
\renewcommand\tabularxcolumn[1]{>{\centering\arraybackslash}S{p{#1}}}
\setlength\cellspacetoplimit{0pt}
\setlength\cellspacebottomlimit{0pt}
\renewcommand\tabularxcolumn[1]{>{\centering\arraybackslash}S{p{#1}}}

\setlength\cellspacetoplimit{0pt}
\setlength\cellspacebottomlimit{0pt}
\renewcommand\tabularxcolumn[1]{>{\centering\arraybackslash}S{p{#1}}}

%%%%%%%% mlp figure for width %%%%%%%
%%%%%%%%%%  128 samples %%%%%%%%%

\section{Overparameterization with Increasing Depth}
\label{sec:depth-appendix}
In the following section, we elaborate on the experiments of the network \emph{depth} analysis described in section~\ref{sec:depth}.

The different layer configurations for the LeNet, specifically the number of convolutional and fully connected layers, along with the corresponding parameter counts are depicted for the MNIST dataset in Table~\ref{tab:depth_parameters_mnist} and for the CIFAR10 dataset in Table~\ref{tab:depth_parameters_cifar}. 
These experiments consider a \emph{width} of $\nicefrac{2}{6}$ for the network, as detailed in sections~\ref{sec:width} and Appendix~\ref{sec:width-appendix}.
In our notation, \emph{c} represents a convolutional layer, and \emph{f} represents a fully connected layer.

\begin{figure*}[htb]
\centering
\setlength\tabcolsep{0pt}
\adjustboxset{width=\linewidth,valign=c}
\centering
\begin{tabularx}{0.75\linewidth}{
@{}l 
S{p{0.02\textwidth}} 
*{3}{S{p{0.24\textwidth}}}}
    %%%%%%%%%%%%%%%%%%%%%%%%%%
    &
    & \multicolumn{1}{c}{\textbf{\qquad Mean test accuracy}}
    & \multicolumn{1}{c}
    {\textbf{\quad \enspace \makecell{Probability to sample\\ 100\% train accuracy}}}
    & \multicolumn{1}{c}{\textbf{\makecell{\quad Distribution of\\ \quad test accuracies}}}\\
    %%%%%%%%%%%%%%%%%%%%%%%%%%%%%%%%%%%
    % 202 - 0 vs 7
    %%%%%%%%%%%%%%%%%%%%%%%%%%%%%%%%%%%
    &\rotatebox[origin=c]{90}{\textbf{\quad 0 vs 7}}
    &\includegraphics{figures/depth/mnist/different_samples/depth_acc_plot_mnist_202.jpg}
    &\includegraphics{figures/depth/mnist/different_samples/depth_num_models_plot_mnist_202_lenet2.jpg}
    &\includegraphics{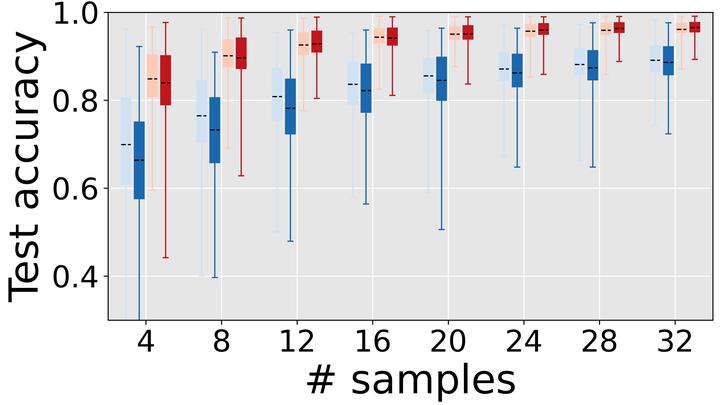}
    \\
    %%%%%%%%%%%%%%%%%%%%%%%%%%%%%%%%%%%
    % 203
    %%%%%%%%%%%%%%%%%%%%%%%%%%%%%%%%%%%
    &\rotatebox[origin=c]{90}{\textbf{\quad 0 vs 8}}
    &\includegraphics{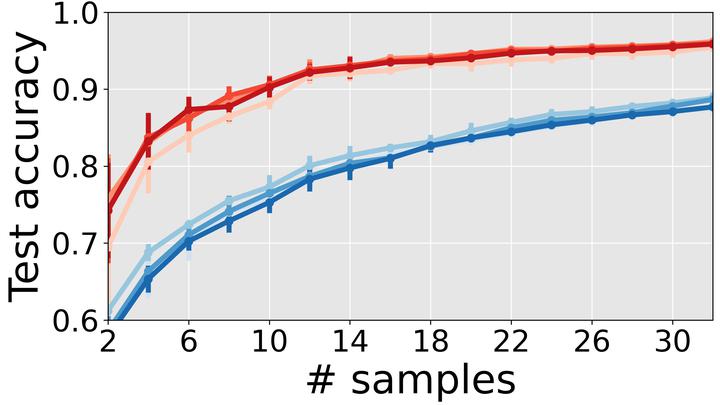}
    &\includegraphics{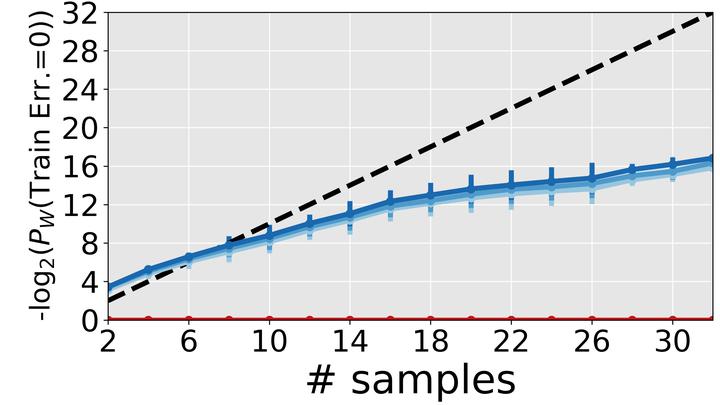}
    &\includegraphics{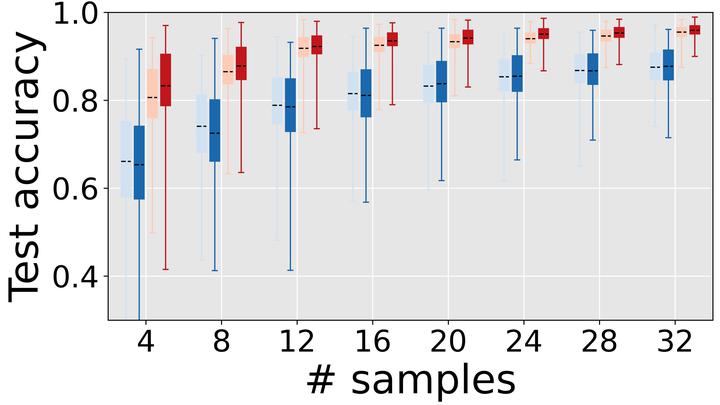}
    \\
   %%%%%%%%%%%%%%%%%%%%%%%%%%%%%%%%%%%
    % 204 (3 vs 5)
    %%%%%%%%%%%%%%%%%%%%%%%%%%%%%%%%%%%
    &\rotatebox[origin=c]{90}{\textbf{\quad 3 vs 5}}
    &\includegraphics{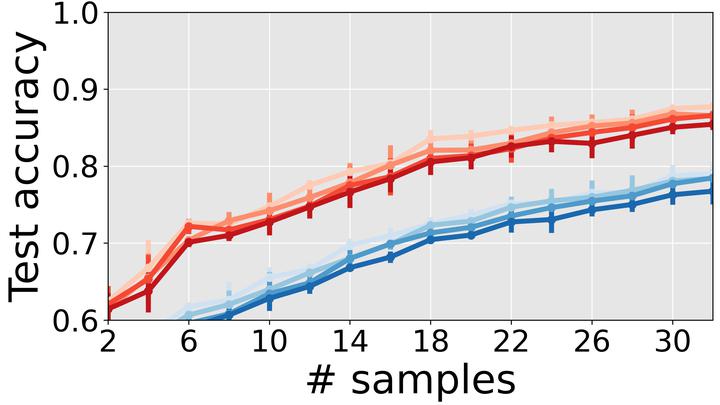}
    &\includegraphics{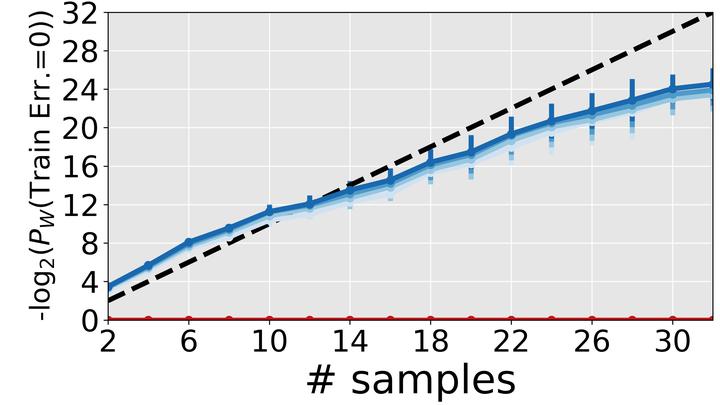}
    &\includegraphics{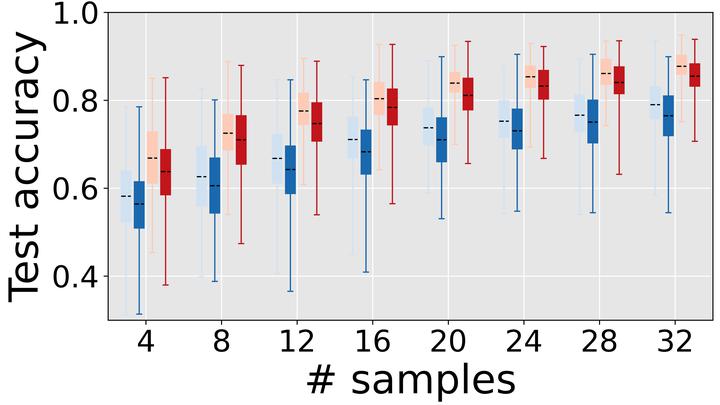}
        \\
%%%%%%%%%%%
%%%%%%%%%%%
%%%%%%%%%%%%%%%%%%%%%%%%%%%%%%%%%%%
    % \textbf{Plane vs Frog (224)}
    %%%%%%%%%%%%%%%%%%%%%%%%%%%%%%%%%%%
    &\rotatebox[origin=c]{90}{\textbf{\quad Plane vs Frog}}
    &\includegraphics{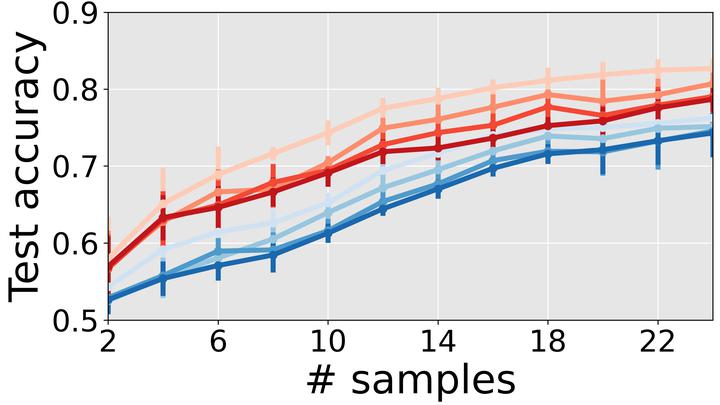}
    &\includegraphics{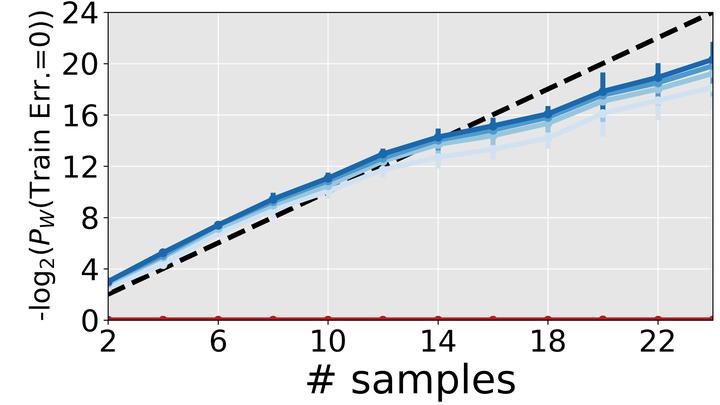}
    &\includegraphics{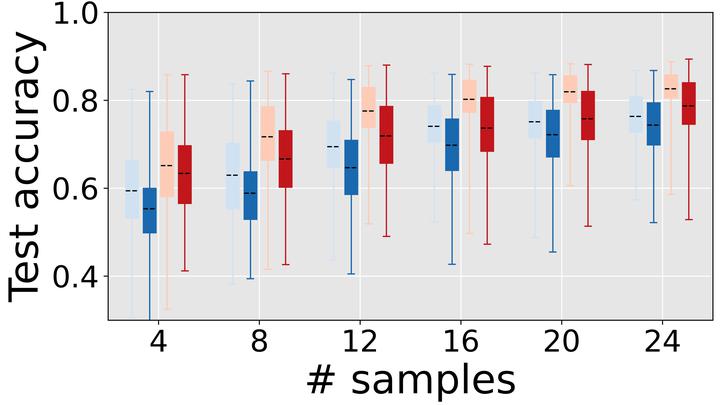}
    \\
    %%%%%%%%%%%%%%%%%%%%%%%%%%%%%%%%%%%
    % Bird vs Ship (201)
    %%%%%%%%%%%%%%%%%%%%%%%%%%%%%%%%%%%
    &\rotatebox[origin=c]{90}{\textbf{\quad Bird vs Ship}}
    &\includegraphics{figures/depth/cifar/seed201/different_samples/depth_acc_plot_cifar_201.jpg}
    &\includegraphics{figures/depth/cifar/seed201/different_samples/depth_num_models_plot_cifar_201_lenet2.jpg}
    &\includegraphics{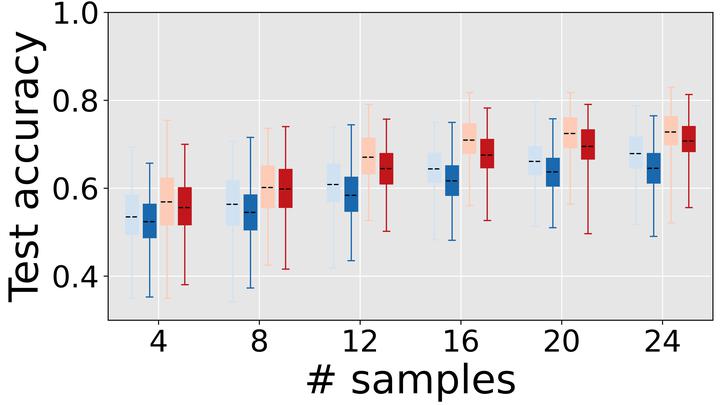}
    \\
%%%%%%%%%%%%%%%%%%%%%%%%%%%%%%%%%%%
    % \textbf{Deer vs Truck (219)}
    %%%%%%%%%%%%%%%%%%%%%%%%%%%%%%%%%%%
    &\rotatebox[origin=c]{90}{\textbf{\quad Deer vs Truck}}
    &\includegraphics{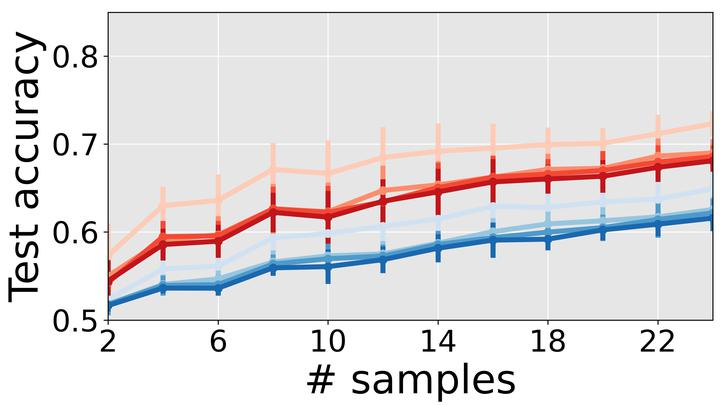}
    &\includegraphics{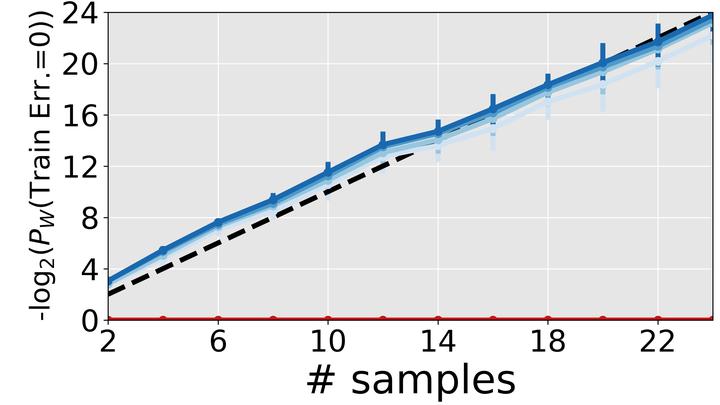}
    &\includegraphics{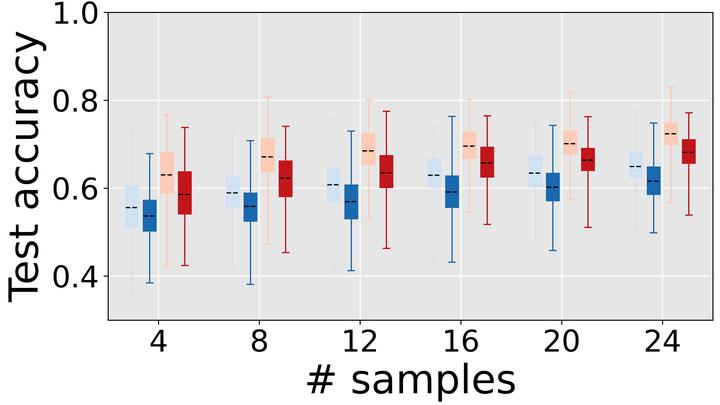}
    \\

    &\multicolumn{4}{c}{
    \includegraphics[width=0.75\linewidth]{figures/depth/legend_depth_lenet.png}}
    \end{tabularx}
    \caption{\textbf{Increasing depth is a negative architectural bias.}
    The rows correspond to different pairs of classes from the MNIST and CIFAR10 datasets using the LeNet architecture.
    Configuration ``2c-1f" means two convolutional layers followed by a fully connected layer.
    \textbf{Column 1:} Mean
    test accuracy vs number of training samples across network depths. 
    As the network becomes deeper, \gnc{} always gets worse, whereas \sgd{} stagnates for simple problems (top) and gets worse for harder problems (bottom). Thus, overparameterization in terms of
    depth results in overfitting instead of better generalization, unlike
    for the width.  
    Since both \gnc{} and \sgd{} follow a similar trend, the decrease in performance with increased depth can be attributed
    to architectural bias.
    \textbf{Column 2:}  
    We report the negative log probability of \gnc{} to find a network fitting the training data (${\Pr_W(\TrainError=0)}$).
    Deeper networks have a lower probability for \gnc{} to fit the training data, indicating that the network produces more complex functions.
    \textbf{Column 3:} We report the combined distribution of the four subsets for the extreme number of layers (``1c-1f", ``2c-3f").
    For more details, please refer to Section~\ref{sec:depth}.
    }
    \label{fig:different_depths_mnist}
\end{figure*}

\begin{figure*}
\centering
\setlength\tabcolsep{0pt}
\adjustboxset{width=\linewidth,valign=c}
\centering
\begin{tabularx}{0.5\linewidth}{
@{}l 
S{p{0.02\textwidth}} 
*{2}{S{p{0.24\textwidth}}}}
    %%%%%%%%%%%%%%%%%%%%%%%%%%
    &
    & \multicolumn{1}{c}{\textbf{\quad \enspace Mean test accuracy}}
    & \multicolumn{1}{c}
    {\textbf{\quad \enspace \makecell{Probability to sample\\ 100\% train accuracy}}}
    \\
    %%%%%%%%%%%%%%%%%%%%%%%%%%%%%%%%%%%
    % 3 vs 5 (204)
    %%%%%%%%%%%%%%%%%%%%%%%%%%%%%%%%%%%
    &\rotatebox[origin=c]{90}{\textbf{\quad 3 vs 5}}
    &\includegraphics{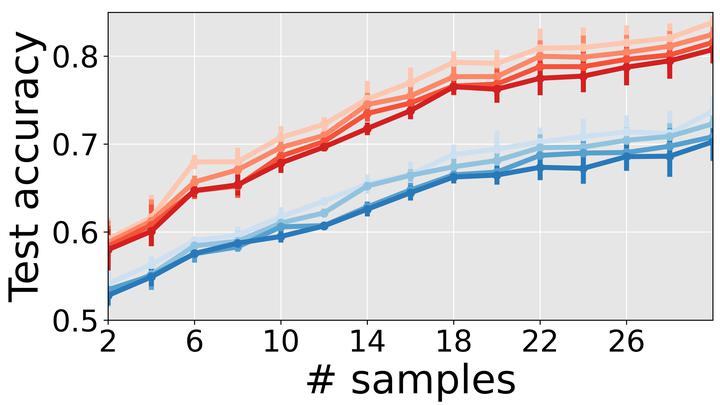}
    &\includegraphics{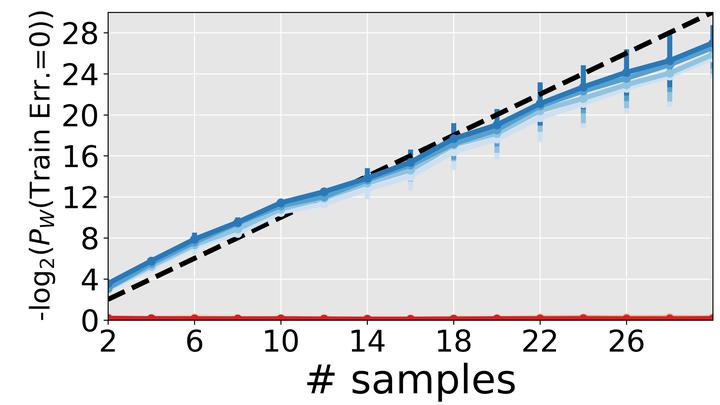}
    \\
    %%%%%%%%%%%%%%%%%%%%%%%%%%%%%%%%%%%
    % Plane vs Frog (224)
    %%%%%%%%%%%%%%%%%%%%%%%%%%%%%%%%%%%
    &\rotatebox[origin=c]{90}{\textbf{\quad Plane vs Frog}}
    &\includegraphics{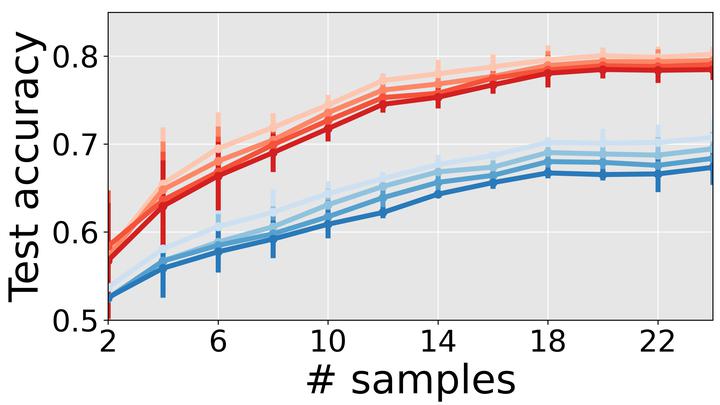}
    &\includegraphics{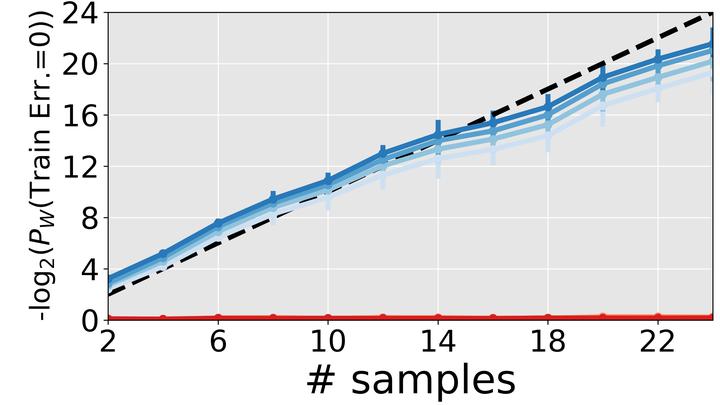}
    \\
    %%%%%%%%%%%%%%%%%%%%%%%%%%%%%%%%%%%
    &\multicolumn{3}{c}{
    \includegraphics[width=0.5\linewidth]{figures/mlp/legend_depth_mlp.png}
    }
    \end{tabularx}
    \caption{\textbf{Increasing depth is a negative architectural bias.}
    The rows represent classes 3 vs 5 from MNIST and Plane vs Frog from CIFAR10 using the MLP architecture.
    \textbf{Left:} Mean
    test accuracy vs number of training samples across network depths. 
    As the network becomes deeper, \gnc{} and \sgd{} get worse. Thus, overparameterization in terms of depth results in overfitting instead of better generalization, unlike
    for the width.  
    Since both \gnc{} and \sgd{} follow a similar trend, the decrease in performance with increased depth can be attributed
    to architectural bias.
    \textbf{Right:}  
    We report the negative log probability of \gnc{} to find a network fitting the training data (${\Pr_W(\TrainError=0)}$).
    Deeper networks have a lower probability for \gnc{} to fit the training data, indicating that the network produces more complex functions.
    For more details, please refer to Section~\ref{sec:depth}.
    }
    \label{fig:different_depths_appendix_mlp}
\end{figure*}

\begin{figure*}
\setlength\tabcolsep{0pt}
\adjustboxset{width=\linewidth,valign=c}
\centering
\begin{tabularx}{0.75\linewidth}{
@{}l 
S{p{0.02\textwidth}} 
*{2}{S{p{0.35\textwidth}}}}
    %%%%%%%%%%%%%%%%%%%%%%%%%%
    &
    & \multicolumn{1}{c}{\quad \enspace \textbf{Mean test accuracy}}
    & \multicolumn{1}{c}
    {\textbf{\quad \enspace \makecell{Probability to sample\\ 100\% train accuracy}}}\\
    %%%%%%%%%%%%%%%%%%%%%%%%%%%%%%%%%%%
    % 202 - 0 vs 7
    %%%%%%%%%%%%%%%%%%%%%%%%%%%%%%%%%%%
    &\rotatebox[origin=c]{90}{\textbf{\quad 0 vs 7}}
    &\includegraphics{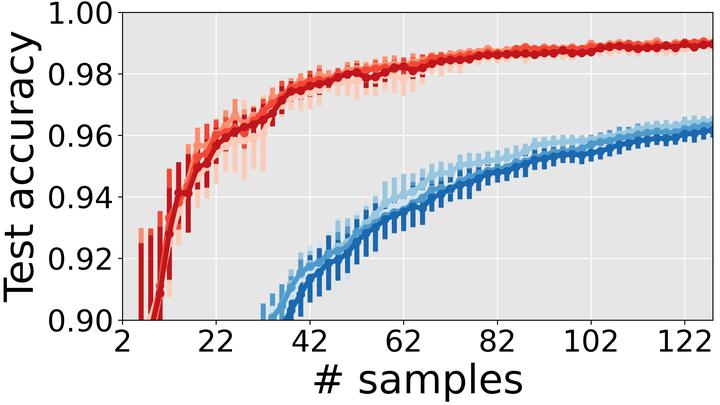}
    &\includegraphics{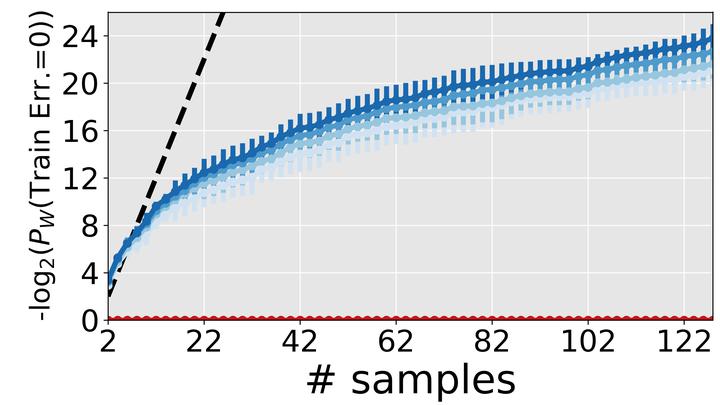}
    \\
    %%%%%
    & \multicolumn{3}{c}{\includegraphics[width=0.75\linewidth]
    {figures/depth/legend_depth_lenet.png}
    }
    \end{tabularx}
    \caption{\textbf{Trend of \sgd{} vs \gnc{} when increasing depth remains consistent with larger training set.}
    Configuration ``2c-1f" means two convolutional layers followed by a fully connected layer.
    \textbf{Left:} Mean
    test accuracy vs number of training samples across network depths. 
    As the network becomes deeper, \gnc{} gets worse, whereas \sgd{} stagnates. Thus, overparameterization in terms of
    depth results in overfitting instead of better generalization, unlike
    for the width.  
    \textbf{Right:} 
    We report the negative log probability of \gnc{} to find a network fitting the training data (${\Pr_W(\TrainError=0)}$).
    Deeper networks have a lower probability for \gnc{} to fit the training data, indicating that the network produces more complex functions.
    The fact that we see the same trend for 128 samples as we saw for 32 samples for classes 0 vs 7, suggests that the results stated in the previous parts are not confined only to the low sample regime, at least for simpler problems like 0 vs 7.}
\label{fig:depths_07_128samples}
\end{figure*}

\begin{figure*}[ht]
\centering
\setlength\tabcolsep{0pt}
\adjustboxset{width=\linewidth,valign=c}
\centering
\begin{tabularx}{1.0\linewidth}{
@{}l 
S{p{0.00\linewidth}} 
S{p{0.045\textwidth}} 
    *{4}{S{p{0.22195\textwidth}}} 
    S{p{0.051\textwidth}}}
    %%%%%%%%%%%%%%%%%%%%%%%%%%
    &
    &
    & \multicolumn{1}{c}{$\mathbf{1}$ \textbf{conv}, $\mathbf{1}$ \textbf{fc}}
    & \multicolumn{1}{c}{$\mathbf{2}$ \textbf{conv}, $\mathbf{1}$ \textbf{fc}}
    & \multicolumn{1}{c}{$\mathbf{2}$ \textbf{conv}, $\mathbf{2}$ \textbf{fc}}
    & \multicolumn{1}{c}{\textbf{standard}}
    & \multicolumn{1}{c}{} \\
    %%%%%%%%%%%%%%%%%%%%%%%%%%%%%%%%%%%
    % GNC 1
    %%%%%%%%%%%%%%%%%%%%%%%%%%%%%%%%%%%
    &
    &\rotatebox[origin=c]{90}{\textbf{\makecell{\gnc{} - 4 samples\\Test accuracy}}}
    &\includegraphics{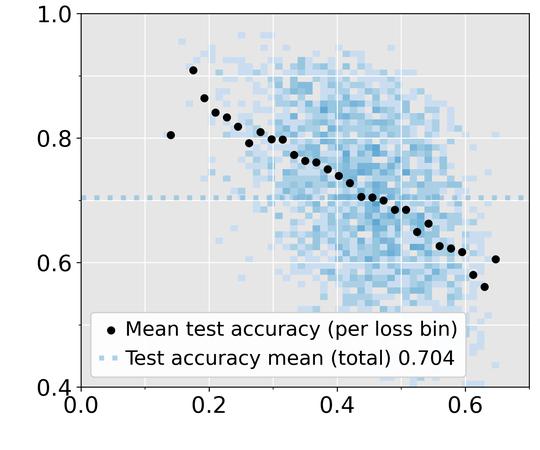}
    &\includegraphics{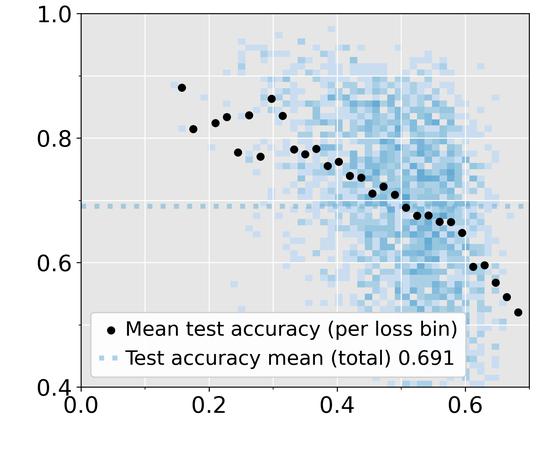}
    &\includegraphics{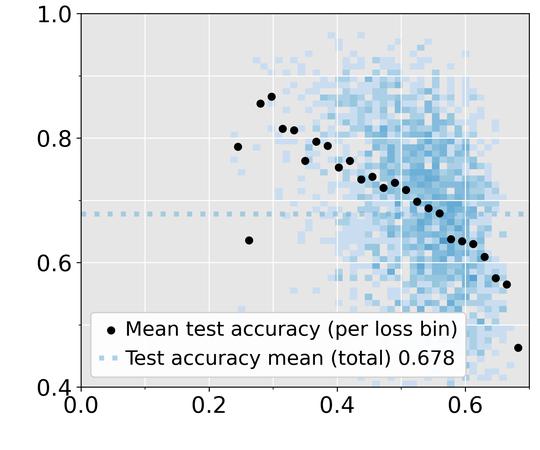}
    &\includegraphics{figures/different_widths/MNIST/seed202/2d_hist_train_loss_normalize_grad_input_test_acc_mnist_guess_initialization_uniform_width033_seed202_permseed202_4_samples.jpg}
    &\includegraphics{figures/different_loss_factor/colorbar.png} \\
    %%%%%%%%%%%%%%%%%%%%%%%%%%%%%%%%%%%
    % SGD 1
    %%%%%%%%%%%%%%%%%%%%%%%%%%%%%%%%%%%
    &
    &\rotatebox[origin=c]{90}{\textbf{\makecell{\sgd{} - 4 samples\\Test accuracy}}}
    &\includegraphics{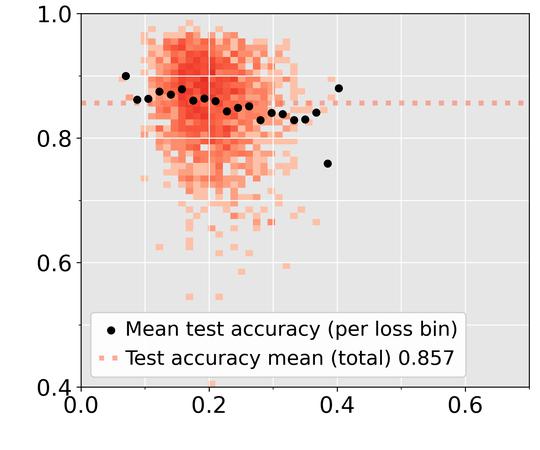}
    &\includegraphics{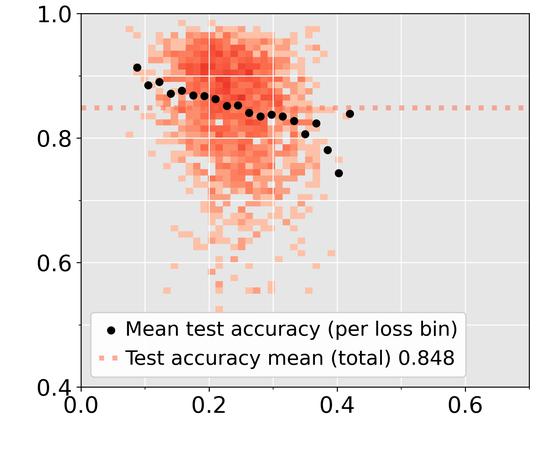}
    &\includegraphics{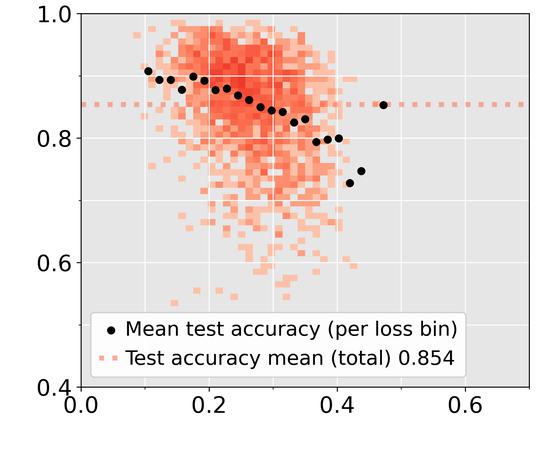}
    &\includegraphics{figures/different_widths/MNIST/seed202/2d_hist_train_loss_normalize_grad_input_test_acc_mnist_sgd_initialization_kaiming_lr01_width033_epoch60_seed202_permseed202_4_samples.jpg}
    &\includegraphics{figures/different_loss_factor/colorbar_SGD.png} \\
    %%%%%%%%%%%%%%%%%%%%%%%%%%%%%%%%%
        %%%%%%%%%%%%%%%%%%%%%%%%%%%%%%%%%%%
    % GNC 1
    %%%%%%%%%%%%%%%%%%%%%%%%%%%%%%%%%%%
    &
    &\rotatebox[origin=c]{90}{\textbf{\makecell{\gnc{} - 32 samples\\Test accuracy}}}
    &\includegraphics{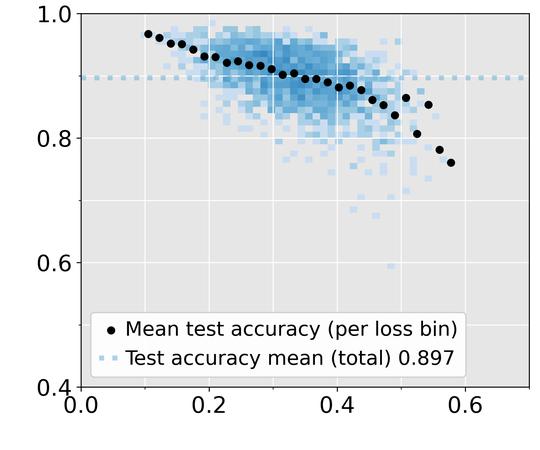}
    &\includegraphics{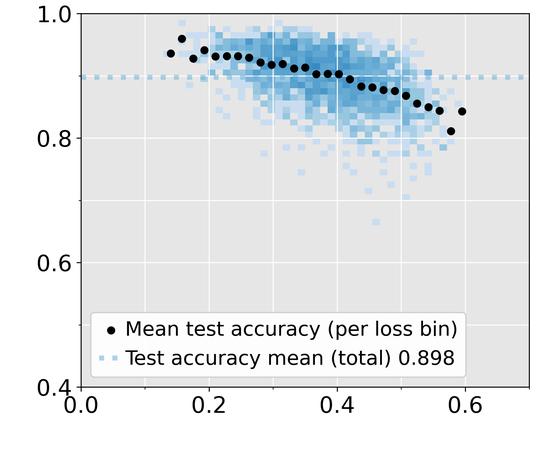}
    &\includegraphics{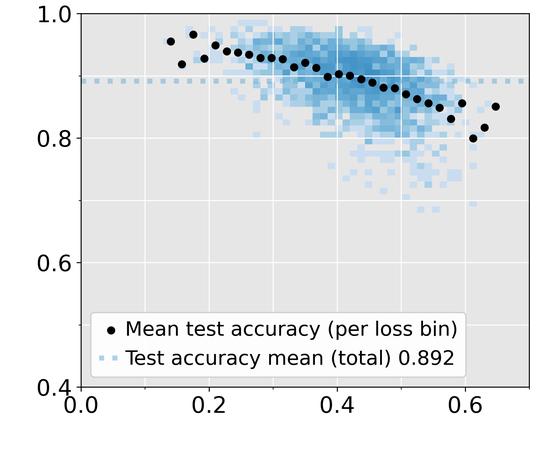}
    &\includegraphics{figures/different_widths/MNIST/seed202/2d_hist_train_loss_normalize_grad_input_test_acc_mnist_guess_initialization_uniform_width033_seed202_permseed202_32_samples.jpg}
    &\includegraphics{figures/different_loss_factor/colorbar.png} \\
        %%%%%%%%%%%%%%%%%%%%%%%%%%%%%%%%%%%
    % SGD 1
    %%%%%%%%%%%%%%%%%%%%%%%%%%%%%%%%%%%
    &
    &\rotatebox[origin=c]{90}{\textbf{\makecell{\sgd{} - 32 samples\\Test accuracy}}}
    &\includegraphics{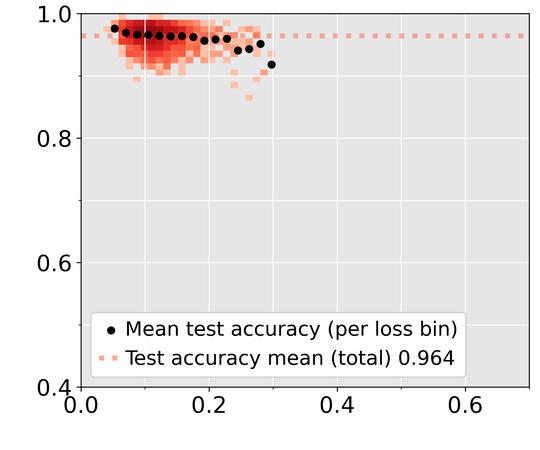}
    &\includegraphics{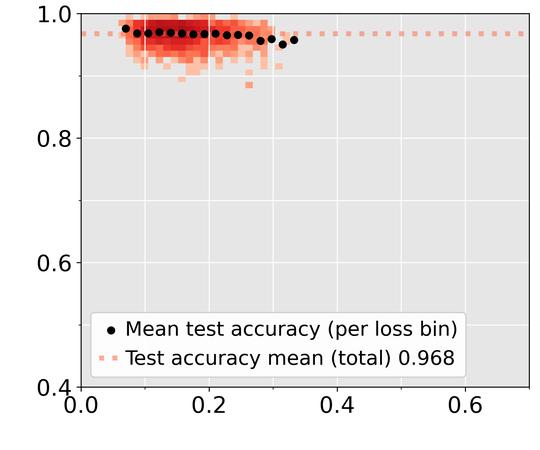}
    &\includegraphics{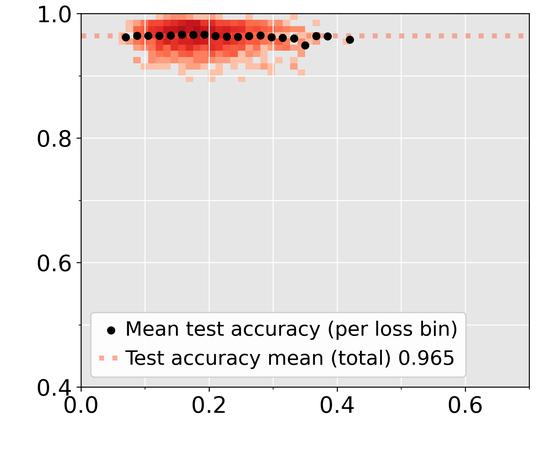}
    &\includegraphics{figures/different_widths/MNIST/seed202/2d_hist_train_loss_normalize_grad_input_test_acc_mnist_sgd_initialization_kaiming_lr01_width033_epoch60_seed202_permseed202_32_samples.jpg}
    &\includegraphics{figures/different_loss_factor/colorbar_SGD.png} \\
    %%%%%%%%%%%%%%%%%%%%%%%%%%
    &
    &
    & \multicolumn{1}{c}{\hspace{1.5em}\textbf{\makecell{Train loss\\(normalized)}}}
    & \multicolumn{1}{c}{\hspace{1.5em}\textbf{\makecell{Train loss\\(normalized)}}}
    & \multicolumn{1}{c}{\hspace{1.5em}\textbf{\makecell{Train loss\\(normalized)}}}
    & \multicolumn{1}{c}{\hspace{1.5em}\textbf{\makecell{Train loss\\(normalized)}}}
    &  \\
    \end{tabularx}
    \caption{\textbf{Qualitative analysis of overparametrization in the depth.} 
    In contrast to increasing width, increasing the depth decreases the geometric margin (higher Lipschitz normalized loss). This decrease is true both for \colorgnc{\textbf{\gnc}} (Rows 1 and 3)  and \colorsgd{\textbf{\sgd}} (Rows 2 and 4).
    We show 500 networks for each depth for MNIST with classes \emph{0} and \emph{7} using a training set of size 4 and 32.}
\label{fig:different_depths_loss_mnist_appendix_4_32_samples}
\end{figure*}

\begin{figure*}[ht]
\centering
\setlength\tabcolsep{0pt}
\adjustboxset{width=\linewidth,valign=c}
\centering
\begin{tabularx}{1.0\linewidth}{
@{}l 
S{p{0.022\linewidth}} 
S{p{0.045\textwidth}} 
    *{4}{S{p{0.22195\textwidth}}} 
    S{p{0.051\textwidth}}}
    %%%%%%%%%%%%%%%%%%%%%%%%%%
    &
    &
    & \multicolumn{1}{c}{$\mathbf{1}$ \textbf{conv}, $\mathbf{1}$ \textbf{fc}}
    & \multicolumn{1}{c}{$\mathbf{2}$ \textbf{conv}, $\mathbf{1}$ \textbf{fc}}
    & \multicolumn{1}{c}{$\mathbf{2}$ \textbf{conv}, $\mathbf{2}$ \textbf{fc}}
    & \multicolumn{1}{c}{\textbf{standard}}
    & \multicolumn{1}{c}{} \\
    %%%%%%%%%%%%%%%%%%%%%%%%%%%%%%%%%%%
    % GNC 1
    %%%%%%%%%%%%%%%%%%%%%%%%%%%%%%%%%%%
    % &
    &\multirow{2}{*}[-4em]{\rotatebox{90}{\textbf{3 vs 5}}}
    &\rotatebox[origin=c]{90}{\textbf{\makecell{\gnc{}\\Test accuracy}}}
    &\includegraphics{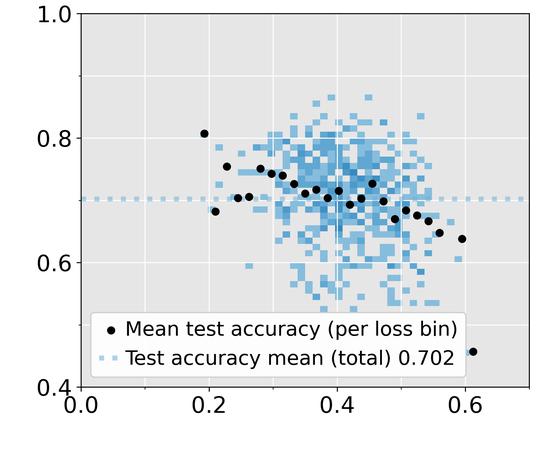}
    &\includegraphics{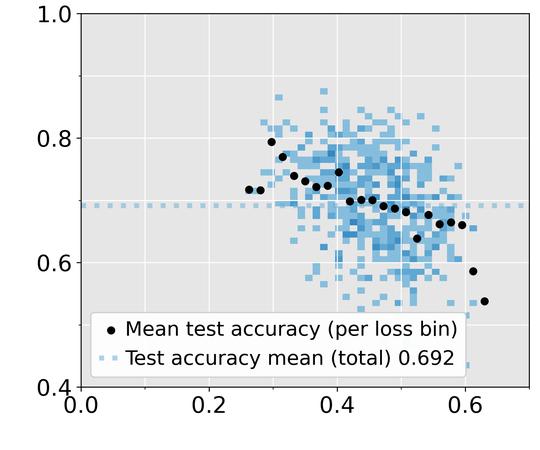}
    &\includegraphics{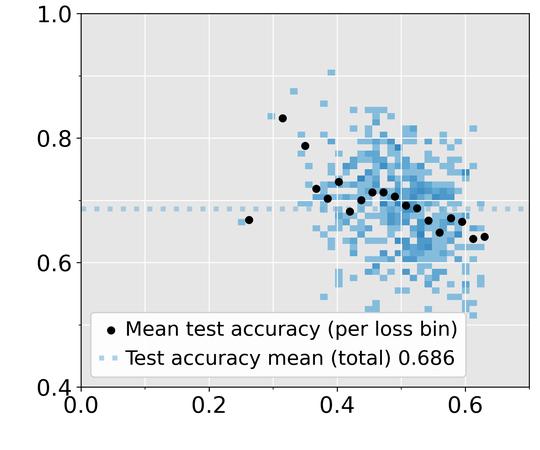}
    &\includegraphics{figures/different_widths/MNIST/seed204/16samples/2d_hist_train_loss_normalize_grad_input_test_acc_mnist_guess_initialization_uniform_width033_seed204_permseed202_16_samples.jpg}
    &\includegraphics{figures/different_loss_factor/colorbar.png} \\
    %%%%%%%%%%%%%%%%%%%%%%%%%%%%%%%%%%%
    % SGD 1
    %%%%%%%%%%%%%%%%%%%%%%%%%%%%%%%%%%%
    &
    &\rotatebox[origin=c]{90}{\textbf{\makecell{\sgd{}\\Test accuracy}}}
    &\includegraphics{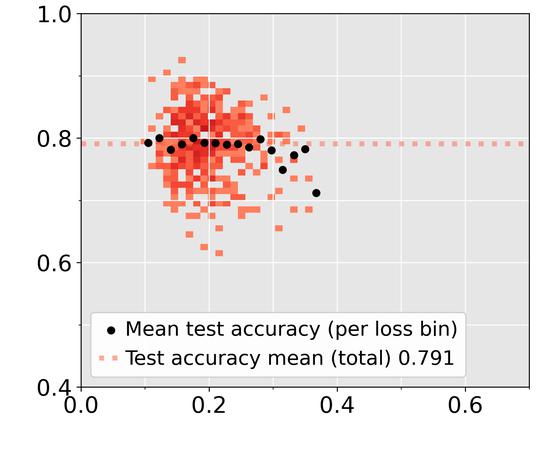}
    &\includegraphics{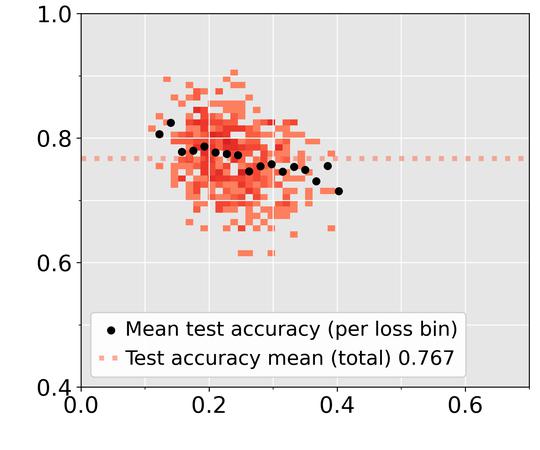}
    &\includegraphics{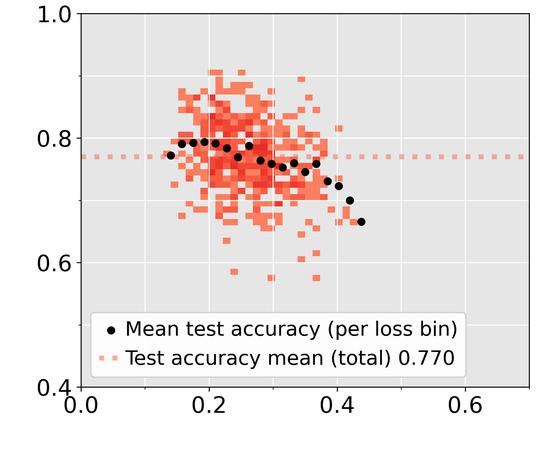}
    &\includegraphics{figures/different_widths/MNIST/seed204/16samples/2d_hist_train_loss_normalize_grad_input_test_acc_mnist_sgd_initialization_kaiming_lr01_width033_epoch60_seed204_permseed202_16_samples.jpg}
    &\includegraphics{figures/different_loss_factor/colorbar_SGD.png} \\
    %%%%%%%%%%%%%%%%%%%%%%%%%%%%%%%%%%%
    % GNC 1
    %%%%%%%%%%%%%%%%%%%%%%%%%%%%%%%%%%%
    &\multirow{2}{*}[-2em]{\rotatebox{90}{\textbf{Bird vs Ship}}}
    &\rotatebox[origin=c]{90}{\textbf{\makecell{\gnc{}\\Test accuracy}}}
    &\includegraphics{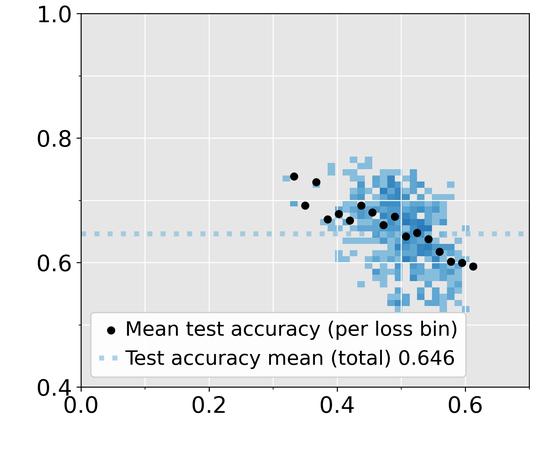}
    &\includegraphics{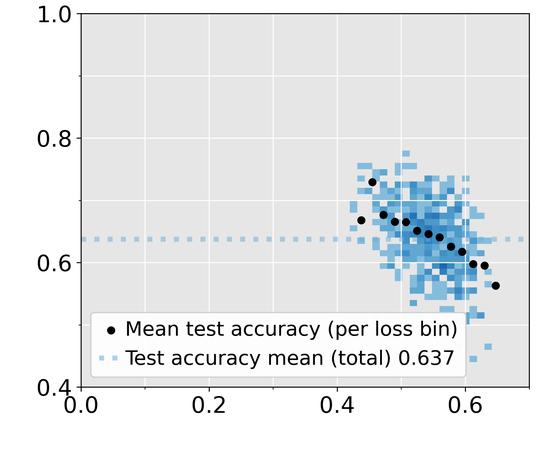}
    &\includegraphics{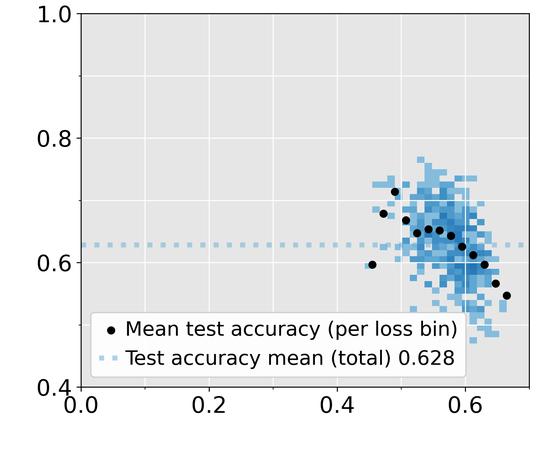}
    &\includegraphics{figures/width_16samples_cifar/2d_hist_train_loss_normalize_grad_input_test_acc_cifar_guess_initialization_uniform_width033_seed201_permseed201_16_samples.jpg}
    &\includegraphics{figures/different_loss_factor/colorbar.png} \\
    %%%%%%%%%%%%%%%%%%%%%%%%%%%%%%%%%%%
    % SGD 1
    %%%%%%%%%%%%%%%%%%%%%%%%%%%%%%%%%%%
    &
    &\rotatebox[origin=c]{90}{\textbf{\makecell{\sgd{}\\Test accuracy}}}
    &\includegraphics{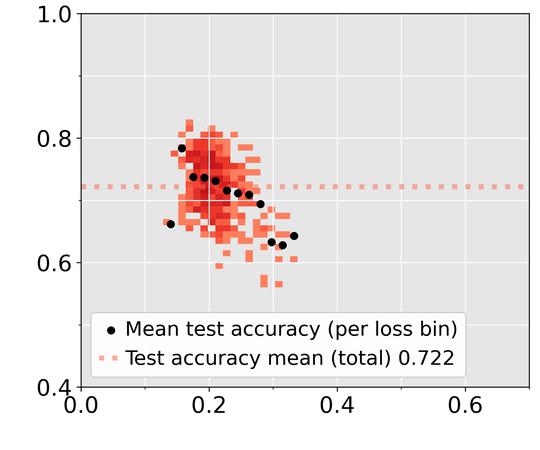}
    &\includegraphics{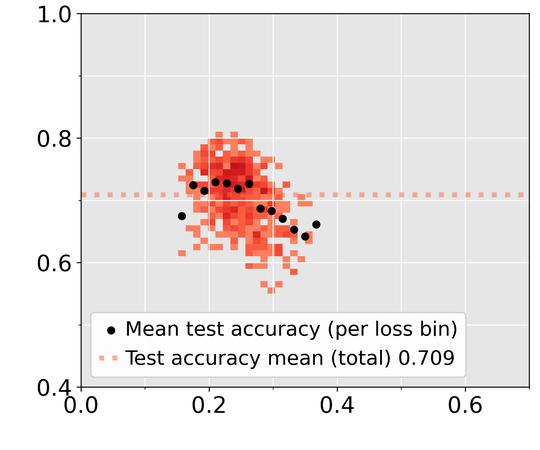}
    &\includegraphics{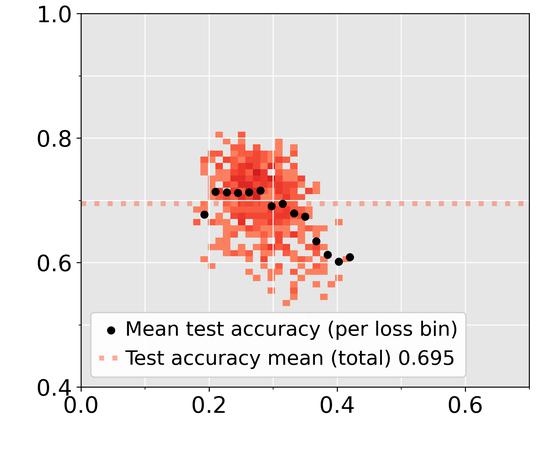}
    &\includegraphics{figures/width_16samples_cifar/2d_hist_train_loss_normalize_grad_input_test_acc_cifar_sgd_initialization_kaiming_lr01_width033_epoch60_seed201_permseed201_16_samples.jpg}
    &\includegraphics{figures/different_loss_factor/colorbar_SGD.png} \\
    %%%%%%%%%%%%%%%%%%%%%%%%%%%%%%%%%
        %%%%%%%%%%%%%%%%%%%%%%%%%%%%%%%%%%%
    % GNC 1
    %%%%%%%%%%%%%%%%%%%%%%%%%%%%%%%%%%%
    % &
    &\multirow{2}{*}[-2em]{\rotatebox{90}{\textbf{Deer vs Truck}}}
    &\rotatebox[origin=c]{90}{\textbf{\makecell{\gnc{}\\Test accuracy}}}
    &\includegraphics{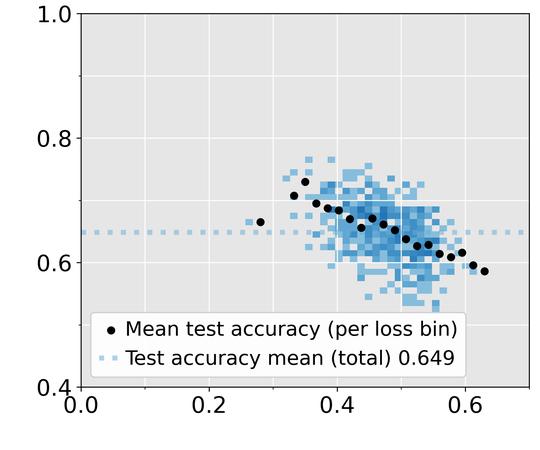}
    &\includegraphics{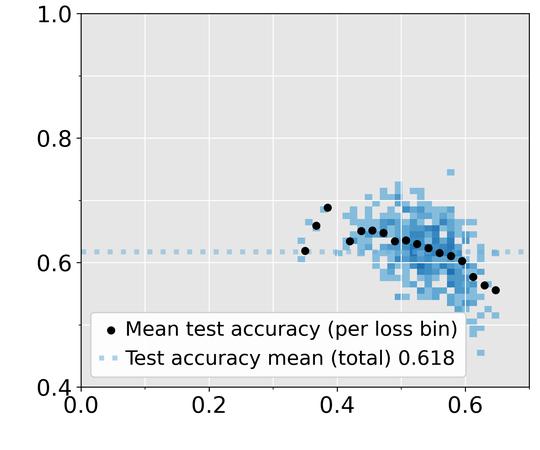}
    &\includegraphics{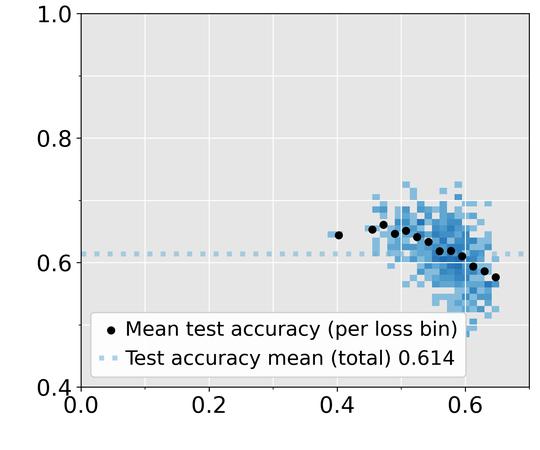}
    &\includegraphics{figures/different_widths/CIFAR10/seed219/16samples/2d_hist_train_loss_normalize_grad_input_test_acc_cifar_guess_initialization_uniform_width033_seed219_permseed201_16_samples.jpg}
    &\includegraphics{figures/different_loss_factor/colorbar.png} \\
    %%%%%%%%%%%%%%%%%%%%%%%%%%%%%%%%%%%
    % SGD 1
    %%%%%%%%%%%%%%%%%%%%%%%%%%%%%%%%%%%
    &
    &\rotatebox[origin=c]{90}{\textbf{\makecell{\sgd{}\\Test accuracy}}}
    &\includegraphics{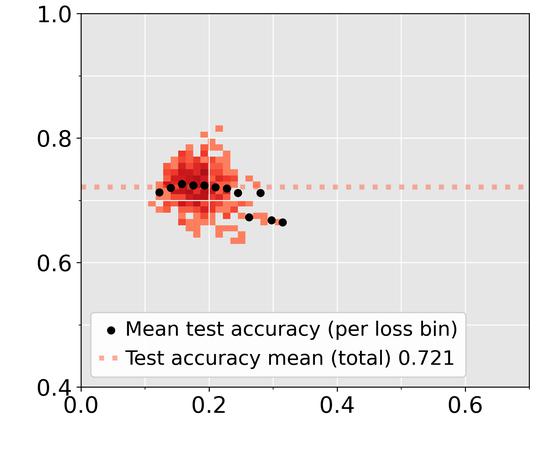}
    &\includegraphics{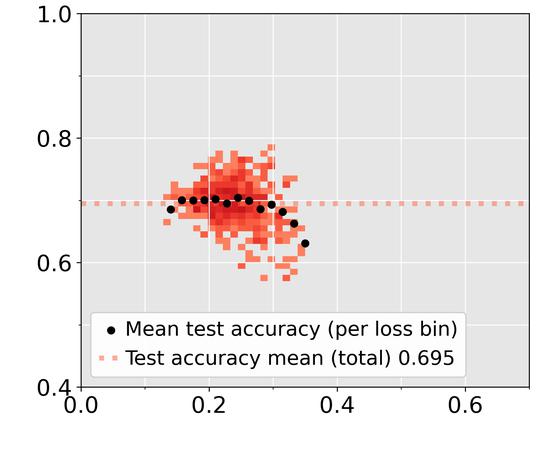}
    &\includegraphics{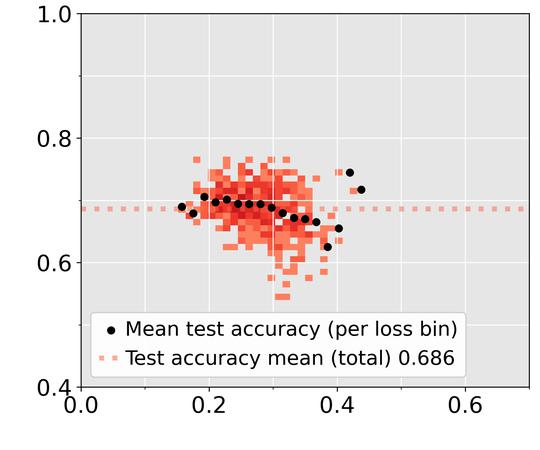}
    &\includegraphics{figures/different_widths/CIFAR10/seed219/16samples/2d_hist_train_loss_normalize_grad_input_test_acc_cifar_sgd_initialization_kaiming_lr01_width033_epoch60_seed219_permseed201_16_samples.jpg}

    &\includegraphics{figures/different_loss_factor/colorbar_SGD.png} \\
    %%%%%%%%%%%%%%%%%%%%%%%%%%%%%%%%%
    &
    &
    & \multicolumn{1}{c}{\hspace{1.5em}\textbf{\makecell{Train loss\\(normalized)}}}
    & \multicolumn{1}{c}{\hspace{1.5em}\textbf{\makecell{Train loss\\(normalized)}}}
    & \multicolumn{1}{c}{\hspace{1.5em}\textbf{\makecell{Train loss\\(normalized)}}}
    & \multicolumn{1}{c}{\hspace{1.5em}\textbf{\makecell{Train loss\\(normalized)}}}
    &  \\
    %%%%%%%%%%%%%%%%%%%%%%%%%%
    \end{tabularx}
    \caption{\textbf{Qualitative analysis of overparametrization in the depth.} 
    In contrast to increasing width, increasing the depth decreases the geometric margin (higher Lipschitz normalized loss) and decreases the test accuracy. This decrease is true both for \colorgnc{\textbf{\gnc}}~(top)  and \colorsgd{\textbf{\sgd}}~(bottom).
    We show 500 networks for each depth for MNIST with classes \emph{3} and \emph{5}, and CIFAR10 with classes Bird vs Ship and Deer vs Truck, using a training set of size 16.}
\label{fig:different_depths_loss_mnist_appendix}
\end{figure*}

In Figure~\ref{fig:different_depths_mnist},  
we show the performance of the LeNet with different numbers of layers along various sizes of training sets, as in Figure~\ref{fig:different_depths}, but for other classes from the MNIST and CIFAR10 datasets. In Figure~\ref{fig:different_depths_appendix_mlp}, we do the same for MLP.
Additionally, in Figure~\ref{fig:depths_07_128samples}, we present this for classes 0 and 7 for a training set of 128 to illustrate that the results hold for a larger number of samples.
In Figures~\ref{fig:different_depths_loss_mnist_appendix_4_32_samples} and~\ref{fig:different_depths_loss_mnist_appendix},
we display the loss changes as in Figure~\ref{fig:different_depths_loss_mnist}, but for a different number of samples and different pairs of classes, respectively.

\setlength\cellspacetoplimit{0pt}
\setlength\cellspacebottomlimit{0pt}
\renewcommand\tabularxcolumn[1]{>{\centering\arraybackslash}S{p{#1}}}

%%%%%%%% mlp figure for depth %%%%%%%

%%%%%%%%%%  128 samples %%%%%%%%%

\setlength\cellspacetoplimit{0pt}
\setlength\cellspacebottomlimit{0pt}
\renewcommand\tabularxcolumn[1]{>{\centering\arraybackslash}S{p{#1}}}

%%%%%%%%%%%%%%%
%%% mnist - seed 202 - 4 and 32 samples
%%%%%%%%%%%%%%%

%%%%%%%%%%%%%%%
%%% mnist - seed 204
%%%%%%%%%%%%%%%

%%%%%%%%%%%%%%%%%%%%%%%%%%%%%%%%%%%%%%%%%%%%%%%%%%%%%%%%%%%%%%%%%%%%%%%%%%%%%%%
%%%%%%%%%%%%%%%%%%%%%%%%%%%%%%%%%%%%%%%%%%%%%%%%%%%%%%%%%%%%%%%%%%%%%%%%%%%%%%%

\end{document}